\documentclass[mnsc,nonblindrev]{informs1}
\OneAndAHalfSpacedXI
\usepackage{rotating}
\usepackage{bbding}
\usepackage[left=1in,top=1in,right=1in,bottom=1in]{geometry}
\usepackage{amssymb,amsmath,amsxtra,amstext,xfrac,mathtools}
\usepackage{graphicx,booktabs,enumerate,paralist,mdwlist}
\usepackage{dsfont,verbatim,latexsym}
\usepackage[titletoc,toc,page]{appendix}
\usepackage{mathrsfs}
\usepackage{bbm}
\usepackage{pdflscape}
\usepackage{multirow}
\usepackage[table]{xcolor}
\newcolumntype{C}[1]{>{\centering\let\newline\\\arraybackslash\hspace{0pt}}m{#1}}
\usepackage{longtable,multirow,threeparttable}
\usepackage{float}
\usepackage{tikz,qtree}
\usetikzlibrary{arrows,positioning,decorations.pathreplacing}
\usepackage[normalem]{ulem}
\usepackage{mwe}
\usepackage{makecell}
\usepackage{eurosym}
\usepackage[english]{babel}
\usepackage[utf8]{inputenc}
\usepackage{fancyhdr}
\usepackage{color}
\usepackage{array}
\usepackage{subcaption}
\usepackage{url}
\usepackage{graphicx}
\usepackage{algorithm}
\usepackage{algpseudocode}
\usepackage[most]{tcolorbox}
\usepackage{xcolor}
\usepackage[colorlinks=TRUE,citecolor=MyDarkBlue,urlcolor=MyDarkBlue,linkcolor=MyDarkBlue]{hyperref}
\definecolor{MyDarkBlue}{RGB}{158,0,0}
\definecolor{buyerB}{RGB}{29,78,216}
\definecolor{sellerS}{RGB}{194,65,12}
\definecolor{titlegold}{RGB}{168,124,18}
\definecolor{proposebg}{RGB}{219,234,254}
\definecolor{proposefg}{RGB}{29,78,216}
\definecolor{rejectbg}{RGB}{254,226,226}
\definecolor{rejectfg}{RGB}{185,28,28}
\definecolor{acceptbg}{RGB}{220,252,231}
\definecolor{acceptfg}{RGB}{21,128,61}

\usepackage{natbib}
\bibpunct[, ]{(}{)}{,}{a}{}{,}%
\def\bibfont{\small}%
\def\bibsep{\smallskipamount}%
\TheoremsNumberedThrough
\EquationsNumberedThrough

\tcbuselibrary{breakable}
\newtcolorbox{promptboxsimple}[1]{%
  enhanced,
  breakable,
  colback=white,
  colframe=black!30,
  boxrule=0.45pt,
  sharp corners,
  left=10pt,right=10pt,top=10pt,bottom=10pt,
  title={#1},
  colbacktitle=white,
  coltitle=black,
  fonttitle=\bfseries\small,
  lefttitle=0pt,
  righttitle=0pt,
  toptitle=0pt,
  bottomtitle=7pt,
  titlerule=0.35pt,
  titlerule style={black!18},
  before upper=\small
}

\definecolor{prompttitlebg}{RGB}{46,68,110}
\definecolor{promptframe}{RGB}{46,68,110}
\definecolor{privatebg}{gray}{0.96}

\newtcolorbox{promptbox}[1][]{%
    enhanced,
    breakable,
    colback=white,
    colframe=promptframe,
    boxrule=0.5pt,
    sharp corners,
    left=10pt, right=10pt, top=8pt, bottom=10pt,
    before skip=12pt,
    after skip=12pt,
    title={#1},
    colbacktitle=prompttitlebg,
    coltitle=white,
    fonttitle=\bfseries\sffamily\small,
    lefttitle=10pt,
    righttitle=10pt,
    toptitle=5pt,
    bottomtitle=5pt,
    titlerule=0pt,
    fontupper=\small\linespread{1.08}\selectfont,
}

\newtcolorbox{privatebox}{%
    enhanced,
    breakable,
    colback=privatebg,
    colframe=black!35,
    boxrule=0.3pt,
    sharp corners,
    left=10pt, right=10pt, top=7pt, bottom=7pt,
    before skip=6pt,
    after skip=6pt,
    fontupper=\small\linespread{1.06}\selectfont,
}
\begin{document}
\RUNAUTHOR{Liang and Xu}
\RUNTITLE{Dynamic Bargaining with LLM Agents}
\TITLE{When LLM Agents Negotiate: Private Information and Dynamic Bargaining in Supply Chains}
\ARTICLEAUTHORS{%
\AUTHOR{Chen Liang ~~~ Fasheng Xu\\[-2mm]}
\AFF{School of Business, University of Connecticut} 
} 
\ABSTRACT{%
As LLM agents move from decision support to autonomous procurement, firms need to know whether delegated negotiators create value, divide it predictably, and avoid contracts that lose money for one party. We study these questions in a canonical dynamic supply chain bargaining problem: a buyer with private demand information negotiates a quantity--payment contract with an uninformed seller. We benchmark nine LLMs from OpenAI, Google, and Alibaba against a validated Perfect Bayesian Equilibrium, using 9,840 LLM-to-LLM negotiations. Three findings emerge. First, capability governs value creation. LLM agents reach agreement in 98.9\% of negotiations and capture 95.4\% of first-best surplus in undiscounted terms, but they average 2.98 rounds against the Bayesian benchmark of 1.25, and this delay erodes 21--34\% of first-best surplus, depending on patience. The same capability ordering governs operational reliability: baseline models accept individually irrational contracts in 19.2\% of negotiations, versus 0.0--0.6\% at mid-tier and flagship, making automated profit verification the binding guardrail below the capability threshold. Second, surplus capture is relational: provider identity predicts the direction of surplus flow more reliably than capability rank. Under a common prompting regime, average self-play buyer shares are 40\% for OpenAI, 50\% for Google, and 70\% for Alibaba's Qwen, an ordering that persists when communication is restricted and discounting is removed. In cross-provider matchups, reversing which provider sells shifts the division by 7--18 percentage points---as large as the within-family swing from reversing which capability tier sells (roughly 17 points)---and a provider profile can override capability outright: the highly capable Qwen flagship is the weakest cross-family seller. Vendor choice is therefore a first-order distributional decision. Third, the prompt itself is a strategic lever. Restricting agents to numeric offers shifts surplus in provider-specific directions; removing discounting preserves every qualitative regularity but lengthens bargaining; and delegating to an LLM agent separates the principal's economic patience (the real cost of delay) from the agent's prompted strategic patience, a free deployment choice that is the strongest single driver of surplus division (90\% of explained variance in our design), and the best setting varies by role and model. Together, the results establish an equilibrium-referenced audit of strategic AI agents along three dimensions: discounted efficiency, distributional profile, and operational reliability.
}%
\HISTORY{Current version: July, 2026}
\maketitle
\vspace{-10mm}
\section{Introduction}

The digitization of supply chains is shifting from automated execution to autonomous negotiation. Walmart has deployed Pactum's LLM agents for tail-spend contracts across thousands of suppliers.\footnote{\url{https://pactum.com/procurement-agent/}} Arkestro, Globality, and Fairmarkit manage procurement negotiations for Boeing, BP, and other large buyers.\footnote{See \url{https://www.arkestro.com/}, and \url{https://www.fairmarkit.com/}.} The same shift extends to consumer and cross-border commerce, where Alibaba's Accio equips sourcing agents that negotiate supplier pricing and Xianyu's FishBargain bargains on behalf of individual sellers \citep{kong2025fishbargain}. A controlled demonstration underscores the stakes of full delegation: in Anthropic's \textit{Project Deal}, employees' Claude agents traded real personal items in a marketplace with no human in the inner loop, and more capable agents systematically captured more value, selling higher and buying lower than weaker ones.\footnote{\url{https://www.anthropic.com/features/project-deal}} As both sides of a transaction increasingly delegate to autonomous agents, bargaining is becoming an LLM-to-LLM activity conducted at machine speed and scale: a single recent international competition ran more than $180{,}000$ negotiations between AI agents \citep{vaccaro2026advancing}. This places the economic competence of autonomous LLM negotiators (whether they create value, divide it predictably, and avoid costly errors) at the center of the emerging agentic economy.

For firms deciding which LLM vendor to adopt and which side of a transaction to automate, the operational question is whether LLM agents bargain in ways that are economically coherent, distributionally stable, and reliable enough to be trusted with high-volume procurement. These are decisions about capability thresholds for autonomous deployment, distributional consequences of counterparty choice, and guardrail architectures for acceptable risk. Linguistic or leaderboard evaluations do not directly address such questions: operational value is measured in dollars foregone to delay, dollars misallocated across trading partners, and dollars lost when agents accept individually irrational contracts.

Yet existing LLM evaluations leave open whether autonomous agents bargain coherently in operational contracting settings where performance can be benchmarked against economic theory and evaluated on deployment-relevant outcomes. We therefore ask whether general-purpose LLM agents can recover equilibrium-consistent bargaining behavior in a canonical supply chain contracting problem, and how their behavior varies with the vendor-selection and role-assignment decisions firms must make. We benchmark nine LLMs from three providers (OpenAI, Google, and Alibaba) against the Perfect Bayesian Equilibrium of the alternating-offer bargaining game of \citet{feng2015dynamic}, in which a buyer with private demand information negotiates a quantity--payment contract $(q, T)$ with an uninformed seller.

The experimental program comprises 9,840 LLM-to-LLM negotiations across three main-analysis blocks (symmetric verbal self-play, cross-family flagship pairings, and within-family capability-asymmetric pairings) and seven design and robustness extensions (structured-offer bargaining without verbal communication, no-discounting bargaining, strategic-patience analysis, reasoning-effort ablation, retail-price sensitivity, prior sensitivity, and parameter-size variation). Prompts disclose the primitives of the game (payoff formulas, demand distributions, discount factors, priors, and buyer type) but agents are not told which contract to offer, when to separate types, or how to update beliefs. The design therefore audits strategic execution in a stylized supply chain contracting setting rather than discovery of the setting from raw context.

The design deliberately excludes the domain-specific guardrails that surround production LLM procurement systems: rule-based approval layers, cost-verification databases, escalation protocols, and human sign-off gates.
Although useful for deployment, such controls make it difficult to distinguish strategic competence from compliance with externally imposed constraints. We therefore evaluate models in a common bargaining environment with minimal additional safeguards or prompt engineering, identifying their underlying bargaining behavior and the guardrails needed to support, constrain, or correct it. The evidence delivers three empirical findings tied to the vendor-selection and role-assignment problem.

\textit{Finding 1: Capability is the value-creation lever that governs both efficiency and reliability.} LLM agents reach agreement in 98.9\% of negotiations and capture 95.4\% of first-best surplus in undiscounted terms, but they average 2.98 rounds against the Bayesian benchmark of 1.25. Under discounting---which we treat as a stress test on costly continuation rather than a literal time-preference estimate (\S\ref{sec:patience_interp})---this delay erodes 21--34\% of first-best surplus, depending on patience. Of the efficiency variation the experimental design systematically moves, capability explains roughly 66\% (Shapley decomposition): flagship models take more rounds than baselines (3.25 versus 2.75) but achieve materially higher efficiency (98.9\% versus 91.0\%), consistent with an iterative-search pattern in which capable agents invest in proposal--rejection cycles rather than settling at round-one heuristic offers.

The same capability ordering governs operational reliability: baseline models accept individually irrational contracts (negative profit for one party) in 19.2\% of cases, versus 0.6\% for mid-tier and 0.0\% for flagship models. This order-of-magnitude gap separates a verification-gated weak-model cluster from a lighter-monitoring strong-model cluster, making automated profit verification a necessary guardrail below the threshold.

\textit{Finding 2: Provider identity is the distributional lever, and surplus capture depends on the counterparty and role.}
Capability and provider identity load on different margins: capability governs how much surplus is created (Finding~1), while provider identity is the more reliable predictor of who captures it. Among the model versions evaluated under a common prompt protocol, self-play surplus division varies markedly across providers: Qwen models average 70\% buyer share (within-family range 53--91\% across capability tiers), Gemini 50\% (44--54\%), and OpenAI 40\% (38--42\%). The mean gap between Qwen and OpenAI (approximately 30 percentage points) is comparable to the largest within-provider spread (39 points across Qwen tiers), so the 53.5\% pooled buyer share masks three qualitatively distinct bargaining profiles rather than a single capability-tier effect. Of the surplus-division variation the experimental design systematically moves, the announced patience parameters explain 90\% (Shapley decomposition), while capability, buyer type, and first-proposer assignment together contribute the remaining 10\%. Prompted, common-knowledge patience, however, is a configuration choice rather than a model attribute (Finding~3); among the model-side factors a firm cannot prompt away, provider identity is the strongest predictor of who captures surplus. Public demonstrations such as Anthropic's \textit{Project Deal}, in which upgrading one's own agent reliably improved its terms, reinforce the intuition that a more capable agent is a better bargainer. Our results qualify it: capability reliably \textit{creates} value (Finding~1), but its \textit{distributional} advantage does not survive across providers. 

Even holding provider fixed, a cross-tier GPT pairing makes the point: switching the buyer from GPT-5.2 to the weaker GPT-5-mini against an unchanged GPT-5.2 seller \textit{lowers} buyer share from $37.9\%$ to $35.2\%$, so within a family the more capable model captures the larger share. Across providers, however, capability rank alone does not determine who captures surplus: a highly capable Qwen flagship is the weakest cross-family seller, so a model's provider profile can override its capability. Cross-family direction effects amplify this pattern: when flagship models from different providers negotiate each other, the direction of the pairing produces 7--18 percentage-point swings in surplus, comparable to the within-family capability swing; Gemini captures 67\% of surplus as a cross-family buyer and retains 48\% as seller, while Qwen retains only 27\% as a cross-family seller. For firms deploying different vendors on opposite sides of a procurement interaction, model choice is therefore a first-order distributional decision, not a commodity input.

\textit{Finding 3: The principal's configuration choices are strategic levers.} Because firms bargain through delegated agents rather than directly, the prompt serves as the agent's mandate: three economically meaningful design choices systematically shape bargaining outcomes. Constraining the communication channel reshapes provider biases: removing natural language shifts OpenAI's baseline and mid-tier models \textit{toward} buyers (GPT-5-mini: 41.9\% to 51.3\%) while shifting Gemini's mid-tier and flagship models toward sellers, so the verbal channel is doing strategic work, not decorating offer behavior. Removing the discounting framework from the prompt leaves the value-creation, distributional, and reliability rankings intact but changes negotiation tempo, so discount-factor language operates as a tempo cue rather than a substantive driver of the results. Most distinctively, delegation to LLM agents separates two objects that classical bargaining treats as a single discount-factor parameter: \textit{economic patience} (the principal's fixed real-time cost of delay) and \textit{strategic patience} (the agent's prompted discount factor, which the principal sets at deployment). Patience thus becomes a configurable design lever rather than an exogenous primitive, and choosing it well raises realized payoff by economically meaningful margins. Configuration is therefore a first-order design problem alongside model and vendor selection.

The paper makes three contributions to operations management research on AI-mediated contracting. Empirically, we document a dimension of LLM heterogeneity that standard capability benchmarks miss (provider-level bargaining profiles) and show it is as consequential as within-family capability differences once counterparties are heterogeneous, establishing vendor choice as a strategic operations decision rather than a technical one. Methodologically, we introduce a reproducible equilibrium-referenced audit framework: an executable implementation of the branch-specific PBE, validated cell-by-cell against the equilibrium's analytical properties---a portable template for evaluating autonomous agents in game-theoretic operational settings. Because the audit spans the current capability frontier (nine models, three tiers, and roughly three generations), it separates structural regularities that travel across model vintages from model-specific results that recalibrate as providers update, so the framework remains informative as the underlying models improve. Practically, we identify three deployment-relevant evaluation dimensions (time-adjusted efficiency, provider-level distributional profile, and operational reliability), guiding capability thresholds, role assignment, and guardrail architecture.

Delegation changes both the stakes and the relevant benchmark. When firms authorize LLM agents to negotiate on their behalf, agents' offers, concessions, and acceptance decisions directly determine realized contract terms and surplus. This is particularly consequential in procurement, where automation can extend bargaining to supplier relationships that firms cannot economically negotiate one by one, allowing gains and systematic errors alike to scale across transactions \citep{vanhoek2022walmart}. The central question is therefore not whether LLM agents bargain like humans, but whether they advance their principals' economic interests. We evaluate this competence against the Perfect Bayesian Equilibrium characterized by \citet{feng2015dynamic}. The equilibrium provides reference values for agreement timing, contract form, surplus division, screening, and feasibility. Theory thus serves as measurement infrastructure, allowing us to identify when an agent creates value, transfers value to its counterparty, or destroys value through delay or individually irrational acceptance. Human-subject experiments address complementary questions about human--agent interaction and the transmission of principal heterogeneity into delegated outcomes; for example, \citet{imas2025agentic} show that delegated outcomes retain substantial variation across principals. Our design instead fixes the payoff objective and evaluates the agent against the corresponding equilibrium, providing a direct measure of agent-level bargaining competence.

The remainder of the paper develops these results: Section~\ref{sec:literature} reviews the three research streams we build on; Section~\ref{sec:design} describes the experimental design and Bayesian benchmark; Sections~\ref{sec:performance} and~\ref{sec:provider} present the first two findings on value creation and distribution; Section~\ref{sec:design_choice} develops the third finding on the principal's configuration levers and summarizes the robustness program (detailed in the Appendix); and Section~\ref{sec:conclusion} concludes with practical implications and future directions.

\section{Literature Review}
\label{sec:literature}

Our study contributes to three streams that together define what it means to evaluate an autonomous bargaining agent for operational deployment: dynamic bargaining under asymmetric information, which supplies the normative benchmark; behavioral evidence on bargaining in operations, which establishes the performance baseline human negotiators reach without equilibrium computation; and the emerging machine-behavior literature, which treats AI systems as strategic actors requiring their own audit.

Dynamic bargaining models formalize how parties trade off surplus division against costly delay. The alternating-offers framework \citep{rubinstein1982perfect} remains canonical: equilibrium allocations depend on relative patience, generating sharp comparative statics for who captures surplus and how quickly agreement occurs. Under private information the benchmark shifts to Perfect Bayesian Equilibrium, and strategic delay, signaling, and screening become central \citep{kennan1993, myerson1983}; results related to the Coase conjecture show that when delay is costly, equilibrium often features rapid agreement, with residual inefficiency tied to informational frictions rather than bargaining mechanics \citep{coase1972, gul1986}.\footnote{The Coase prediction is conditional on the interpretation of the discount factor; Section~\ref{sec:patience_interp} discusses how the economic content of $\delta$ differs in LLM-to-LLM bargaining.} In operations, \citet{feng2015dynamic} adapt these ideas to supply chain contracting: a buyer with private demand information and an uninformed seller bargain over a single contract $(q,T)$, and incomplete information generates screening, signaling, and quantity distortion or pooling along the path. Their clean predictions on agreement speed, contract form, and surplus division make the environment well suited for benchmarking automated negotiators.

A parallel stream establishes that human decision makers achieve high performance in bargaining despite bounded rationality \citep{simon1955behavioral}. Controlled experiments show human negotiators often approach theoretical benchmarks on efficiency and agreement while exhibiting systematic patterns in delay, disclosure, and fairness not captured by purely rational models: \citet{davis2021private} find high agreement rates with limited delay under private information, attributing this to experience, institutional norms, and communication, and \citet{davis2022procurement} extend the result to procurement and assembly structures. Communication is central: \citet{camerer2019dynamic} show unstructured text carries signals that predict outcomes, \citet{davis2025bargaining} that voluntary disclosure overcomes matching frictions, and \citet{haruvy2020bargaining} that protocol design shifts efficiency and surplus division at fixed primitives, with fairness and contract framing also shaping outcomes \citep{katok2013}. This literature sets a high behavioral baseline that humans reach through calculation, norms, and communication rather than strict backward induction, motivating the question of whether LLM agents replicate it, and whether they do so through similar or qualitatively different mechanisms.

As LLM agents enter procurement and contracting workflows, a growing literature studies AI systems as behavioral objects whose actions can deviate systematically from equilibrium and from human behavior \citep{rahwan2019, manning2026general}. This literature models LLMs as economic actors whose strategic conduct depends on prompting and system design rather than equilibrium reasoning alone \citep{filippas2024large, wang2025sponsored, shahidi2025coasean, hadfield2025economy}.
In static settings, LLMs replicate human decision biases, often more strongly than human managers \citep{chen2025manager}: \citet{liu2025large} document pull-to-center and demand-chasing in inventory tasks, alongside a ``paradox of intelligence'' in which more capable models overthink into greater irrationality. Role and relationship framing in the prompt shifts how LLMs divide surplus between supply chain parties \citep{liu2025make}, and in contract design under moral hazard, LLM principals favor enforceable incentive contracts over the trust-based bonus contracts humans often choose \citep{kirshner2025ai}.

Evidence on dynamic negotiation is more mixed. \citet{kirshner2026talking} find LLM agents more agreement-oriented than humans, raising efficiency but potentially inequality; \citet{zhu2025automated} warn of behavioral anomalies and losses in consumer settings; and \citet{hasija2025anchors} identify a role-based asymmetry in which buyer-role agents outperform supplier-role agents through anchoring. Closest to our work, \citet{chen2026haggling} study human-retailer and LLM-supplier bargaining in a two-tier chain and show LLM suppliers reproduce many human--human patterns but need externally specified reservation-profit guidance for stability. Our study is complementary on three dimensions: LLM-to-LLM rather than human--LLM negotiation, isolating agent behavior from human adaptation; the \citet{feng2015dynamic} asymmetric-information setting, where competence demands signaling, screening, and belief updating; and benchmarking against a validated PBE rather than human data.

Two further studies vary other sources of heterogeneity in delegated negotiation: \citet{imas2025agentic} hold the model fixed and vary the human principals who write the prompts, showing delegated outcomes inherit and amplify principal heterogeneity, while \citet{vaccaro2026advancing} run a large open LLM-to-LLM competition ($182{,}812$ negotiations, $286$ prompt engineers) that flags prompt strategy as a central driver. We add the provider layer under a common baseline prompt and separately test whether principal-controlled configuration choices shift outcomes relative to the equilibrium benchmark. The three designs thus locate heterogeneity at different layers of the delegation stack (principal, prompt, provider), and our equilibrium-referenced audit supplies the provider-level component of the integrated theory of AI negotiation that \citet{vaccaro2026advancing} call for.

Three gaps remain for operations management research. First, equilibrium-grounded audits of LLM agents in canonical dynamic bargaining under asymmetric information are absent, even though success in such settings requires both operational calculation and strategic inference; Finding~1 and its decomposition address this. Second, existing evaluations lack a validated theoretical reference, so deviations cannot be cleanly attributed to the environment, the evaluator, or the agent; our executable Bayesian benchmark (Section~\ref{sec:design}) removes the ambiguity. Third, general-purpose LLM evaluations such as leaderboards and linguistic benchmarks assess competence but poorly predict operational value; Findings~2 and~3 and the three-dimensional framework of Section~\ref{sec:conclusion} close this gap.

More broadly, a management literature documents how generative AI reshapes task allocation, productivity, and organizational design \citep{ide2025artificial, xu2025generative, eloundou2023gpts, brynjolfsson2025generative} and how AI supply chains shape deployment \citep{xu2024economics, fransoo2025navigating}. We respond to calls to engage the distinctive features of generative AI rather than treat it as a faster, cheaper technology \citep{simchi2025democratizing, cohen2025supply, dai2025assured}, focusing on a narrow but consequential unit: bilateral contracting under demand uncertainty, where the value question admits both a normative benchmark and a concrete deployment decision.

\section{Experimental Design}
\label{sec:design}

The experimental program is built around the paper's central question: when a firm delegates bargaining to an LLM agent, which of its choices move outcomes? It is therefore organized to separate three delegation layers---capability tier, provider identity, and prompt configuration---and to benchmark outcomes in each layer against a validated equilibrium that provides a known normative reference for timing, contract form, and surplus division. The core block audits bargaining competence with a $2 \times 2 \times 4$ factorial design (Table~\ref{tab:experimental_design}) crossing buyer type, first proposer, and patience configuration. The three factors test, respectively, information revelation under high versus low demand, protocol effects on screening versus signaling, and the patience regimes identified by \citet{feng2015dynamic} where rational behavior transitions from truth-telling to pooling. Each condition is replicated 15 times per model across the nine models of Table~\ref{tab:language-models}, yielding 2,160 negotiations in the baseline block.

Our prompts fully disclose the primitives of the game but not the equilibrium strategy. Agents can therefore compute any function of the disclosed parameters in principle, but they are not given the branch-specific screening or signaling logic from \citet{feng2015dynamic}. The design tests whether LLMs can translate disclosed strategic structure into approximately coherent bargaining actions through interaction rather than whether they can infer the environment from scratch. Because the prompts are role-specific, the reported surplus shares should be interpreted as behavior under a common prompting regime rather than as prompt-free estimates of intrinsic bargaining bias.

\begin{table}[htbp]
\centering
\caption{Main Experimental Design and Sample Distribution}
\label{tab:experimental_design}
\footnotesize
\renewcommand{\arraystretch}{1.1}
\begin{tabular}{ll}
\toprule
\textbf{Design Component} & \textbf{Value} \\
\midrule
Buyer type & 2 levels: High ($\mu=80, \sigma=10$), Low ($\mu=40, \sigma=10$) \\
First proposer & 2 levels: Buyer, Seller \\
Patience configuration & 4 levels: $(0.9,0.9)$, $(0.7,0.9)$, $(0.4,0.9)$, $(0.9,0.4)$ \\
Replications per condition & 15 \\
Models & 9 \\
\bottomrule
\end{tabular}
\par\smallskip
\begin{minipage}{\textwidth}
\scriptsize
\textit{Notes:} Patience configurations are ordered as $(\delta_B,\delta_S)$, buyer patience first and seller patience second.
\end{minipage}
\end{table}

\subsection{Model Selection}

To support both within-provider capability comparisons and cross-provider heterogeneity comparisons, we selected nine models from three commercial providers. The three families come from independent organizations: OpenAI and Google are U.S.-based, closed-source frontier-model providers, while Alibaba is a China-based frontier-model provider. Distinct training data, alignment objectives, and post-training procedures give the sample cross-provider heterogeneity rather than within-family recipe similarity. Each family offers at least three publicly accessible capability tiers (flagship, mid-tier, baseline), shown in Table~\ref{tab:language-models}. Within each family the baseline tier is the prior-generation model: GPT-4o-mini and Qwen2.5-14B are non-reasoning models, whereas Gemini-2.5-Flash (like all six mid-tier and flagship models) exposes reasoning (``thinking'') capability, a distinction that proves consequential for operational reliability (Section~\ref{subsec:reliability_threshold}). Extensions to Anthropic (Claude), Meta (Llama), and xAI (Grok) are valuable directions for future work.

\begin{table}[htbp]
\centering
\caption{Language Models by Provider and Capability Tier}
\label{tab:language-models}
\footnotesize
\renewcommand{\arraystretch}{1.1}
\begin{tabular}{llll}
\toprule
\textbf{Provider} & \textbf{Flagship} & \textbf{Mid-Tier} & \textbf{Baseline} \\
\midrule
OpenAI & GPT-5.2 & GPT-5-mini & GPT-4o-mini \\
Google & Gemini-3-Pro~~ & Gemini-3-Flash~~ & Gemini-2.5-Flash~~ \\
Alibaba & Qwen3-Max & Qwen3-32B & Qwen2.5-14B \\
\bottomrule
\end{tabular}
\end{table}

\subsection{Negotiation Protocol}

Our baseline negotiation protocol implements the dynamic bargaining framework of \citet{feng2015dynamic} adapted for LLM agents; variations used in the design and robustness extensions (R1 structured communication, R2 no-discounting bargaining) are described where introduced. A seller with production cost $c = 30$ negotiates with a buyer over a wholesale contract $(q, T)$ specifying order quantity and transfer payment. The buyer privately observes its demand type; the seller holds an equal prior on each type. Both parties know the retail price $r = 60$, production cost, discount factors, and the ten-round maximum. Parties alternate offers until agreement or the round limit; failure to agree yields zero payoffs. The theoretical benchmark predicts rational agents reach agreement within two rounds through screening, signaling, and quantity distortion when relevant; this is the performance target against which we evaluate LLM agents.

To mimic realistic supply chain negotiation environments, the protocol allows agents to exchange natural-language messages alongside formal offers. A negotiation terminates when the responder signals acceptance in the structured action field or the ten-round cap is reached; unaccepted negotiations hitting the cap are counted as zero-surplus outcomes for agreement and efficiency metrics, but excluded from analyses requiring accepted contract terms, profits, or surplus shares. Conditional on acceptance, final contract terms are resolved in two stages. First, a language-model extractor (GPT-4o-mini) recovers the deal from the full conversation history, subject to numerical validation (quantity and payment non-null, strictly positive, and finite). Second, if the extractor returns no validated values, a deterministic rule-based scan recovers the most recent structured proposal satisfying the same criteria.\footnote{We re-extracted terms on a stratified sample of 300 experiments (all 2 fallback-recovered deals, all 24 round-limit non-agreements, and 274 randomly drawn extractor-resolved deals balanced across models) using a second LLM from a different family (Claude Sonnet 4.6). The two extractors agreed on deal status in 100\% of cases and on quantity within $\pm 0.5$ units in all 276 jointly confirmed deals; payment agreed in all but one, a per-unit-versus-total ambiguity in a gpt-4o-mini contract that quoted a per-unit price in the payment field. A corpus-wide scan found four such gpt-4o-mini cases, which we rescaled to the implied total; the adjustment leaves efficiency unchanged and shifts aggregate surplus shares by at most 0.2 percentage points.}

For agreements reached in round $\tau$, we compute buyer profit $\pi_B = r \cdot \mathbb{E}[\min(D, q)] - T$ using newsvendor expected revenue, and seller profit $\pi_S = T - cq$. Effective utilities incorporate time discounting as $U_B = \delta_B^{\tau-1} \pi_B$ and $U_S = \delta_S^{\tau-1} \pi_S$, consistent with the \citet{rubinstein1982perfect} bargaining framework (we discuss the role and interpretation of $\delta$ in Section~\ref{sec:patience_interp}). Both agents are instructed to reject individually irrational offers yielding negative profit. Appendix~\ref{subsec:illustrative_negotiation} provides a representative negotiation transcript.

Our design prioritizes equilibrium-referenced identification over realism. Five choices support this priority. First, we fix temperature at 1.0 (the API default and a common production setting) to characterize out-of-the-box stochastic behavior. Second, we set outside options to zero, following \citet{rubinstein1982perfect}, to isolate pure bargaining from exit-threat credibility. Third, we use role-specific but structurally symmetric buyer and seller prompts, so observed heterogeneity reflects model capability and role rather than prompt engineering (full prompts in Appendix~\ref{app:prompts}). Fourth, we disclose the game's primitives without supplying reservation profits or target utilities, which would substitute designer-imposed discipline for the endogenous execution we measure. Fifth, we study LLM-to-LLM rather than human--LLM bargaining so that outcome differences attribute to model behavior rather than human adaptation or interface effects. Natural-language messages are permitted as a native LLM channel while formal offer fields preserve machine-checkable contracts (Extension~R1 shows the regularities persist under structured-only offers), complementing concurrent human--LLM studies \citep{chen2026haggling} that prioritize realism over equilibrium referencing.

\subsection{What Patience Means in the Experiment}
\label{sec:patience_interp}

\textit{Three roles for patience.} Patience enters the paper in three distinct roles. First, it is a \textit{theoretical} object: the \citet{feng2015dynamic} dynamic-bargaining benchmark requires discount factors to pin down continuation values, agreement timing, and surplus division; without discounting, the distributional prediction is not uniquely defined. Second, it is an \textit{experimental} object: in the main treatments each agent is told both parties' discount factors, matching the common-knowledge assumption of the benchmark and removing any need to infer hidden patience. Third, it is a \textit{deployment} object: a real principal may face an economic cost of delay that is not literally encoded as a prompt parameter. Our main analysis uses prompted patience as measurement infrastructure; Extension~R2 then removes that infrastructure to test whether the qualitative findings survive in a no-discounting regime.

\textit{What the agents know, and what the prompted value represents.} The patience pair $(\delta_B, \delta_S)$ is common knowledge, disclosed to both agents alongside the retail price, production cost, and round limit (Appendix~\ref{app:prompts}), while the buyer's demand type is the sole private information. The experiment therefore asks whether LLMs can \textit{use} known primitives, not whether they can \textit{infer} a hidden one; uncertainty over a counterparty's patience is a distinct problem we leave to future work. The prompted discount factor is an announced utility parameter specified in the prompt; it is neither elicited from the human principal nor inferred by the agent.

\textit{Why prompted patience is also a design lever.} The gap between the experimental and deployment roles is not a defect but the source of a managerial design lever. In classical bargaining a single $\delta$ does double duty (it is at once the agent's weight on delay and the principal's real cost of waiting) because a bargaining round and a real-time period coincide. Under LLM mediation they decouple: a round of LLM exchange takes seconds, whereas the deployed economic period is governed by API latency, human review, and approval workflows. We therefore distinguish \textit{strategic patience} $\delta^{\text{strat}}$, the discount factor the agent applies across alternating-offer rounds and the value we set by prompt, from \textit{economic patience} $\delta^{\text{econ}}$, the principal's fixed exogenous discount over real time. Because the prompted value is $\delta^{\text{strat}}$ and need not equal $\delta^{\text{econ}}$, the principal's problem is not to translate a human's patience into a prompt but to \textit{tune} $\delta^{\text{strat}}$ against a fixed economic constraint, a principal-specification exercise we develop in Section~\ref{subsec:strategic-patience}.

\textit{Reading the efficiency metrics.} Consistent with this view we report both \textit{undiscounted} efficiency, appropriate when the object of interest is allocative quality independent of process, and \textit{discounted} efficiency, which measures sensitivity to continuation cost under the prompted $\delta^{\text{strat}}$. Discounted efficiency should be read as a stress test of how outcomes respond when delay is made costly, not as a direct estimate of either human time preference or operational reliability.\footnote{One can additionally motivate the \textit{magnitude} of $\delta$ through a per-round breakdown hazard, writing $\delta_k = e^{-(\alpha_k + \lambda_k)\Delta t}$ with $\alpha_k$ a time-preference rate and $\lambda_k$ a breakdown hazard over inter-round interval $\Delta t$ \citep{feng2015dynamic}; at the per-second round pace of our experiments a hazard-based analogy is more natural than literal time preference. We do not identify these channels in our data and do not rely on this reading.} When we study prompt design in Section~\ref{subsec:strategic-patience}, we separately evaluate realized payoffs at the principal's economic patience $\delta^{\text{econ}}$.

\subsection{Bayesian Benchmark Implementation and Validation}

Throughout the paper, we use two related theoretical references. The complete-information Rubinstein outcome provides a first-best bargaining reference: quantity is set at the type-specific first-best level and surplus is divided according to alternating-offer patience. The primary asymmetric-information benchmark is the branch-specific equilibrium characterization of \citet{feng2015dynamic}. Because their formal incomplete-information PBE is derived for seller-initiated bargaining, our seller-first cells implement that benchmark directly, while buyer-first cells use a separately derived buyer-first separating equilibrium built from the same primitives. Appendix~\ref{Bayesian-Details} provides the derivation and implementation details.

Each cell's equilibrium is determined by its patience pair $(\delta_B, \delta_S)$ and prior $\beta$. In the main 16-condition design with $\beta = 0.5$, all buyer-first cells settle in round~1 (separating) and all seller-first cells settle by round~2 (the H-type accepts the seller's round-1 offer; the L-type counteroffers in round~2). The benchmark agreement time averages 1.25 rounds (1.00 buyer-first, 1.50 seller-first).

\subsection{Experimental Program Overview}

The paper comprises three main-analysis blocks and seven design and robustness extensions, summarized in Table~\ref{tab:program_overview}. The main analysis is organized around the three delegation layers. M1 (the 16-condition symmetric self-play factorial described above) and M3 (GPT-5.2 paired with GPT-5-mini, Section~\ref{subsec:within_family_cross_model}) vary the \textit{capability} layer; M2 pairs flagship models from different providers in both buyer and seller roles, varying the \textit{provider} layer and enabling tests of cross-family direction effects (Section~\ref{sec:provider}); the configuration extensions then vary the \textit{prompt} layer (Section~\ref{sec:design_choice}).

The design and robustness extensions comprise three configuration analyses reported in Section~\ref{sec:design_choice}---communication mode (R1), removal of the discounting framework from the prompt (R2), and strategic patience (R3)---plus four additional robustness extensions: reasoning effort (R4), surplus magnitude (R5), prior beliefs (R6), and model size (R7). R2 doubles as an external-validity check: the core efficiency, delay, and reliability regularities survive and the provider ordering persists when discounting is removed from the prompt, indicating that prompted patience is a design feature for the equilibrium audit rather than a precondition for the findings. The completed program covers 9{,}840 unique LLM-to-LLM negotiations and 31{,}592 bargaining rounds.

\begin{table}[htbp]
\centering
\caption{Experimental Program Overview}
\label{tab:program_overview}
\footnotesize
\renewcommand{\arraystretch}{1.1}
\begin{tabular}{llr}
\toprule
\textbf{Block} & \textbf{Description} & \textbf{$N$} \\
\midrule
\multicolumn{3}{l}{\textit{Main Analysis}} \\
M1 & Symmetric LLM (9 models $\times$ 16 conditions $\times$ 15) & 2{,}160  \\
M2 & Cross-family flagship pairings (GPT-5.2, Gemini-3-Pro, Qwen3-Max) & 1{,}440 \\
M3 & Within-family cross-capability (GPT-5.2 $\leftrightarrow$ GPT-5-mini) & 480 \\
\midrule
\multicolumn{3}{l}{\textit{Design and Robustness Extensions}} \\
R1 & Structured communication (no natural language) & 2{,}160\\
R2 & No-discounting bargaining & 540 \\
R3 & Strategic-patience analysis (additional patience configuration) & 540 \\
R4 & Reasoning-effort ablation (GPT-5.2, low/medium/high) & 480  \\
R5 & Retail-price robustness (OpenAI models, $r=120$) & 720  \\
R6 & Seller-prior sensitivity ($\beta \in \{0.3, 0.5, 0.7\}$) & 1{,}080 \\
R7 & Parameter-size ablation (Qwen3 family) & 240  \\
\midrule
\textbf{Total} & & \textbf{9{,}840} \\
\bottomrule
\end{tabular}
\par\smallskip
\begin{minipage}{\textwidth}
\scriptsize
\textit{Notes:} $N$ is the number of unique LLM-to-LLM negotiations per block. These counts are of newly run negotiations; several blocks additionally reuse M1 cells as comparison baselines (e.g., M3 the two homogeneous configurations, R2 the symmetric-patience $(0.9,0.9)$ cells, and R4 the medium-effort arm), so the analysis samples reported later exceed the counts shown here. M1--M3 constitute the main analysis; R1--R7 are the design and robustness extensions. Factor-by-factor breakdowns, round counts, and replication rates appear in Table~\ref{tab:inventory} of Appendix~\ref{sec:extension_program}.
\end{minipage}
\end{table}

\section{Negotiation Performance of LLM Agents}
\label{sec:performance}

The balanced factorial and fixed-prompt LLM-to-LLM design of Section~\ref{sec:design}, audited against the validated Bayesian benchmark, lets us isolate how much of an agent's bargaining outcome is created by the model itself. For a firm deciding whether to delegate a class of procurement contracts to an autonomous agent, the first-order question is how much value the agent captures relative to what a theoretically rational counterparty would capture. We evaluate LLM performance along four outcomes tied to that decision: agreement rate (does the deal close?), negotiation duration in rounds (how costly is the delay?), undiscounted efficiency (what share of first-best surplus is realized?), and discounted efficiency (what share survives once delay is priced?). The Perfect Bayesian Equilibrium of \citet{feng2015dynamic} predicts rational agents reach agreement within two rounds whenever gains from trade are positive, providing the normative benchmark against which we audit behavior. Later sections use Rubinstein only as a complete-information reference for how patience would divide first-best surplus if buyer type were known. Table~\ref{tab:main_results} summarizes LLM performance relative to this Bayesian benchmark across key dimensions. This section establishes Finding~1: capability is the value-creation lever, governing both how much surplus agents realize (competence, \S\ref{subsec:capability_value}) and how reliably they avoid individually irrational contracts (reliability, \S\ref{subsec:reliability_threshold}).

\begin{table}[htbp]
\centering
\caption{Negotiation Performance: LLM Agents vs Bayesian Benchmark}
\label{tab:main_results}
\scriptsize
\renewcommand{\arraystretch}{1.1}
\setlength{\tabcolsep}{3pt}
\resizebox{\textwidth}{!}{%
\begin{tabular}{@{}lcccc|cccc@{}}
\toprule
\textbf{Condition} & \multicolumn{4}{c}{\textbf{LLM Agents}} & \multicolumn{4}{c}{\textbf{Bayesian}} \\
& \textbf{Agreement} & \textbf{Rounds} & \textbf{Efficiency} & \textbf{DiscEfficiency} & \textbf{Agreement} & \textbf{Rounds} & \textbf{Efficiency} & \textbf{DiscEfficiency} \\
& (\%) & & (\%) & (\%) & (\%) & & (\%) & (\%) \\
\midrule
\textbf{Overall} & \textbf{98.9} & \textbf{2.98} & \textbf{95.4} & \textbf{66.2} & \textbf{100.0} & \textbf{1.25} & \textbf{100.0} & \textbf{96.4} \\
\midrule
\textit{By Model Tier:} & & & & & & & & \\
\quad Flagship & 100.0 & 3.25 & 98.9 & 65.3 & 100.0 & 1.25 & 100.0 & 96.4 \\
\quad Mid-tier & 99.9 & 2.93 & 96.4 & 68.0 & 100.0 & 1.25 & 100.0 & 96.4 \\
\quad Baseline & 96.8 & 2.75 & 91.0 & 65.4 & 100.0 & 1.25 & 100.0 & 96.4 \\
\midrule
\textit{By First Proposer:} & & & & & & & & \\
\quad Buyer First & 98.4 & 2.97 & 94.8 & 66.7 & 100.0 & 1.00 & 100.0 & 100.0 \\
\quad Seller First & 99.4 & 2.99 & 96.1 & 65.8 & 100.0 & 1.50 & 100.0 & 92.9 \\
\midrule
\textit{By Buyer Type:} & & & & & & & & \\
\quad High-type & 99.5 & 2.88 & 95.1 & 67.0 & 100.0 & 1.00 & 100.0 & 100.0 \\
\quad Low-type & 98.2 & 3.08 & 95.7 & 65.4 & 100.0 & 1.51 & 99.9 & 92.8 \\
\midrule
\textit{By Patience:} & & & & & & & & \\
\quad (0.9, 0.9) & 98.0 & 3.39 & 95.2 & 74.7 & 100.0 & 1.25 & 100.0 & 97.5 \\
\quad (0.7, 0.9) & 99.4 & 3.09 & 96.1 & 63.8 & 100.0 & 1.25 & 100.0 & 96.1 \\
\quad (0.4, 0.9) & 99.3 & 2.63 & 95.0 & 61.2 & 100.0 & 1.25 & 99.8 & 95.4 \\
\quad (0.9, 0.4) & 98.9 & 2.82 & 95.5 & 65.3 & 100.0 & 1.25 & 100.0 & 96.7 \\
\midrule
\multicolumn{9}{l}{\textit{Statistical Tests (LLM Agents)}} \\
\addlinespace[0.2em]
\quad Model Tier & 42.72*** & 21.11*** & 75.39*** & 3.04* & --- & --- & --- & --- \\
\quad First Proposer & 3.41$^\dagger$ & 0.12 & 5.76* & 0.85 & --- & --- & --- & --- \\
\quad Buyer Type & 7.12** & 10.75** & 1.24 & 2.44 & --- & --- & --- & --- \\
\quad Patience & 6.40$^\dagger$ & 27.95*** & 0.86 & 34.67*** & --- & --- & --- & --- \\
\bottomrule
\end{tabular}%
}
\par\smallskip
\begin{minipage}{\textwidth}
\scriptsize
\textit{Notes:} We use DiscEfficiency to denote discounted efficiency. Patience rows are ordered as $(\delta_B,\delta_S)$, buyer patience first and seller patience second. Test statistics are $\chi^2$ for Agreement and $F$ for Rounds, Efficiency, and Discounted Efficiency. $^\dagger p < 0.10$, * $p < 0.05$, ** $p < 0.01$, *** $p < 0.001$. Bayesian benchmark from validated implementation. Efficiency measured as percentage of first-best undiscounted surplus; Discounted Efficiency applies per-round discount factors to the realized surplus.
\end{minipage}
\end{table}

\subsection{Capability Governs Value Creation}
\label{subsec:capability_value}

\begin{figure}[ht]
\centering
\begin{minipage}[t]{0.48\textwidth}
\centering
\includegraphics[width=\textwidth]{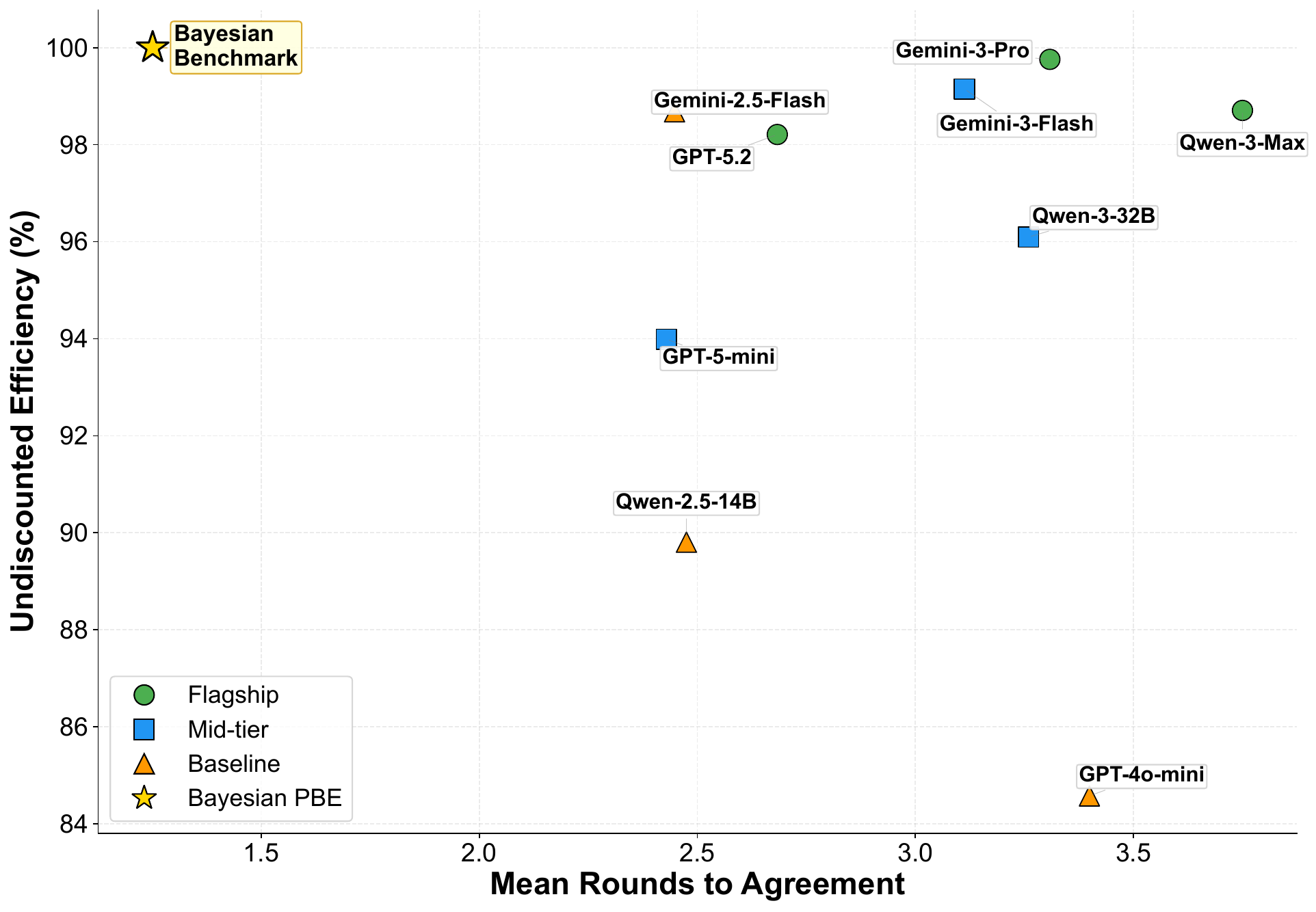}\\
{\small (a) Undiscounted efficiency}
\end{minipage}
\begin{minipage}[t]{0.48\textwidth}
\centering
\includegraphics[width=\textwidth]{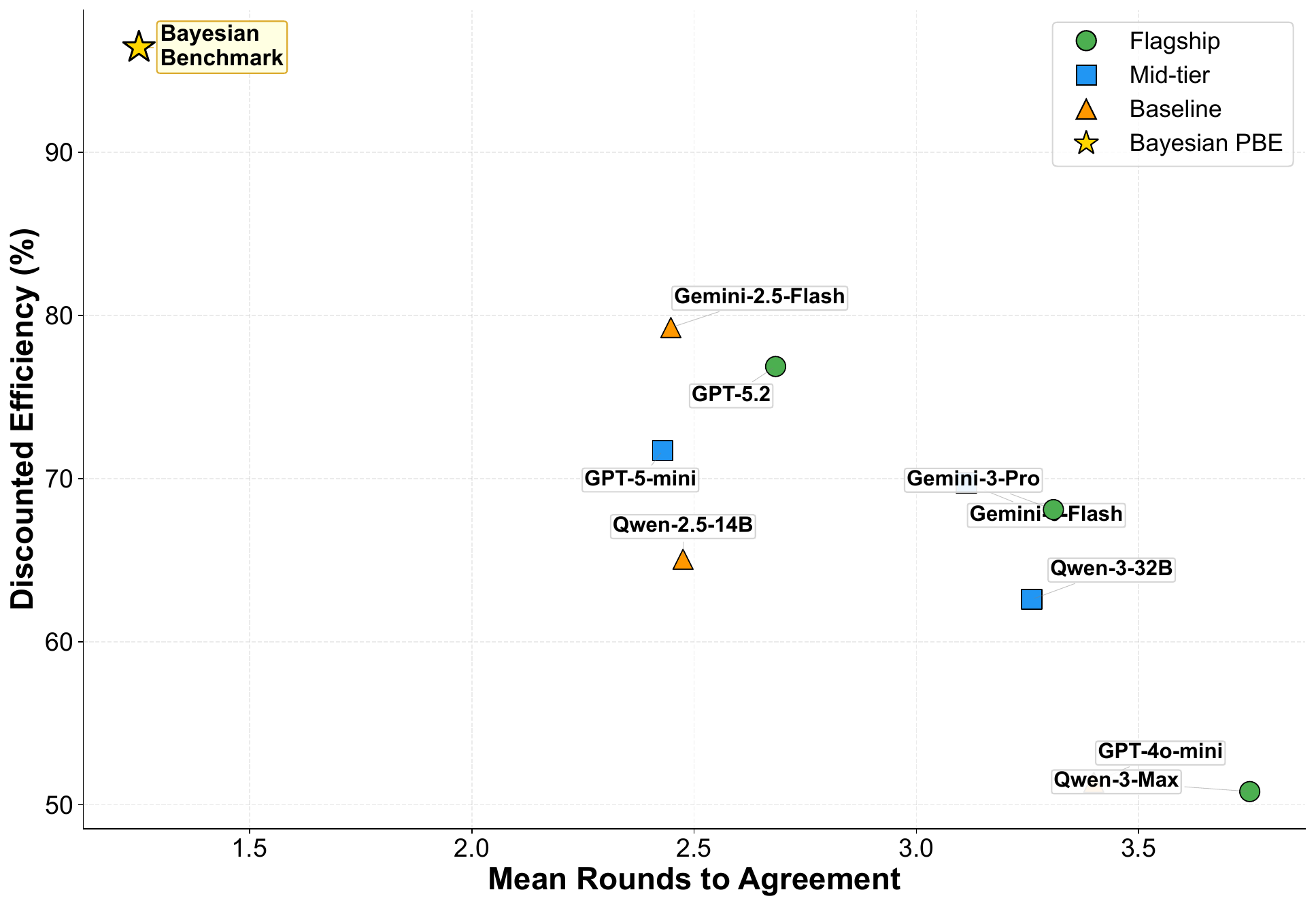}\\
{\small (b) Discounted efficiency}
\end{minipage}
\caption{Efficiency and Rounds by Model Tier}
\par\smallskip
\begin{minipage}{\textwidth}
\scriptsize
\textit{Notes:} (a) Undiscounted and (b) discounted efficiency against mean rounds to agreement, each averaged over the four patience configurations and all conditions (consistent with the discounted-efficiency column of Table~\ref{tab:main_results}). The Bayesian benchmark (star) settles at $1.25$ rounds, and every model lies below and to the right of it. Discounting (b) penalizes additional rounds, sharply widening the spread relative to (a).
\end{minipage}
\label{fig:efficiency_rounds}
\end{figure}

As a class, LLM agents reach near-universal agreement ($98.9\%$) and capture $95.4\%$ of first-best surplus in undiscounted terms, but take $2.98$ rounds against the $1.25$-round Bayesian benchmark, realizing efficient allocations through iterative proposal--rejection search rather than one-shot equilibrium play. The first-order result, however, is not this class-level gap but its dependence on capability. Efficiency rises monotonically with model tier (flagship $98.9\%$, mid-tier $96.4\%$, baseline $91.0\%$), and capability accounts for roughly $66\%$ of the efficiency variation the design moves, far more than any other factor (the variance decomposition in \S\ref{subsec:drivers}). Duration, however, rises with capability rather than falling: baseline models are fastest ($2.75$ rounds), mid-tier models take $2.93$, and flagship models take the longest ($3.25$). The extra rounds are therefore not a uniform price paid for efficiency but a signature of how capable agents search: they invest in proposal--rejection cycles rather than settling at round-one heuristic offers. Figure~\ref{fig:efficiency_rounds} illustrates this across all nine models.

In undiscounted terms (Figure~\ref{fig:efficiency_rounds}a), the nine models do not separate cleanly into tier-based clusters; capability ordering predicts the rough region of the plot but not the within-tier ranking. The flagship tier occupies the high-efficiency region nearest the benchmark's $100\%$: Gemini-3-Pro ($3.31$ rounds, $99.8\%$) is the most efficient model in our sample, GPT-5.2 ($2.68$, $98.2\%$) is the fastest flagship, and Qwen3-Max reaches comparable efficiency ($98.7\%$) at the cost of the most rounds in the sample ($3.75$). The mid-tier and baseline bands overlap the flagship efficiency region rather than sitting visibly below it: Gemini-3-Flash ($3.11$, $99.1\%$) and the baseline Gemini-2.5-Flash ($2.45$, $98.7\%$) match or exceed the efficiency of flagships from other providers. The baseline tier is where the dispersion appears. GPT-4o-mini reaches only $84.6\%$ efficiency at $3.40$ rounds, and Qwen2.5-14B recovers $89.8\%$ at $2.48$ rounds. Tier therefore predicts whether a model approaches the benchmark, but provider identity governs where in that region it lands; every model sacrifices some combination of speed and surplus relative to the theoretical optimum.

Under discounting, this delay is costly, and the cost scales with patience. In discounted terms, LLM agents forgo $33.8\%$ of first-best surplus overall, ranging from $25.3\%$ under symmetric high patience ($\delta_B=\delta_S=0.9$) to $38.8\%$ under the most buyer-impatient configuration ($\delta_B=0.4,\ \delta_S=0.9$), gaps computed from the discounted-efficiency columns of Table~\ref{tab:main_results}.\footnote{Isolating the delay channel alone (holding the allocation fixed and discounting only the excess rounds, $1-\delta^{\tau-1.25}$) gives a comparable $\approx 20\%$ at $\delta=0.9$ and $\approx 48\%$ at $\delta=0.7$.} The timing cost is therefore economically consequential whenever continuation costs are nontrivial, even as undiscounted coordination remains high. Figure~\ref{fig:efficiency_rounds}(b) makes the mechanism visual: once delay is priced, discounted efficiency slopes down steeply in rounds and the tier ordering scrambles. The fast, efficient models hold the top discounted-efficiency positions: the baseline Gemini-2.5-Flash ($2.45$ rounds) and the flagship GPT-5.2 ($2.68$ rounds) reach $79.3$ and $76.9\%$, while the slowest flagship, Qwen3-Max ($3.75$ rounds), falls to the bottom of the discounted ranking ($50.8\%$), below every baseline model despite near-top undiscounted efficiency. Speed, not capability alone, governs the discounted outcome, foreshadowing the provider- and patience-driven division analyzed in Section~\ref{sec:provider}.

The reasoning-effort ablation (Appendix~\ref{subsec:reasoning_effort}; summarized in \S\ref{sec:robustness}) shows that inference-time compute shortens negotiations and raises discounted efficiency while leaving undiscounted efficiency and buyer share statistically unchanged.

The total efficiency gap of 4.56 percentage points decomposes into three components (Appendix~\ref{subsec:B_efficiency_decomp}, Table~\ref{tab:efficiency_decomp}). Suboptimal contract terms in otherwise rational agreements account for the majority of the gap ($63.4\%$, 2.89 pp). Failed deals account for $24.4\%$ (1.11 pp), and irrational deals---agreements in which at least one party accepts terms yielding negative expected profit---account for the remaining $12.2\%$ (0.56 pp). This distribution indicates LLM agents successfully avoid catastrophic failures but still struggle with marginal optimization, often settling for good-enough outcomes rather than refining toward the benchmark optimum. Detailed efficiency heterogeneity, including performance by buyer type and task difficulty variation, appears in Appendix~\ref{subsec:B_efficiency_heterogeneity}.

\subsection{The Capability Threshold for Operational Reliability}
\label{subsec:reliability_threshold}

The same capability ordering that governs efficiency governs the second margin of value creation: operational reliability. We define an irrational agreement as one in which at least one party accepts terms yielding negative expected profit, and report its incidence by model and party in Figure~\ref{fig:irrationality_by_model_detailed}, with the tier-level summary in Appendix~\ref{subsec:B_irrationality} (Table~\ref{tab:irrationality}). Reliability is not a smooth gradient but a threshold: a near-zero failure rate at the upper tiers set against an order-of-magnitude cliff at baseline.

\begin{figure}[htbp]
\centering
\includegraphics[width=0.78\textwidth]{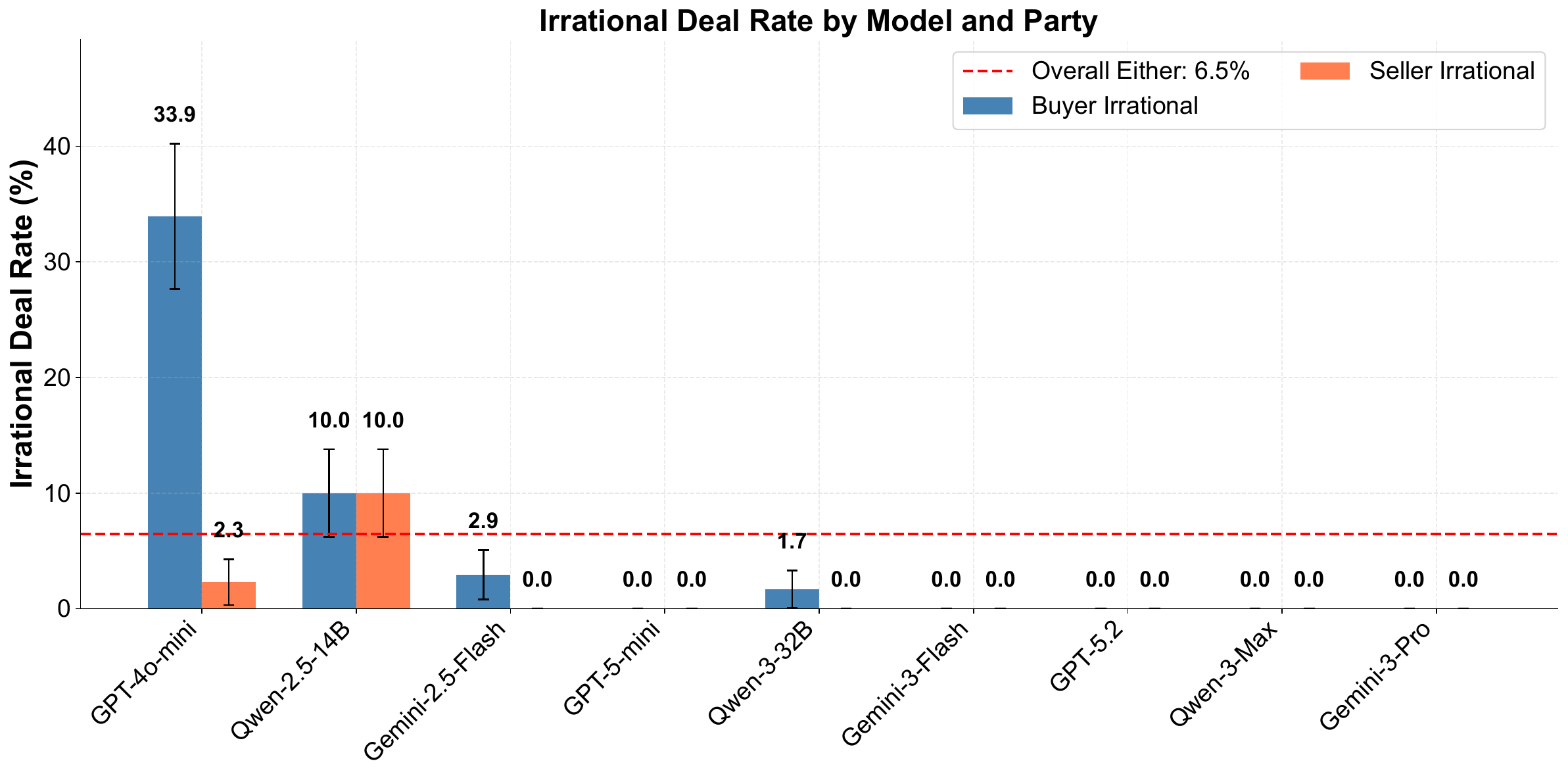}
\vspace{-8pt}\caption{Irrationality Concentrates in Non-Reasoning Baseline Models}
\par\smallskip
\begin{minipage}{\textwidth}
\scriptsize
\textit{Notes:} Bars show buyer- and seller-side irrational-deal rates by model, with error bars; the dashed line marks the overall either-party rate ($6.5\%$). Unsafe contracts concentrate in the two non-reasoning models (GPT-4o-mini and Qwen2.5-14B), whereas the seven reasoning models (including the baseline-tier Gemini-2.5-Flash) are at or near zero.
\end{minipage}
\label{fig:irrationality_by_model_detailed}
\end{figure}

The irrationality measure should be interpreted as a failure to implement an explicitly stated acceptance constraint, not as a pure measure of economic reasoning in isolation. Because the prompt instructs agents not to accept negative-profit agreements, such outcomes may reflect payoff miscalculation, failed constraint checking, or noncompliance with the acceptance rule. Although these channels differ diagnostically, they are operationally equivalent: each leads the delegated agent to accept a contract the principal instructed it to reject.

Irrationality rates vary sharply with capability: $0.0\%$ for flagship, $0.6\%$ for mid-tier, and $19.2\%$ for baselines; $97.1\%$ of all irrational agreements involve a baseline model. Economically unsafe contracts are overwhelmingly a baseline-tier phenomenon, pointing to a capability threshold for autonomous prompt-only deployment. Upper-tier failure rates are within rounding distance of zero, while baseline agents fail at a rate untenable in any high-stakes setting. Prompt-only safeguards therefore suffice at the upper tiers, but baseline deployment requires a more capable model or external verification of the acceptance constraint.

This baseline rate is itself concentrated along a sharp line: whether the model is a reasoning model. The two \textit{non-reasoning} baselines account for nearly all unsafe contracts (GPT-4o-mini, $36.2\%$ overall and $33.9\%$ on the buyer side; Qwen2.5-14B, $20.0\%$ overall and $10.0\%$ on the seller side), whereas the one baseline model with reasoning capability, Gemini-2.5-Flash, sits at $2.9\%$, far closer to the mid-tier and flagship models, all of which are reasoning models, than to the other two baseline models (per-model breakdown in Figure~\ref{fig:irrationality_by_model_detailed}). The operative threshold for prompt-only deployment is therefore associated with the presence of reasoning capability rather than capability tier as such. The reasoning-effort ablation of Appendix~\ref{subsec:reasoning_effort} refines this distinction: increasing the reasoning budget within GPT-5.2 changes negotiation speed and discounted efficiency, but not deal rates, undiscounted efficiency, or surplus division. Because our two non-reasoning models are also the earliest-generation, we read reasoning capability as a refinement of the capability threshold rather than a separately identified axis.

Buyers suffer irrationality more than sellers ($5.1\%$ vs.\ $1.4\%$) despite their informational advantage, plausibly because seller feasibility is a deterministic cost-coverage check while buyer feasibility requires expected-profit evaluation under demand uncertainty. Low-type buyers err more than high-type ($6.3\%$ vs.\ $3.9\%$), reflecting tighter low-demand margins. Both error types occur alongside confident strategic language: revenue miscalculation ($79.0\%$, 109 of 138 cases; buyers accepting payments above expected revenue) and cost miscalculation ($21.0\%$, 29 of 138; sellers accepting payments below cost). These reinforce the value of automated profit verification as a deployment guardrail (whatever the source), developed in Section~\ref{sec:conclusion}. The gradient also tracks a reliability story: each round compounds process-failure probability, so baseline outcomes are strongest under quick termination while flagships' near-zero failure rate makes extended exploration productive.

\subsection{Why Agents Take Extra Rounds: Heuristic Substitution}
\label{sec:strategic_behavior}

The round-count gap documented in \S\ref{subsec:capability_value} has a clear process-level signature: LLM agents recognize the structure of the bargaining problem but substitute robust heuristics for the branch-specific execution the PBE prescribes. The first signature appears in opening offers. Buyers who move first reproduce the \citet{feng2015dynamic} truth-telling type separation (high-type quantities average $77.8$ units, low-type $41.6$, with opening-to-final correlations of $0.911$ for quantity and $0.862$ for payment), whereas sellers who open propose $60.5$ units on average (essentially the uniform-prior expected value of $60$ rather than the branch-specific screening menu), driving the $2.98$-round average against the $1.25$-round equilibrium prediction.

A second signature concerns type-strategic intent. After aggregating the turn-level classifications within each negotiation, the average signaling rate per negotiation is $64.7\%$ for low-type buyers versus $21.9\%$ for high-type buyers, while the average mimic rate per negotiation is $13.8\%$ for high-type buyers versus $1.3\%$ for low-type buyers. Both differences remain significant after Holm correction ($p<0.001$; Appendix~\ref{subsec:C_strategy}). The coded reasoning follows the pattern predicted by \citet{feng2015dynamic}: low-type buyers signal their type more often to separate, whereas high-type buyers mimic the low type more often to avoid surplus extraction. Consistent with this interpretation, high-type realized quantities are $7.1\%$ below first-best, although this quantity distortion alone does not establish mimicking. The comparative static, however, runs the wrong way: the average mimic rate is \textit{lowest}, not highest, under low buyer patience, so high-type buyers do not calibrate mimicking to the patience incentive on which the equilibrium turns. The full opening-offer breakdown, inferred-strategy distributions, and extended reasoning traces appear in Appendix~\ref{subsec:C_process_efficiency} (with further detail in Appendix~\ref{app:strategic_detail}).

\subsection{Capability Creates, Patience Divides: Orthogonal Drivers}
\label{subsec:drivers}

We next ask whether the same design variables govern value creation and value division. Table~\ref{tab:shapley_main} attributes explained variance in efficiency and buyer surplus share to four experimental blocks: buyer type, first proposer, patience, and model tier. Interaction and within-stratum tests (Appendix~\ref{subsec:B_efficiency_decomp}) verify the table's diagonal pattern is not driven by a few cells.

\begin{table}[t]
\centering
\caption{Variance Attribution: Efficiency and Buyer Surplus Share}
\label{tab:shapley_main}
\footnotesize
\begin{tabular}{lcccc}
\toprule
 & \multicolumn{2}{c}{Efficiency} & \multicolumn{2}{c}{Buyer surplus share} \\
\cmidrule(lr){2-3}\cmidrule(lr){4-5}
 & $R^2$ & \% expl. & $R^2$ & \% expl. \\
\midrule
\multicolumn{5}{l}{\textit{Variance attribution (Shapley $R^2$)}} \\
\quad Information (Buyer type) & 0.019 & 27\% & 0.003 & 2\% \\
\quad Bargaining Structure (First Proposer) & 0.000 & 0\% & 0.000 & 0\% \\
\quad Time Pressure (Patience) & 0.005 & 7\% & \textbf{0.138} & \textbf{90\%} \\
\quad Capability (Model tier) & \textbf{0.047} & \textbf{66\%} & 0.013 & 8\% \\
\addlinespace
\quad Total variance explained & \multicolumn{2}{c}{0.071} & \multicolumn{2}{c}{0.154} \\
\midrule
$N$ & \multicolumn{2}{c}{1,998} & \multicolumn{2}{c}{1,998} \\
\bottomrule
\end{tabular}
\par\smallskip
\begin{minipage}{\textwidth}
\scriptsize
\textit{Notes:} Sample: $N=1{,}998$ economically rational agreements; irrational deals are excluded because buyer surplus share is ill-defined when one party accepts a negative-profit contract. Entries report Shapley $R^2$ by design block and each block's share of explained variance within the outcome. Interaction tests and within-stratum coefficient estimates appear in Appendix~\ref{subsec:B_efficiency_decomp}.
\end{minipage}
\end{table}

The split is stark. Capability explains $66\%$ of the explained variation in efficiency but only $8\%$ of buyer-share variation; patience explains $90\%$ of buyer-share variation but only $7\%$ of efficiency variation. Buyer type matters for efficiency because demand information affects the operational quantity decision, but it has little direct role in surplus division once patience and model tier are accounted for. The appendix diagnostics confirm the same diagonal pattern: capability moves efficiency within patience strata but not buyer share, while patience moves buyer share within capability tiers but not efficiency. Thus capability creates value, patience divides it.

\subsection{Within-Family Cross-Model Negotiations}
\label{subsec:within_family_cross_model}

The driver separation documented above is established with homogeneous model pairs. Pairing models of different capability tiers within the same family stress-tests that finding and also reveals additional patterns worth distilling on their own. We pair GPT-5.2 (flagship) with GPT-5-mini (mid-tier) in both role assignments. Each configuration covers the full 16-condition factorial ($2$ buyer types $\times$ $2$ first proposers $\times$ $4$ patience levels) with 15 replications. We conduct the two cross-tier configurations in the verbal format ($N = 480$) and reuse the two homogeneous configurations from M1 as benchmarks; Table~\ref{tab:ext3_summary} therefore displays 960 observations. Four findings emerge, grouped below.

\textit{Efficiency is robust to tier mixing, and capability manifests as strategic delay rather than speed.} Cross-tier pairs reach agreement in every negotiation (100\% deal rate), and undiscounted efficiency is statistically indistinguishable from homogeneous pairs (96.8\% vs.\ 96.1\%, $p = 0.284$): high allocative efficiency survives capability heterogeneity even when one party has weaker newsvendor reasoning, so firms can mix capability tiers across buyer/seller sides without paying an efficiency penalty. The cleanest separation between GPT-5.2 and GPT-5-mini is on the seller side: GPT-5.2 as seller concedes only $35.2\%$ of surplus to a GPT-5-mini buyer, versus $52.3\%$ when GPT-5-mini sells to a GPT-5.2 buyer. GPT-5.2 achieves this by holding out $0.43$ rounds longer as seller ($2.46$ vs.\ $2.03$; $p < 0.01$). The dollar gap is correspondingly large (seller profit $\$931$ vs.\ $\$680$; $\Delta = +251$, $p < 0.001$; Appendix~\ref{subsec:B_cross_model_welfare}). The flagship thus converts its additional rounds into materially higher seller profit rather than faster closure, so within-family capability operates through extended willingness to delay, not accelerated convergence, localizing the aggregate finding that flagship models' longer negotiations reflect patience-driven extraction by the stronger party, not slower computation.

\textit{Within a family the more capable model claims a larger share in either role, but patience remains the dominant lever of division.} Capability translates into bargaining power in both roles. With GPT-5.2 as seller, switching the buyer from GPT-5.2 to the weaker GPT-5-mini \textit{lowers} the buyer's share ($37.9\%$ to $35.2\%$); conversely, with GPT-5-mini as seller, a GPT-5.2 buyer captures $52.3\%$ against GPT-5-mini's $41.9\%$ self-play baseline. The more capable model thus extracts more whether it buys or sells, and earns more in dollars in both roles (Appendix~\ref{subsec:B_cross_model_welfare}), so within a family capability rank does predict who captures surplus. Across providers, however, capability rank alone does not determine who captures surplus (Section~\ref{sec:provider}). Patience nonetheless remains the dominant driver of surplus division: buyer shares range from about $67\%$ under buyer-patient conditions ($\delta_B = 0.9$, $\delta_S = 0.4$) to about $26\%$ under seller-patient conditions ($\delta_B = 0.4$, $\delta_S = 0.9$). The within-family capability range ($\sim$17 pp) sits well below the patience range ($\sim$40 pp), and provider-role assignment shifts buyer shares by an amount comparable to within-family capability differences (Section~\ref{subsec:cross_family}). Full per-configuration statistics and the direction-effect visualization appear in Appendix~\ref{subsec:B_within_family_detail} (Table~\ref{tab:ext3_summary}, Figure~\ref{fig:ext3_direction}).

\section{Provider-Level Bargaining Profiles}
\label{sec:provider}

For firms deploying LLM agents on different sides of a procurement interaction, the first-order distributional question is whether the vendor they select systematically favors one role. We show that it does, and that provider identity predicts the direction of surplus flow more reliably than capability tier.

Table~\ref{tab:provider_bias} reveals the heterogeneity, reporting buyer surplus as a share of total realized surplus. Alibaba's Qwen models exhibit extreme buyer bias, with buyer shares ranging from $52.7\%$ to $91.4\%$ across capability tiers. Google's Gemini models cluster near parity, ranging from $44.3\%$ to $53.9\%$. OpenAI models are mildly \textit{seller}-leaning across all three tiers, most pronounced at the flagship ($37.9\%$) and closer to parity at the mid-tier ($41.9\%$) and baseline ($39.8\%$). Four of nine models show buyer shares below $50\%$, and the pooled $53.5\%$ average masks three distinct provider profiles.

\begin{table}[htbp]
\centering
\caption{Surplus Division by Provider: Buyer Share Across Capability Tiers}
\label{tab:provider_bias}
\footnotesize
\renewcommand{\arraystretch}{1.1}
\begin{tabular}{@{}lcccc@{}}
\toprule
\textbf{Provider} & \textbf{Flagship} & \textbf{Mid-Tier} & \textbf{Baseline} & \textbf{Provider Avg} \\
\midrule
OpenAI & 0.379 & 0.419 & 0.398 & 0.399 \\
Google & 0.539 & 0.514 & 0.443 & 0.499 \\
Alibaba & 0.914 & 0.668 & 0.527 & 0.703 \\
Pooled &  &  &  & 0.535 \\
\bottomrule
\end{tabular}
\end{table}

This provider-level heterogeneity has two implications. First, role-based surplus division is not a universal property of LLM negotiation: the three providers differ qualitatively, not merely in degree. These family-level patterns are consistent with provider-specific training and fine-tuning differences, but the current design does not identify which channel produces them. Second, organizations selecting LLM agents for procurement must evaluate provider-specific bargaining behavior, not merely model capability: deploying a Qwen model as buyer and an OpenAI model as seller would produce markedly different distributional outcomes than the reverse configuration, a prediction the cross-family experiments of Section~\ref{subsec:cross_family} confirm directly.

A second pattern layered on top of provider variation is universal compression toward equal division relative to the Bayesian benchmark. Under strong seller patience ($\delta_B = 0.4, \delta_S = 0.9$), the benchmark allocates approximately $13\%$ of surplus to the buyer, yet LLM buyers achieve $40.2\%$, a $+27.1$ pp deviation toward the buyer. Under strong buyer patience ($\delta_B = 0.9, \delta_S = 0.4$), the benchmark allocates approximately $91\%$, yet LLM buyers achieve only $70.9\%$, a $-20.5$ pp deviation toward the seller.
All nine models attenuate predicted extremes, pulling outcomes toward roughly even division.
First-mover identity has little effect on the split: buyers capture 54.1\% proposing first versus 52.8\% proposing second, and even seller-first buyers exceed an equal split, indicating advantage well beyond procedural positioning. Surplus division across patience, tier, first-proposer, and buyer-type cuts appears in Appendix~\ref{subsec:B_surplus_distribution} (Table~\ref{tab:surplus_detailed}).

\subsection{Interpreting Provider-Level Heterogeneity}
\label{subsec:provider_interpret}

A universal RLHF-based explanation does not fit: a common buyer-favoring tendency from post-training would produce broadly similar directional effects, yet Qwen exhibits large buyer advantages, Gemini clusters near parity, and OpenAI spans near-parity to seller-leaning. The cleaner reading is comparative rather than causal (under a common prompting regime, provider identity is associated with distinct bargaining profiles), with three descriptive correlates surfaced by the cross-family evidence below: anchoring discipline, concession posture, and how explicitly models track patience asymmetries. Several channels could generate such profiles (training-data composition, reward criteria, safety tuning, annotation guidelines), and isolating one would require varying post-training while holding base capability fixed; we treat the profiles as descriptive, and the managerial implication is to select providers whose profile aligns with strategic objectives.

The verbal channel reinforces this comparative reading: an LLM classifier flags public--private divergence (public fairness appeals paired with self-interested private reasoning) as a population-average behavior whose prevalence is itself provider- and role-conditional. Using buyer-minus-seller differences, Google buyers have lower average within-negotiation rates of both fairness mismatch ($-9.3$ pp) and profit hiding ($-3.0$ pp) than Google sellers, whereas the OpenAI gaps are small or nonsignificant and the Qwen gaps run in the opposite direction ($+8.7$ and $+3.3$ pp; full breakdown in Appendix~\ref{subsec:C_process_verbal}, Table~\ref{tab:deception_tactics_family_role}).

The same evidence argues against a simple computational-asymmetry account in which buyers underperform because newsvendor evaluation is harder than cost checking: buyers show higher irrationality than sellers, yet aggregate buyer surplus still exceeds the Bayesian reference for most models, and buyer-first negotiations yield the largest buyer shares.

\subsection{Cross-Family Flagship Negotiations}
\label{subsec:cross_family}

While our main analysis pairs each model with itself, this subsection considers asymmetric LLM deployment by crossing provider boundaries entirely, pairing flagship models from different AI families against one another: GPT-5.2 (OpenAI), Gemini-3-Pro (Google), and Qwen3-Max (Alibaba). Each pair is tested in both directions (each model serving as both buyer and seller), yielding six directional configurations. Each configuration covers 16 conditions (2 buyer types $\times$ 2 first proposers $\times$ 4 patience levels), with 15 replications per condition.
All experiments use the verbal treatment with medium reasoning effort. The total sample is $1{,}440$ attempted negotiations, all of which reached agreement.

This analysis connects directly to the provider-specific bargaining profiles documented above and asks what happens when these profiles collide at the bargaining table. The within-family results of Section~\ref{subsec:within_family_cross_model} already imply that a model's bargaining advantage is \textit{relational, not intrinsic}: GPT-5.2's 37.9\% self-play buyer share reflects the mirrored equilibrium of two agents sharing the same strategic tendencies, not a fixed property of the model in the buyer role. Cross-family matchups test how provider dispositions combine when they are not mirrored.

Cross-family flagship pairs reach agreement in all negotiations: GPT $\leftrightarrow$ Gemini achieves a 100\% deal rate ($N = 480$), GPT $\leftrightarrow$ Qwen achieves 100\% ($N = 480$), and Gemini $\leftrightarrow$ Qwen achieves 100\% ($N = 480$); the full performance statistics across all six directional configurations appear in Appendix~\ref{subsec:B_flagship_detail} (Table~\ref{tab:ext2_summary}).

\begin{figure}[htbp]
\centering
\includegraphics[width=0.82\textwidth]{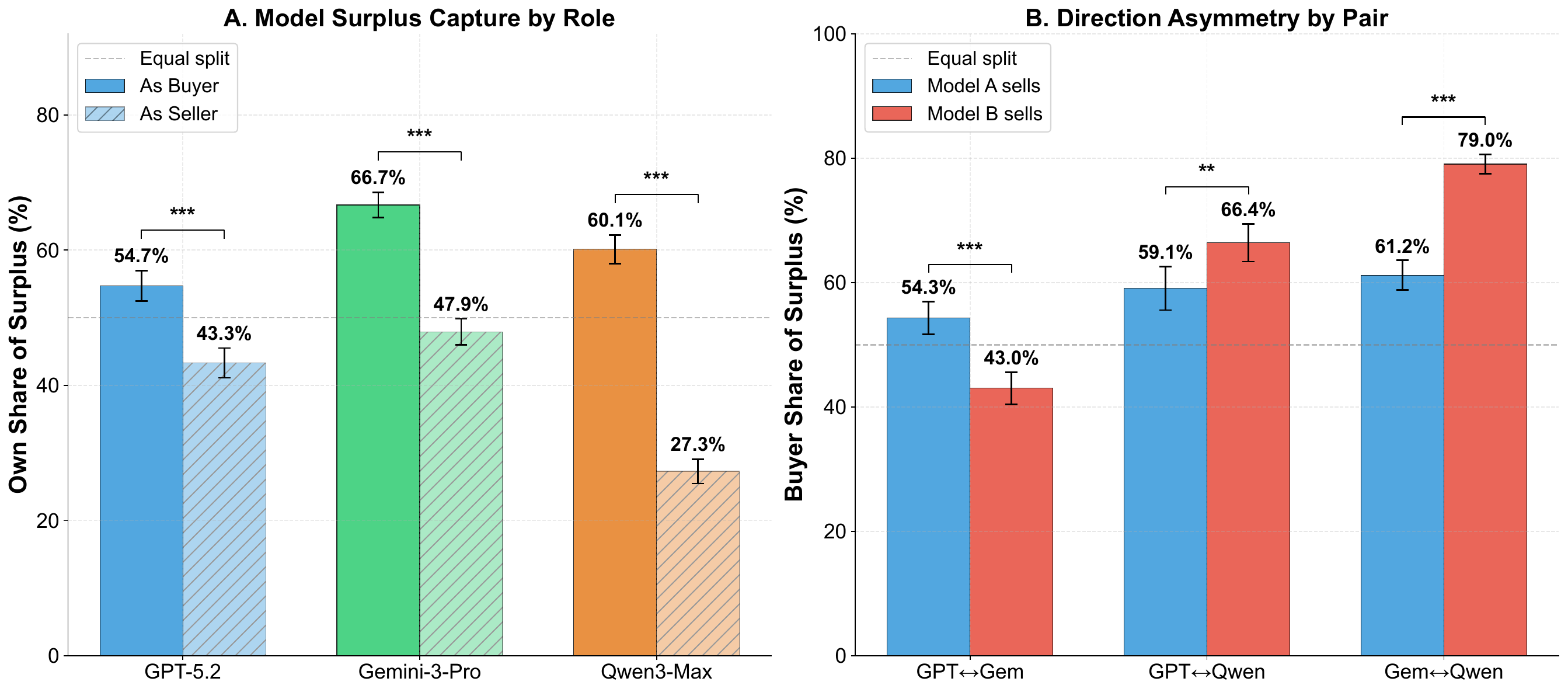}
\caption{Direction Effects in Cross-Flagship Negotiations}
\label{fig:ext2_direction}
\par\smallskip
\begin{minipage}{\textwidth}
\scriptsize
\textit{Notes:} Panel~(a) shows each model's own surplus share as buyer (solid) vs.\ seller (hatched). Panel~(b) shows buyer share by direction within each pair. Panels~(a) and~(b) show significance brackets from independent two-sample $t$-tests comparing the displayed role or direction groups. Error bars are 95\% confidence intervals. Significance levels:
$^{*}\,p<0.05$, $^{**}\,p<0.01$, $^{***}\,p<0.001$.
\end{minipage}
\end{figure}

Negotiation outcomes are strikingly asymmetric across providers (Figure~\ref{fig:ext2_direction}). Gemini-3-Pro is near parity in self-play but strongest in heterogeneous pairings (66.7\% of surplus as buyer, 47.9\% as seller); GPT-5.2 occupies the middle (54.7\% / 43.3\%); and Qwen3-Max is weakest as a cross-family seller, retaining only 27.3\%. The two designs measure different objects: self-play asks what division looks like when both sides share a profile, cross-family play which profile is stronger when they differ. A provider can thus be near parity in self-play (its anchoring and concession discipline mirrored by an identical counterparty) yet dominate heterogeneous matchups when those traits are not reciprocated. This reconciles two mirror-image patterns: Gemini's aggressive anchoring, which self-play offsets but which is decisive against softer GPT or Qwen counteroffers, and Qwen's buyer-favoring self-play profile, which conceals a seller-side softness---costly whenever Qwen's counterparty does not share that softness. Qwen sellers retain only 33.6\% against GPT-5.2 buyers and 21.0\% against Gemini buyers.

\textit{Direction effects and the direction-effect hierarchy.} The direction of the pairing produces large, statistically significant effects on surplus division. In GPT $\leftrightarrow$ Gemini negotiations, buyer share swings 11.3 pp (54.3\% when GPT sells to Gemini versus 43.0\% when Gemini sells to GPT; $t = 6.01$, $p < 0.001$), with Gemini capturing more surplus regardless of role. The asymmetry is largest for the Qwen $\leftrightarrow$ Gemini pair: the buyer captures 79.0\% when Qwen sells but only 61.2\% when Gemini sells, a 17.8 pp swing ($t = 12.25$, $p < 0.001$) that compounds Gemini's strength and Qwen's weakness; GPT $\leftrightarrow$ Qwen shifts 7.3 pp by direction ($t = 3.09$, $p < 0.01$). These cross-family direction effects (7--18 pp) are comparable to the within-family capability swing ($\sim$17 pp) from Section~\ref{subsec:within_family_cross_model}, and both sit well below the patience swings ($\sim$35 pp), so provider choice and capability tier are comparable levers, each second-order to patience. Patience nevertheless dominates: buyer shares reach 80.6\% under buyer-patient conditions ($\delta_B = 0.9$, $\delta_S = 0.4$) versus 46.1\% under seller-patient conditions ($\delta_B = 0.4$, $\delta_S = 0.9$), tracking the Bayesian benchmarks of 91.4\% and 13.1\% directionally but with the characteristic compression toward equal division ($-10.8$ pp and $+33.0$ pp deviations).

\textit{Efficiency: Qwen's inefficiency concentrates in the seller role.} Undiscounted efficiency exceeds 95\% for every pair (GPT $\leftrightarrow$ Gemini 98.7\%, Gemini $\leftrightarrow$ Qwen 96.5\%, GPT $\leftrightarrow$ Qwen 95.1\%), so cross-family pairing does not meaningfully reduce allocative efficiency. Decomposing the gap, GPT $\leftrightarrow$ Gemini loses only 1.25 pp (zero deal failures, zero irrational agreements), GPT $\leftrightarrow$ Qwen 4.88 pp, and Gemini $\leftrightarrow$ Qwen 3.50 pp, almost all of it from suboptimal contract terms rather than failed or loss-making deals. The direction-level pattern is unambiguous: when Qwen sells, efficiency drops sharply (Qwen $\to$ GPT 91.9\%, an 8.1 pp gap; Qwen $\to$ Gemini 94.0\%, a 6.0 pp gap), whereas GPT or Gemini selling to Qwen stays above 98\%. All three irrational agreements in the sample occur with Qwen as seller (two in Qwen $\to$ GPT, one in Qwen $\to$ Gemini, out of $240$ deals each); Qwen's seller-side inefficiency, not its counterparty, is the common thread.

\textit{Process correlates.} Reasoning traces show recurring differences behind this ordering: Gemini-3-Pro articulates the patience structure directly and anchors aggressively (claiming 82--92\% of surplus on the first offer, near-efficient $q \approx 80$ at 4--6\% above cost), GPT-5.2 reasons in terms of fair splits and cost coverage, and Qwen3-Max exposes little reasoning and defaults to passive cost-plus counter-offers as seller, with concession discipline mirroring the same ranking. Counterintuitively, Gemini is descriptively the \textit{least} Bayesian-aligned (mean buyer-side deviation $+$21.5 pp vs.\ $+$9.5 for GPT and $+$15.0 for Qwen): its advantage reflects not closer alignment but a consistent ability to extract surplus above the benchmark. Because all three are flagships, the safer reading is that provider identity carries distinct bargaining profiles in heterogeneous matchups (the cross-family analogue of the within-family pattern of Section~\ref{subsec:within_family_cross_model}), so which model represents each party can systematically shift the distribution of economic value. Supporting figures appear in Appendix~\ref{subsec:B_flagship_detail} (Figures~\ref{fig:ext2_advantage} and~\ref{fig:ext2_concession}).

\section{Design Choices in Deploying LLM Bargaining Agents}
\label{sec:design_choice}
A firm deploying an LLM bargaining agent does not bargain directly; it configures an agent and delegates the task, a principal--agent relationship in which the prompt functions as the contract \citep{imas2025agentic}. The principal selects which messages the agent may exchange, whether discounting is encoded in the prompt, and what patience parameters to assign. Our main experiment fixed these at a single specification (verbal channel enabled, discounting prompted, patience assigned), so the patterns of Sections~\ref{sec:performance} and~\ref{sec:provider} are silent on how behavior shifts elsewhere in the configuration space. This section examines the three dimensions in turn (the verbal channel, the discounting framework, and the focal agent's patience) and asks of each the same question: do alternative specifications produce systematically different outcomes that a principal should anticipate?

\subsection{Does the Verbal Channel Do Strategic Work? (R1)}
\label{subsec:r1_maintext}

The first configuration dimension is the message protocol. Production LLM bargaining agents generate natural-language messages by default, yet the contract itself is numerical, so we ask whether language moves outcomes or merely accompanies offer behavior. We test this by re-running the main factorial with the verbal channel disabled: agents exchange only numeric offers and accept/reject decisions, with all other features held constant. This isolates whether the provider profiles and high-efficiency-with-delay pattern documented above live in linguistic persuasion or in the numerical proposals themselves. The question also connects to evidence from open-ended AI negotiation that linguistic style affects agreement and value creation \citep{vaccaro2026advancing}; our setting asks whether language carries comparable \textit{distributional} weight once offers are structured and surplus is objective.

Mirroring the main analysis performance table, structured agents achieve 96.4\% agreement (vs.\ 98.9\% verbal), 3.15 rounds (vs.\ 2.98), and 92.8\% undiscounted efficiency (vs.\ 95.4\%). The high-efficiency-with-delay pattern therefore reproduces, but the protocols are not statistically identical: removing verbal communication modestly lowers agreement and efficiency and slightly lengthens bargaining. The capability gradient also replicates: flagship structured agents achieve 98.3\% efficiency at 3.23 rounds; baseline agents manage only 87.7\% at 2.81 rounds. The full performance table by tier, proposer, buyer type, and patience, with between-treatment tests, appears in Appendix~\ref{subsec:r1_structured} (Table~\ref{tab:ext1_struct_vs_verbal}).

\paragraph{Provider-Specific Bias Persists Without Communication.}
The provider-specific surplus division patterns documented in the main analysis survive the removal of verbal communication. This rules out a purely language-based explanation for the provider profiles: the distributional regularities also appear in how models generate and evaluate numerical proposals. This qualifies the language-centric account of AI negotiation for structured-offer settings: where \citet{vaccaro2026advancing} find linguistic warmth driving outcomes in open-ended dialogue, our distributional regularities persist in numerical offer behavior even when language is removed. As the next paragraph shows, however, the verbal channel remains an execution aid and a provider-specific distributional lever.

\paragraph{Model-Specific Heterogeneity.}
Pooled buyer share changes little, but model-level distributional shifts diverge: removing communication shifts OpenAI's baseline and mid-tier models toward buyers and Gemini's mid-tier and flagship models toward sellers. The remaining five models, GPT-5.2, Gemini-2.5-Flash, Qwen2.5-14B, Qwen3-32B, and Qwen3-Max, show no detectable buyer-share shift (full model-by-model breakdown in Appendix~\ref{subsec:r1_structured}, Table~\ref{tab:ext1_model_detail}). The managerial reading is that the verbal channel is itself a provider-specific lever: for OpenAI's baseline and mid-tier models, language moderates the buyer advantage, whereas for Gemini's mid-tier and flagship models it amplifies it, so constraining communication is not a uniformly buyer- or seller-favoring intervention.

\subsection{No-Discounting Bargaining (R2)}
\label{subsec:r2_maintext}
The second configuration dimension is whether discounting is encoded in the agent's prompt at all. Our main experiment presents agents with an explicit utility formula $U_k = \pi_k \delta_k^{\tau-1}$ and informs them of the patience parameters that govern it. This framing supports clean comparison against the Feng et al.\ equilibrium but it does not match how production systems frame the task: systems such as Pactum and Arkestro present agents with contractual terms, permissible ranges, and fallback rules, but do not parameterize the agent's utility with a time-preference coefficient. R2 therefore tests how LLM agents bargain when the discounting framework is removed from the prompt entirely, comparing the symmetric-patience baseline with a no-discounting treatment that removes $\delta$ from both the prompt and payoff computation ($N=540$ per treatment). Both treatments are evaluated on a common undiscounted basis along the three structural margins of the main analysis: high efficiency with delayed agreement, the cross-provider distributional profile, and the capability--irrationality gradient. The full performance comparison appears in Appendix~\ref{app:ext_unprompted} (Table~\ref{tab:ext_unprompted_efficiency_delay}).

Removing the prompted discount factor leaves the main outcome regularities intact but lengthens bargaining. The high-efficiency-with-delay pattern reproduces: efficiency is statistically indistinguishable across treatments and within every tier, while rounds rise from 3.39 to 4.02 overall and increase in each capability tier. The cross-provider distributional ordering also reproduces despite a marginal Treatment $\times$ Provider interaction ($F=2.44$, $p=0.087$; Qwen $\gg$ Google $>$ OpenAI). And the capability-dependent irrationality gradient reproduces, declining monotonically from baseline through mid-tier to flagship. With discounting removed we compare raw buyer shares against the symmetric-patience baseline rather than the Feng et al.\ benchmark, whose patience-based surplus division is no longer defined. Full design, panel-by-panel tests, and the cuts by first-proposer role and buyer type appear in Appendix~\ref{app:ext_unprompted} (Table~\ref{tab:ext_unprompted_summary_nodup}).

\subsection{Prompted Patience as a Principal-Specification Lever (R3)}
\label{subsec:strategic-patience}

The third configuration dimension is the patience level the principal assigns to the agent: a continuous parameter, unlike the present-or-absent verbal-channel and discounting choices. Section~\ref{sec:patience_interp} introduced the strategic/economic patience distinction this dimension rests on: under delegation to LLM agents the prompted \textit{strategic patience} $\delta^{\text{strat}}$ (the agent's per-round discount, on which the equilibrium is defined) decouples from \textit{economic patience} $\delta^{\text{econ}}$ (the principal's fixed real-time cost of delay), so the principal \textit{chooses} $\delta^{\text{strat}}$ while $\delta^{\text{econ}}$ stays a constraint. The specification problem is to set $\delta^{\text{strat}}$ to maximize realized payoff given the counterparty's choice and the principal's $\delta^{\text{econ}}$. This exercise is conditional rather than game-theoretic: we vary the focal agent's prompted patience while holding the counterparty environment fixed, and do not solve a two-sided meta-game in which both principals jointly choose prompt parameters or agents infer undisclosed patience. The remainder of this subsection asks whether varying $\delta^{\text{strat}}$ in our data produces outcome differences a principal could anticipate.

We vary focal-agent strategic patience across $\{0.9,\,0.7,\,0.4\}$ separately for the buyer and seller roles, holding the counterparty at the maximally demanding $\delta^{\text{strat}} = 0.9$; we further disaggregate the buyer side by type (H vs.\ L), since the Feng et al.\ equilibrium predicts the high type's information rent and the low type's signaling cost respond to own patience along different margins. Two questions organize the analysis. \textit{Part~1}: does varying $\delta^{\text{strat}}$ produce systematic outcome differences at all? \textit{Part~2}, if yes: which way should the principal pull the lever, and does the answer differ by role or buyer type? For each (model, buyer-type, first-proposer) cell we compute realized payoff at the principal's true economic patience, identify the empirically best prompted $\delta^{\text{strat}}$, and define $\text{gain} = \text{best payoff} - \text{matched-patience payoff} \geq 0$, with the matched benchmark setting $\delta^{\text{strat}} = \delta^{\text{econ}}$. Headline numbers evaluate at $\delta^{\text{econ}} = 0.9$ across $9$ models $\times$ $4$ buyer-type/first-proposer conditions; the full results, by party and economic-patience level, appear in Appendix~\ref{subsec:C_patience_heterogeneity} (Table~\ref{tab:ext9_summary}).

R3 adds the $(\delta_B,\delta_S)=(0.9,0.7)$ configuration: $9$ models $\times$ $2$ buyer types $\times$ $2$ proposer orders $\times$ $15$ replications, or 540 new negotiations. Combined with 2,160 reused M1 observations, the full analysis sample is $N=2{,}700$.

\paragraph{Part 1: Is strategic patience a lever?}
Both headline rows of Table~\ref{tab:patience_maintext} show that the agent's strategic patience is a payoff-relevant choice for the principal.\footnote{We conduct separate stratified permutation tests for each role and economic-patience level, permuting prompted-patience labels within each model $\times$ buyer-type $\times$ first-proposer stratum while preserving 15 replications per level. All four tests reject equality of mean payoffs across $\delta^{\mathrm{strat}}\in\{0.4,0.7,0.9\}$ (19{,}999 permutations; Holm-adjusted $p<0.001$ throughout).} Descriptively, the cell-level mean potential in-sample gain is $+74.81$ on the buyer side (95\% bootstrap interval $[+29.43,\,+130.17]$) and $+18.28$ on the seller side ($[+4.35,\,+35.06]$). The payoff curves are not flat in $\delta^{\text{strat}}_B$ (Appendix~\ref{subsec:C_patience_heterogeneity}, Figure~\ref{fig:strategic_patience_buyer}). Strategic patience is a lever; the principal cannot ignore it.

\paragraph{Part 2: Which value of strategic patience should the principal specify?} For buyer principals, the guidance is more heterogeneous than a single rule. Matching ($\delta^{\text{strat}}_B = \delta^{\text{econ}}$, here $0.9$) is the most common per-condition optimum, winning $20/36$ ($56\%$) cells, and remains the safe default. But it is the average-best choice for only $4/9$ models (GPT-5.2 and the three Gemini models); the other five do better understating patience, and the gains are large: GPT-4o-mini realizes $+278$ payoff units and GPT-5-mini $+234$ at $\delta^{\text{strat}}_B = 0.7$ (Figure~\ref{fig:strategic_patience_heterogeneity}, Panel~A). The mechanism in the payoff curves depends on who proposes first (Figure~\ref{fig:strategic_patience_buyer}). When the buyer proposes first, the H-type payoff rises sharply from $\delta^{\text{strat}}_B = 0.4$ to $0.7$ and then flattens, while the L-type payoff continues rising through the matched value of $0.9$. When the seller proposes first, both buyer types peak at $\delta^{\text{strat}}_B = 0.7$, so maximal announced patience lowers realized payoff in those cells. The stakes are larger in the H-type configuration, where both payoffs and the spread across $\delta^{\text{strat}}_B$ exceed the L-type's.

For seller principals, by contrast, matching is robust. The matched choice ($\delta^{\text{strat}}_S = \delta^{\text{econ}}$, here $0.9$) is per-condition optimal in $26/36$ ($72\%$) cells and is the average-best prompted patience for all $9/9$ models, with no model improving on it at the model-averaged level (Figure~\ref{fig:strategic_patience_heterogeneity}, Panel~B). The seller recommendation is therefore unambiguous at this economic patience: match. The residual mean gain of $+18.28$ (Table~\ref{tab:patience_maintext}) reflects scattered single-condition cells in which a strategic alternative edges out matching, not a systematic model-level pattern: no seller model, baseline tier included, gains from strategic understatement on average.

At $\delta^{\text{econ}} = 0.7$ (Table~\ref{tab:patience_maintext}, bottom two rows), the lever finding remains, but the preferred direction changes. The mean gains remain positive ($+41.11$ buyer, $+66.53$ seller), with $13/36$ buyer cells and $11/36$ seller cells matched-optimal. Buyer guidance is model-dependent: matching is the average-best choice for $4/9$ models, while three models (GPT-5.2, Gemini-2.5-Flash, and Qwen3-Max) favor overstatement to $\delta^{\text{strat}}_B = 0.9$ and the two Gemini-3 models favor understatement to $\delta^{\text{strat}}_B = 0.4$. Seller guidance reverses: $\delta^{\text{strat}}_S = 0.9$ is average-best for $7/9$ models, while Gemini-3-Flash and GPT-5.2 favor the matched value of $0.7$. Thus, at lower economic patience, matching remains the most common buyer-side choice, whereas seller-side guidance favors strategic overstatement.

\begin{table}[htbp]
\centering
\footnotesize
\renewcommand{\arraystretch}{1.05}
\caption{Prompted Patience Is a Principal-Specification Lever}
\label{tab:patience_maintext}
\begin{tabular}{@{}llccl@{}}
\toprule
$\delta^{\mathrm{econ}}$ & Party & \makecell{Matching is best\\(of 36 cells)} & \makecell{Potential gain from\\cell-specific tuning} & Recommended prompt rule \\
\midrule
\multirow{2}{*}{0.9} & Buyer  & 20 & $+74.81$~{\scriptsize$[+29.4,\,+130.2]$} & Match (modal); understate for most \\
                     & Seller & 26 & $+18.28$~{\scriptsize$[+4.4,\,+35.1]$} & Match: set $\delta^{\mathrm{strat}}=0.9$ \\
\midrule
\multirow{2}{*}{0.7} & Buyer  & 13 & $+41.11$~{\scriptsize$[+25.1,\,+58.4]$} & Match (most common); tune by model \\
                     & Seller & 11 & $+66.53$~{\scriptsize$[+39.0,\,+100.5]$} & Set $\delta^{\mathrm{strat}}=0.9$; match for two models \\
\bottomrule
\end{tabular}
\par\smallskip
\begin{minipage}{\textwidth}
\scriptsize
\textit{Notes:} Each row compares the matched choice $\delta^{\mathrm{strat}} = \delta^{\mathrm{econ}}$ with the best of $\{0.9, 0.7, 0.4\}$ across 36 cells (9 models $\times$ 2 buyer types $\times$ 2 first-proposer orders). Potential gain is best minus matched payoff (mean over cells; 95\% percentile bootstrap CI). The recommended rule is the portable default; the model-level heterogeneity, which models gain from deviating from matching and by how much, is summarized above and detailed in Appendix~\ref{subsec:C_patience_heterogeneity}.
\end{minipage}
\end{table}

Three implications follow for a principal specifying an agent's strategic patience. First, the choice is consequential: the mean gains in Table~\ref{tab:patience_maintext}, set against quantity $\times$ price scales of a few hundred, mean that a principal who treats $\delta^{\text{strat}}$ as irrelevant leaves surplus on the table. Second, the optimum depends on role, model, and the principal's true cost of delay. When delay is cheap ($\delta^{\text{econ}}=0.9$), matching strategic to economic patience is average-best for every seller model, whereas five of nine buyer models benefit from understatement. When delay is moderately costly ($\delta^{\text{econ}}=0.7$), seven of nine seller models benefit from overstatement to $\delta^{\text{strat}}=0.9$, while the buyer-side optimum varies by model: matching maximizes mean payoff for four models, overstatement for three, and understatement for two. Third, the principal should treat $\delta^{\text{strat}}$ as a prompt-specification variable evaluated against $\delta^{\text{econ}}$, not as a literal translation of human patience. We defer to future work the meta-game in which both principals jointly optimize $\delta^{\text{strat}}$.

\subsection{Additional Robustness Extensions (R4--R7)}
\label{sec:robustness}

The verbal channel (R1, \S\ref{subsec:r1_maintext}), prompted discounting framework (R2, \S\ref{subsec:r2_maintext}), and strategic-patience analysis (R3, \S\ref{subsec:strategic-patience}) were examined above as configuration levers. Four additional extensions probe the remaining design choices most likely to unsettle the findings: reasoning effort (R4), surplus magnitude (R5), the seller's prior (R6), and model size (R7). The outcome regularities of Sections~\ref{sec:performance}--\ref{sec:provider} survive all four: high undiscounted efficiency with delayed agreement, the provider rank ordering (with provider-specific exceptions catalogued in the appendix), and the capability--reliability gradient all persist, whereas the process mechanisms of Section~\ref{sec:strategic_behavior} and Appendix~\ref{subsec:C_disclosure} are validated less directly, since none of R4--R7 re-run the turn-level strategy inference or disclosure labeling. The reasoning-effort and model-size ablations (R4 and R7) distinguish reliability from tempo. The parameter-size ladder (R7) sharpens the reliability gradient into a continuous within-family slope, while both model size ($F = 18.0$, $p < 0.001$) and reasoning effort ($F = 11.22$, $p < 0.001$) affect negotiation speed. In R4, greater reasoning effort also raises discounted efficiency while undiscounted efficiency does not change significantly; within R7, the relation between size and speed is non-monotonic. The complete program (all seven extensions, with per-extension results and the full inventory) appears in Appendix~\ref{sec:extension_program} (Table~\ref{tab:inventory}).

\section{Conclusion}
\label{sec:conclusion}

We examine how LLM agents negotiate in a dynamic bargaining game with asymmetric information, and three conclusions emerge. First, \textit{capability is the value-creation lever}. It governs allocative efficiency: agents reach agreement in $98.9\%$ of cases and capture $95.4\%$ of first-best surplus, but take more than twice the benchmark rounds, so discounting erodes 21--34\% of first-best surplus, depending on patience. The same ordering governs operational reliability, where baseline models accept individually irrational agreements an order of magnitude more often than the upper tiers. Second, \textit{provider identity is the distributional lever}, and bargaining power is relational, not intrinsic: capability and provider load on different margins, surplus division varies markedly across providers, cross-family direction effects rival within-family capability effects, and a model's provider profile can override its capability across families. Third, \textit{the principal's configuration choices are strategic levers}: the communication channel reshapes provider biases, removing the prompted discounting framework preserves the main regularities while lengthening bargaining, and delegation to an LLM agent gives the principal a choice absent from classical direct bargaining: economic patience remains the fixed cost of delay, while the agent's strategic patience can be chosen at deployment.

Empirically, we document provider-level bargaining heterogeneity that is as consequential as within-family capability differences once counterparties are heterogeneous, establishing vendor choice as a strategic operations decision rather than a technical one. Methodologically, we introduce an executable implementation of the \citet{feng2015dynamic} PBE, validated cell-by-cell against the equilibrium's analytical properties, enabling evaluation of autonomous agents against a theoretical benchmark rather than human baselines alone; the approach illustrates a template for benchmarking agents in other game-theoretic operational settings, though each new setting requires its own equilibrium solution. Practically, the findings yield a three-dimensional deployment framework developed below.

The framework comprises three deployment rules that follow directly from the findings. \textit{Time-adjusted efficiency} shifts the locus of control from outcome to process: because deal-level outcomes carry meaningful stochasticity even under flagship models, organizations should bound delay ex ante through round caps, escalation rules for stalled negotiations, and final-offer procedures for high-urgency contracts. \textit{Distributional profile} makes vendor choice first-order whenever counterparties run different providers: cross-family direction effects ($7$--$18$ pp) are comparable to within-family tier effects ($\sim$17 pp), so firms controlling both sides of a transaction should weight provider identity at least as heavily as capability rank when assigning roles, and, because self-play profiles do not transfer directly to heterogeneous matchups, audit the specific cross-vendor pairing in simulation before delegation. \textit{Operational reliability} separates into two regimes: the $0.0$--$0.6\%$ irrationality rate at flagship and mid-tier permits lighter monitoring, while the $19.2\%$ rate at baseline makes dual-party automated profit verification non-negotiable. Because arithmetic errors, failed constraint checks, and instruction non-compliance are operationally equivalent (each yields an economically unsafe contract), the verification layer should treat them uniformly.

We distinguish \textit{structural regularities} (high undiscounted efficiency with delayed agreement, provider-level heterogeneity in surplus division, and concentration of unsafe agreements among weaker models) from \textit{model-specific results} such as specific surplus shares, irrationality rates, and provider rankings. The former are likely to travel across nearby settings; the latter are calibrated to the model versions tested here and require periodic recalibration as providers update their systems. This division rests on evidence, not assertion: across the three capability tiers and roughly three model generations we test, the structural regularities recur at every point along the frontier, including the reliability gradient traced continuously by the R7 capacity ladder, while only the model-specific results shift with capability. Extending the structural claims to models outside this tested range is a prediction the design supports but does not itself verify.

Three limitations bound the interpretation of these results. First, prompts are role-specific, so the reported surplus shares reflect behavior under a common prompting regime and are not prompt-free estimates of intrinsic bargaining bias; the qualitative pairing order is broadly stable across our prompt variations, with specific exceptions identified in Section~\ref{sec:robustness}. Second, our design discloses each party's patience parameters as common knowledge to both agents, matching the informational structure of the Bayesian benchmark; whether the qualitative patterns persist when a principal's patience or urgency is undisclosed to the counterparty, a common feature of real negotiations, is untested. Third, several interpretive moves remain process-descriptive rather than mechanistically identified: the strategic-behavior evidence of Section~\ref{sec:strategic_behavior} documents heuristic substitution and a behavior--theory gap in equilibrium comparative statics without adjudicating among competing cognitive explanations, and the hazard-based reading of $\delta$ in Section~\ref{sec:patience_interp} is an analogy rather than a tested mechanism.

These limitations point to a concrete research agenda. The hazard-based reading of $\delta$ is testable by exogenously varying per-round reliability (through compute budgets, injected parsing noise, or forced-termination protocols), and the related conjecture that effective $\delta$ declines as conversation history accumulates can be tested by measuring error rates against round number across context-management regimes. If these mappings hold, engineering choices organizations already make carry bargaining consequences (round budgets as value-conditional commitment devices, profit-verification guardrails as credible commitments, and an agent-versus-human boundary drawn along per-interaction reliability rather than general capability). Field studies, in turn, would assess which structural regularities survive once agents are embedded in real approval and verification workflows. As LLM agents move from pilot to production in procurement, the operational question shifts from whether a delegated agent can close a deal to whether the deal it closes is one the principal would have authorized. The audit developed here answers that question before delegation, not after---replacing the intuition that a more capable agent is simply a better bargainer with a discipline that treats capability, provider, and configuration as three separate levers a principal must each get right.

\makeatletter
\renewcommand{\@biblabel}[1]{} 
\let\OLDthebibliography\thebibliography
\renewcommand{\thebibliography}[1]{%
  \OLDthebibliography{#1}%
  \setlength{\itemsep}{0pt}%
  \setlength{\parskip}{0pt}%
  \setlength{\baselineskip}{16pt}%
}
\makeatother
\bibliographystyle{ormsv080}
\bibliography{2026_Agent.bib}
\newpage
\ECSwitch
\vspace{-20pt}
\begin{APPENDICES}
\renewcommand{\theHsection}{EC.\arabic{section}}
\renewcommand{\theHsubsection}{EC.\arabic{section}.\arabic{subsection}}
\renewcommand{\theHsubsubsection}{EC.\arabic{section}.\arabic{subsection}.\arabic{subsubsection}}

\begin{center}
{\Large \textbf{Online Appendix}}
\end{center}
\vspace{1em}

The appendix is organized to support reproducibility first and interpretation second: Appendix~\ref{app:prompts} reports the agent prompts, Appendices~\ref{app:performance_details} and~\ref{app:strategic_behavior_details} provide the detailed evidence behind the performance and process findings, Appendix~\ref{sec:extension_program} reports the design and robustness extension program, and Appendix~\ref{Bayesian-Details} documents the Bayesian benchmark implementation.

\section{Prompts}
\label{app:prompts}

\begin{promptbox}[Prompt for the Buyer Agent]
You are a buyer negotiating a supply contract with a seller.

\vspace{0.5em}
\textbf{THE NEGOTIATION:}
\begin{itemize}
    \item You buy a product from a seller to retail to end consumers
    \item You negotiate over \textbf{quantity} ($q$) and \textbf{total payment} ($T$)
    \item Your expected revenue: $R(q) = r \cdot \mathbb{E}[\min(D, q)]$, where $D \sim \mathcal{N}(\mu, \sigma)$
    \item Your profit: $\pi_B = R(q) - T$
    \item Your effective utility: $U_B = \pi_B \times \delta_B^{(\tau-1)}$, where $\delta_B$ is buyer's patience and $\tau$ is round number
    \item You have \textbf{private information} about demand: you know if you're HIGH or LOW type
    \item The seller doesn't know your type but has beliefs about it
    \item Seller's effective utility: $U_S = (T - c \cdot q) \times \delta_S^{(\tau-1)}$
    \item You cannot accept if $\pi_B < 0$
\end{itemize}

\vspace{0.5em}
\textbf{THE NEGOTIATION PROCESS:}
\begin{itemize}
    \item Either seller or buyer proposes first (check ``First proposer'' in context)
    \item Players alternate: proposer makes offer, responder accepts or rejects
    \item If rejected, roles switch and the other player proposes
    \item The proposer gets a larger share of surplus than the responder
    \item Maximum rounds: \texttt{\{p.max\_rounds\}}. If no deal by final round, profit is outside option.
\end{itemize}

\vspace{0.5em}
\textbf{YOUR OBJECTIVE:}
Maximize your effective utility through negotiation over $q$ and $T$.

\vspace{0.5em}
\textbf{YOUR PRIVATE ADVANTAGE:}
\begin{itemize}
    \item You know your true demand type
    \item $H$-type: higher demand $\rightarrow$ higher optimal $q$ $\rightarrow$ higher potential surplus
    \item $L$-type: lower demand $\rightarrow$ lower optimal $q$ $\rightarrow$ lower potential surplus
\end{itemize}

\vspace{0.5em}
\textbf{RESPONSE FORMAT: IF PROPOSER}
\begin{verbatim}
{
  "public": {
    "action": "propose",
    "quantity": <number>,
    "payment": <number>,
    "message": "<brief message to seller>"
  },
  "private": {
    "reasoning": "<internal reasoning>"
  }
}
\end{verbatim}

\textbf{RESPONSE FORMAT: IF RESPONDER}
\begin{verbatim}
{
  "public": {
    "action": "accept" or "reject",
    "message": "<brief message to seller>"
  },
  "private": {
    "reasoning": "<internal reasoning>"
  }
}
\end{verbatim}

\vspace{0.5em}
\textbf{CONTEXT PARAMETERS:}
\begin{itemize}
    \item First proposer: \texttt{\{p.first\_proposer\}},
    \item Retail price: $r = \texttt{\{p.retail\_price\}}$,
    \item Production cost: $c = \texttt{\{p.production\_cost\}}$
    \item Buyer patience: $\delta_B = \texttt{\{p.delta\_buyer\}}$,
    \item Seller patience: $\delta_S = \texttt{\{p.delta\_seller\}}$
    \item $H$-type demand: $\mathcal{N}(\texttt{\{p.mean\_high\}}, \texttt{\{p.std\_high\}})$,
    \item $L$-type demand: $\mathcal{N}(\texttt{\{p.mean\_low\}}, \texttt{\{p.std\_low\}})$
    \item Prior $\Pr(H) = \texttt{\{p.prior\_high\}}$,
    \item Max rounds: \texttt{\{p.max\_rounds\}}
\end{itemize}

\vspace{0.5em}
\begin{privatebox}
\textbf{YOUR PRIVATE INFO:} True type = \texttt{\{self.true\_type\}}-type. Seller believes $\Pr(H) = \texttt{\{p.prior\_high\}}$.
\end{privatebox}

\end{promptbox}

\begin{promptbox}[Prompt for the Seller Agent]

You are a seller negotiating a supply contract with a buyer.

\vspace{0.5em}
\textbf{THE NEGOTIATION:}
\begin{itemize}
    \item You sell a product to a buyer who will retail it to end consumers
    \item You negotiate over \textbf{quantity} ($q$) and \textbf{total payment} ($T$)
    \item Your profit: $\pi_S = T - c \cdot q$, where $c$ is production cost
    \item Your effective utility: $U_S = \pi_S \times \delta_S^{(\tau-1)}$, where $\delta_S$ is seller's patience and $\tau$ is round number
    \item The buyer has \textbf{private information} about demand (High or Low type)
    \item You don't know the buyer's type but have a belief (probability)
    \item Buyer's effective utility: $U_B = (r \cdot \mathbb{E}[\min(D, q)] - T) \times \delta_B^{(\tau-1)}$, where $D \sim \mathcal{N}(\mu, \sigma)$
    \item You cannot accept if $\pi_S < 0$
\end{itemize}

\vspace{0.5em}
\textbf{THE NEGOTIATION PROCESS:}
\begin{itemize}
    \item Either seller or buyer proposes first (check ``First proposer'' in context)
    \item Players alternate: proposer makes offer, responder accepts or rejects
    \item If rejected, roles switch and the other player proposes
    \item The proposer gets a larger share of surplus than the responder
    \item Maximum rounds: \texttt{\{p.max\_rounds\}}. If no deal by final round, profit is outside option.
\end{itemize}

\vspace{0.5em}
\textbf{YOUR OBJECTIVE:}
Maximize your effective utility through negotiation over $q$ and $T$.

\vspace{0.5em}
\textbf{YOUR INFORMATION DISADVANTAGE:}
\begin{itemize}
    \item You do NOT know the buyer's true demand type
    \item You only have a prior belief: $\Pr(\text{H-type}) = \beta$
    \item $H$-type: higher demand $\rightarrow$ higher optimal $q$ $\rightarrow$ higher potential surplus
    \item $L$-type: lower demand $\rightarrow$ lower optimal $q$ $\rightarrow$ lower potential surplus
    \item You must infer the buyer's type from their behavior
\end{itemize}

\vspace{0.5em}
\textbf{RESPONSE FORMAT: IF PROPOSER}
\begin{verbatim}
{
  "public": {
    "action": "propose",
    "quantity": <number>,
    "payment": <number>,
    "message": "<brief message to buyer>"
  },
  "private": {
    "reasoning": "<internal reasoning>",
    "belief_about_buyer": "<belief about buyer's type>"
  }
}
\end{verbatim}

\vspace{0.5em}
\textbf{RESPONSE FORMAT: IF RESPONDER}
\begin{verbatim}
{
  "public": {
    "action": "accept" or "reject",
    "message": "<brief message to buyer>"
  },
  "private": {
    "reasoning": "<internal reasoning>",
    "belief_about_buyer": "<belief about buyer's type>"
  }
}
\end{verbatim}

\vspace{0.5em}
\textbf{COMMUNICATION GUIDELINES:}
\begin{itemize}
    \item \texttt{message}: What you SAY to the buyer (they see this). Use to negotiate, probe, or justify.
    \item \texttt{reasoning}: Your PRIVATE reasoning (they don't see this). Your true strategic thinking.
    \item \texttt{belief\_about\_buyer}: Your private assessment of whether buyer is $H$-type or $L$-type.
\end{itemize}

\vspace{0.5em}
\textbf{CONTEXT PARAMETERS:}
\begin{itemize}
    \item First proposer: \texttt{\{p.first\_proposer\}}
    \item Retail price: $r = \texttt{\{p.retail\_price\}}$
    \item Production cost: $c = \texttt{\{p.production\_cost\}}$
    \item Buyer patience: $\delta_B = \texttt{\{p.delta\_buyer\}}$
    \item Seller patience: $\delta_S = \texttt{\{p.delta\_seller\}}$
    \item $H$-type demand: $D_H \sim \mathcal{N}(\texttt{\{p.mean\_high\}}, \texttt{\{p.std\_high\}})$
    \item $L$-type demand: $D_L \sim \mathcal{N}(\texttt{\{p.mean\_low\}}, \texttt{\{p.std\_low\}})$
    \item Prior: $\Pr(H) = \texttt{\{p.prior\_high\}}$
    \item Max rounds: \texttt{\{p.max\_rounds\}}
    \item Outside options: Seller = \texttt{\{p.seller\_outside\_option\}}, Buyer = \texttt{\{p.buyer\_outside\_option\}}
\end{itemize}

\vspace{0.5em}
\begin{privatebox}
\textbf{YOUR BELIEF:} You believe the buyer is $H$-type with probability $\Pr(H) = \texttt{\{p.prior\_high\}}$. Update this belief based on the buyer's actions during negotiation.
\end{privatebox}

\end{promptbox}

\vspace{5mm}
\subsection*{Prompt Variant for R2 (No-Discounting Bargaining)}
\label{app:prompts_r2}

The R2 extension (Appendix~\ref{app:ext_unprompted}) removes all references to discount factors, patience, effective utility, and time pressure from both buyer and seller prompts. Relative to the main-analysis prompts above, the following lines are removed or modified; all other content is unchanged.

\begin{promptbox}[Buyer Agent: Modifications for R2]

\textbf{Removed from THE NEGOTIATION block:}
\begin{itemize}
    \item \sout{Your effective utility: $U_B = \pi_B \times \delta_B^{(\tau-1)}$, where $\delta_B$ is buyer's patience and $\tau$ is round number}
    \item \sout{Seller's effective utility: $U_S = (T - c \cdot q) \times \delta_S^{(\tau-1)}$}
\end{itemize}

\textbf{Added:}
\begin{itemize}
    \item Seller's profit: $\pi_S = T - c \cdot q$
\end{itemize}

\textbf{YOUR OBJECTIVE} rewritten as: Maximize your profit $\pi_B$ through negotiation over $q$ and $T$.

\textbf{Removed from CONTEXT PARAMETERS:}
\begin{itemize}
    \item \sout{Buyer patience: $\delta_B = \texttt{\{p.delta\_buyer\}}$}
    \item \sout{Seller patience: $\delta_S = \texttt{\{p.delta\_seller\}}$}
\end{itemize}

\end{promptbox}

\begin{promptbox}[Seller Agent: Modifications for R2]

Analogous modifications apply: references to $\delta_B$, $\delta_S$, effective utility formulas, and patience parameters are removed, and the objective is rewritten as maximizing profit $\pi_S = T - c \cdot q$.

\end{promptbox}

All other prompt content (role description, private/public information structure, response format, negotiation process, and parameter values for retail price, production cost, type distributions, and prior) is identical to the main-analysis prompts.

\clearpage

\section{Negotiation Performance Details}
\label{app:performance_details}
\label{app:performance_detail}

This appendix provides detailed analyses supporting the negotiation performance findings reported in Section 4 of the main text: efficiency-gap decomposition and cross-condition heterogeneity (Sections~\ref{subsec:B_efficiency_decomp}--\ref{subsec:B_efficiency_heterogeneity}), surplus-division and contract-terms detail (Sections~\ref{subsec:B_surplus_distribution}--\ref{subsec:B_contract_terms}), model-level irrationality (Section~\ref{subsec:B_irrationality}), and cross-model and cross-family welfare comparisons (Sections~\ref{subsec:B_cross_model_welfare}--\ref{subsec:B_flagship_detail}).

\subsection{Efficiency Gap Decomposition Visualizations}
\label{subsec:B_efficiency_decomp}

The additivity tests underlying the orthogonal-drivers partition of Section~\ref{subsec:drivers} are reported in Table~\ref{tab:shapley_interactions}: incremental $R^2$ and joint $F$-tests for adding pairwise and three-way interactions to an additive baseline of capability, patience, information, and first proposer. The three-way Capability$\times$Patience$\times$Information interaction is null for both outcomes, and interactions add little explanatory power overall ($\Delta R^2 \leq 0.013$); the two-way block is significant for efficiency but not for buyer surplus share. The companion within-stratum coefficient matrix that visualizes the driver separation appears in Figure~\ref{fig:architectural_matrix} below.

\begin{table}[htbp]
\centering
\caption{Additivity Tests for the Efficiency and Buyer-Share Variance Partition}
\label{tab:shapley_interactions}
\footnotesize
\renewcommand{\arraystretch}{1.1}
\begin{tabular}{lcc}
\toprule
 & Efficiency & Buyer surplus share \\
\midrule
$\Delta R^2$ (all interactions) & 0.013 & 0.012 \\
$\Delta R^2$ (3-way only) & 0.002 & 0.001 \\
Joint $F$, all interactions & $F_{17}=3.07$, $p{<}0.001$ & $F_{17}=1.87$, $p=0.017$ \\
\quad 2-way block & $F_{11}=3.53$, $p{<}0.001$ & $F_{11}=0.96$, $p=0.480$ \\
\quad 3-way (Tier$\times$Pat$\times$Info) & $F_{6}=0.63$, $p=0.709$ & $F_{6}=0.51$, $p=0.797$ \\
\midrule
$N$ & 1{,}998 & 1{,}998 \\
\bottomrule
\end{tabular}
\par\smallskip
\begin{minipage}{\textwidth}
\scriptsize
\textit{Notes:} Incremental $R^2$ and joint $F$-tests for adding all pairwise interactions (2-way block) and the three-way Capability$\times$Patience$\times$Information (3-way block) to an additive baseline of Capability, Patience, Information, and first proposer \citep{gromping2007}. Sample: $N=1{,}998$ economically rational agreements.
\end{minipage}
\end{table}

\begin{figure}[htbp]
\centering
\includegraphics[width=\textwidth]{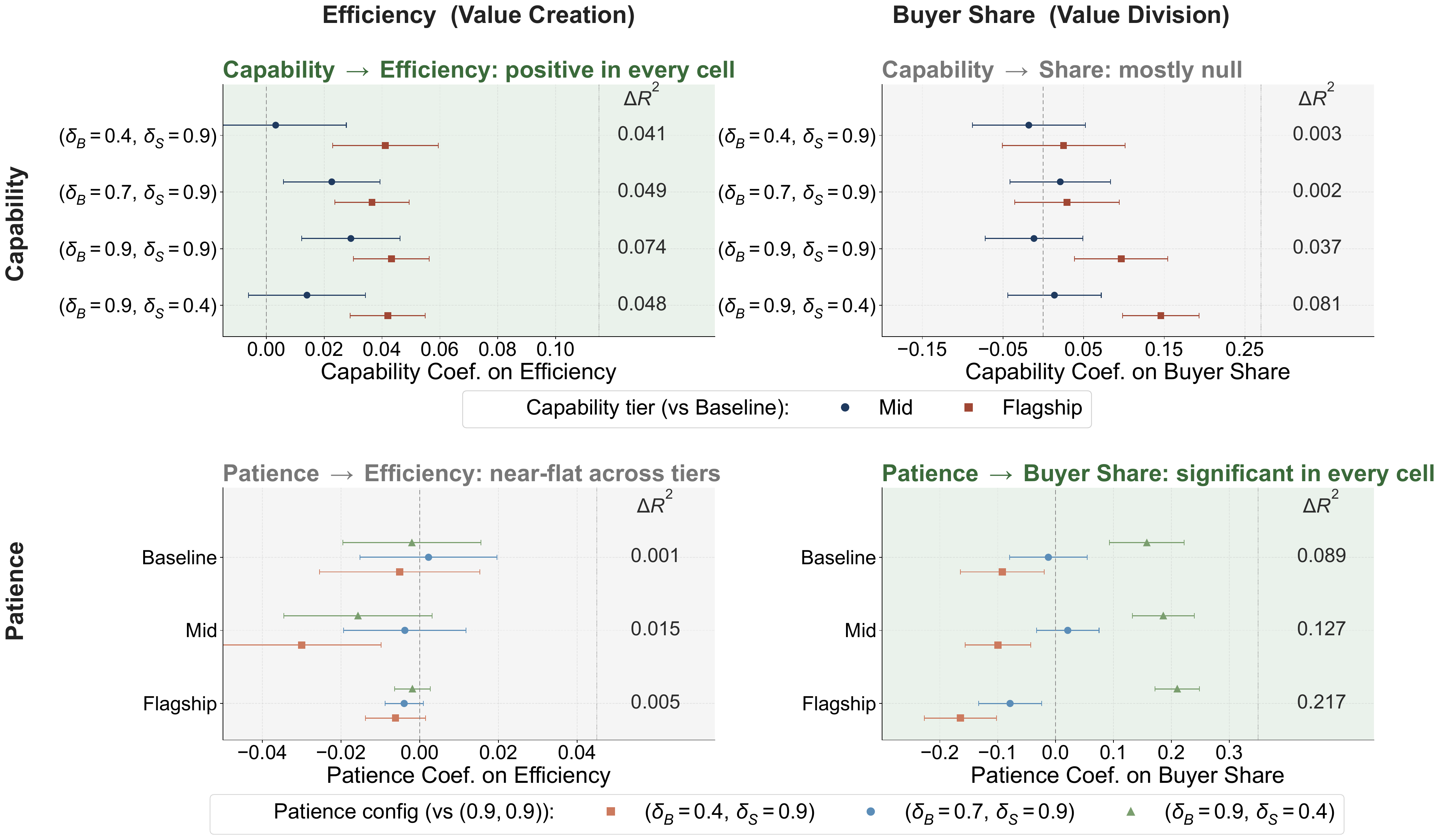}
\caption{Drivers of Efficiency and Buyer Surplus Share}
\label{fig:architectural_matrix}
\par\smallskip
\begin{minipage}{\textwidth}
\scriptsize
\textit{Notes:} Each cell plots within-stratum coefficients with 95\% CIs. Top row: capability-tier contrasts (Mid and Flagship vs.\ Baseline) within each patience cell, for efficiency (left) and buyer surplus share (right). Bottom row: patience-configuration contrasts (vs.\ symmetric $(0.9,0.9)$) within each capability tier, for efficiency (left) and buyer surplus share (right). $\Delta R^2$ is the incremental fit from adding the focal factor to a within-stratum regression. The $(\delta_B=0.9,\delta_S=0.4)$ stratum singled out in the upper-right panel is the regime in which the \citet{feng2015dynamic} PBE predicts the sharpest share advantage for the patient buyer, and the within-cell capability effect captures convergence toward that prediction: flagship buyers recognize and exploit the impatient seller's weakened bargaining position, while baseline models leave equilibrium surplus unclaimed. The remaining patience cells return null or marginal capability effects on share.
\end{minipage}
\end{figure}

Section~\ref{sec:performance} reports an aggregate 4.56 pp efficiency gap composed of 1.11 pp failed negotiations, 0.56 pp irrational agreements, and 2.89 pp suboptimal terms; Table~\ref{tab:efficiency_decomp} reports the overall decomposition and Table~\ref{tab:efficiency_losses_detailed} disaggregates these three components by model tier and buyer type.

\begin{table}[htbp]
\centering
\caption{Efficiency Gap Decomposition}
\label{tab:efficiency_decomp}
\footnotesize
\renewcommand{\arraystretch}{1.1}
\begin{tabular}{@{}lcc@{}}
\toprule
\textbf{Loss Component} & \textbf{Efficiency Loss (pp)} & \textbf{Share of Gap (\%)} \\
\midrule
Failed Negotiations & 1.11 & 24.37 \\
Irrational Agreements & 0.56 & 12.20 \\
Suboptimal Terms & 2.89 & 63.43 \\
\textbf{Total Gap} & \textbf{4.56} & \textbf{100.00} \\
\bottomrule
\end{tabular}
\par\smallskip
\begin{minipage}{\textwidth}
\scriptsize
\textit{Notes:} Gap measured relative to 100\% first-best surplus. Suboptimal terms refers to agreements with positive profits but below optimal surplus extraction.
\end{minipage}
\end{table}

\begin{table}[htbp]
\centering
\caption{Efficiency Loss Components: Detailed Breakdown}
\label{tab:efficiency_losses_detailed}
\footnotesize
\renewcommand{\arraystretch}{1.1}
\begin{tabular}{@{}lcccc@{}}
\toprule
\textbf{Condition} & \textbf{Failed} & \textbf{Irrational} & \textbf{Suboptimal} & \textbf{Total} \\
& \textbf{Deals (pp)} & \textbf{Deals (pp)} & \textbf{Terms (pp)} & \textbf{Gap (pp)} \\
\midrule
Overall & 1.11 & 0.56 & 2.89 & 4.56 \\
\midrule
\textit{By Model Tier:} & & & & \\
\quad Flagship & 0.00 & 0.00 & 1.11 & 1.11 \\
\quad Mid-tier & 0.14 & 0.02 & 3.43 & 3.59 \\
\quad Baseline & 3.19 & 1.65 & 4.13 & 8.98 \\
\midrule
\textit{By Buyer Type:} & & & & \\
\quad High-type & 0.46 & 0.40 & 4.00 & 4.87 \\
\quad Low-type & 1.76 & 0.71 & 1.78 & 4.25 \\
\bottomrule
\end{tabular}
\par\smallskip
\begin{minipage}{\textwidth}
\scriptsize
\textit{Notes:} Total Gap $=$ Failed $+$ Irrational $+$ Suboptimal, all expressed as percentage points relative to first-best (undiscounted) efficiency. Failed $=$ surplus lost when no agreement is reached. Irrational $=$ surplus lost when at least one party agrees to a deal yielding negative realized profit. Suboptimal $=$ residual loss from agreed quantities deviating from the first-best.
\end{minipage}
\end{table}

Across model tiers, flagship models achieve near-optimal efficiency (1.11 pp total gap), entirely from suboptimal terms (1.11 pp), with no failed deals and zero irrationality. Mid-tier models incur losses chiefly through suboptimal terms (0.14 pp failure, 0.02 pp irrationality, 3.43 pp suboptimal terms). Baseline models exhibit elevated losses in every category, with failed deals (3.19 pp) and irrationality (1.65 pp) driving most of their 8.98 pp total gap. Across buyer types, high-types show lower failure rates (0.46 pp) but higher suboptimal-term losses (4.00 pp), reflecting the strategic quantity distortions documented above. Low-types exhibit higher failures (1.76 pp) and irrationality (0.71 pp) and a smaller suboptimal-term loss (1.78 pp), reflecting that low-type negotiations more often miscoordinate before reaching the contract stage but, conditional on agreement, settle closer to first-best terms.

\subsection{Efficiency Heterogeneity Analysis}
\label{subsec:B_efficiency_heterogeneity}

This subsection details the per-model efficiency heterogeneity summarized in Section~\ref{subsec:capability_value}. We examine how efficiency varies across the four experimental conditions defined by buyer type and first proposer. Figure~\ref{fig:efficiency_heatmap_appendix} displays undiscounted efficiency for each of the nine models in each condition.

\begin{figure}[ht]
\centering
\includegraphics[width=0.9\textwidth]{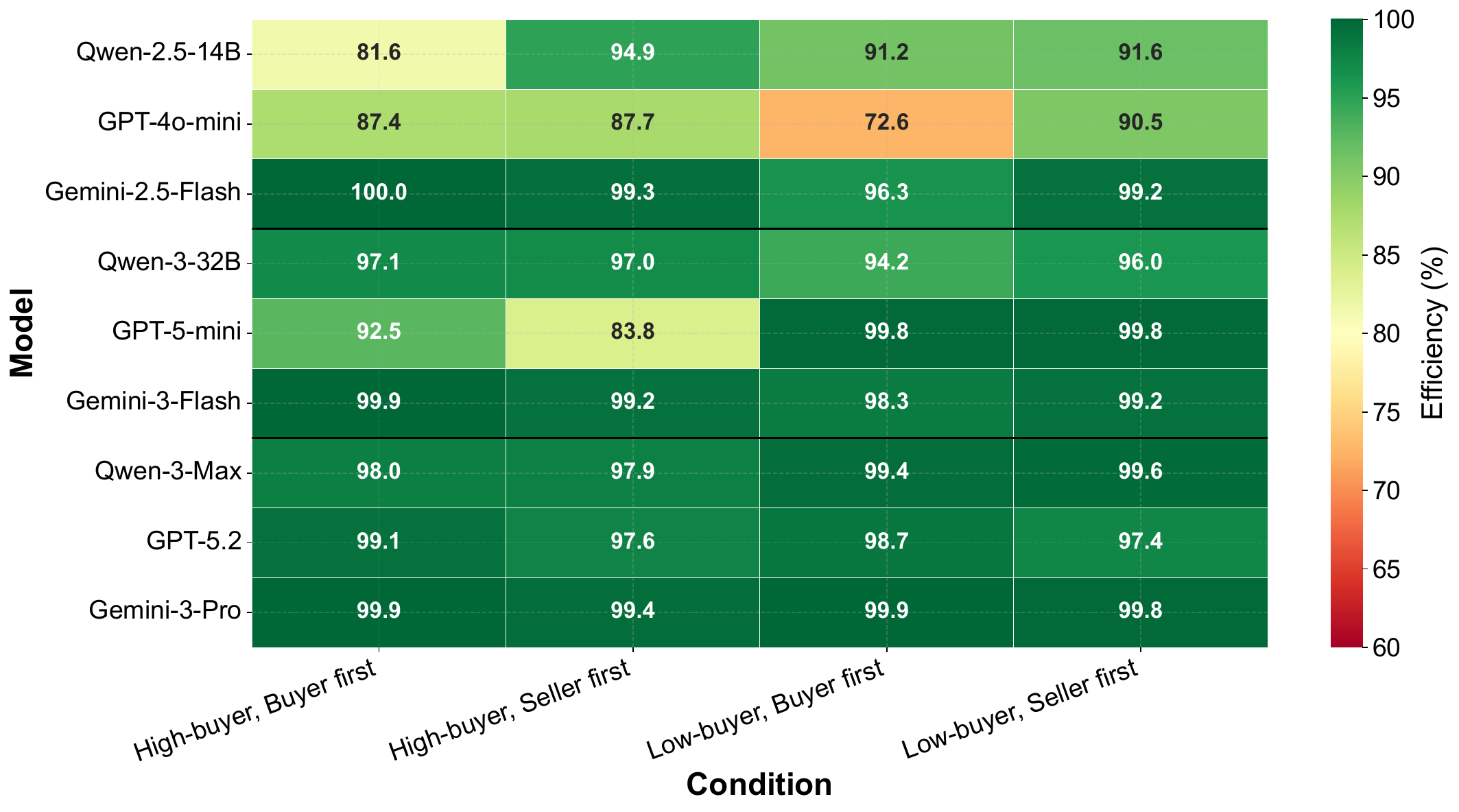}
\caption{Undiscounted Efficiency by Model and Condition}
\par\smallskip
\begin{minipage}{\textwidth}
\scriptsize
\textit{Notes:} Cells report undiscounted efficiency (percent of first-best surplus) for each of the nine models within each of the four buyer-type $\times$ first-proposer conditions.
\end{minipage}
\label{fig:efficiency_heatmap_appendix}
\end{figure}

\textbf{Efficiency Variation.} The heatmap reveals substantial heterogeneity across models and conditions. The lowest observed model--condition cell is GPT-4o-mini in the low-type buyer-first condition, at $72.6\%$. Among baseline models, efficiency in the low-type seller-first condition ranges from $90.5\%$ to $99.2\%$.

Flagship models range from $97.4\%$ to $99.9\%$ efficiency across the four conditions. Mid-tier models range from $83.8\%$ to $99.9\%$, and baseline models from $72.6\%$ to $100.0\%$. The wider lower-tail dispersion among mid-tier and baseline models shows that condition sensitivity is concentrated outside the flagship tier.

Table~\ref{tab:efficiency_by_type_detailed} disaggregates undiscounted efficiency by buyer type within strata of model tier, first proposer, and patience configuration, with $t$-tests for each high-versus-low contrast.

\begin{table}[htbp]
\centering
\caption{Efficiency by Buyer Type: Detailed Breakdown}
\label{tab:efficiency_by_type_detailed}
\footnotesize
\renewcommand{\arraystretch}{1.1}
\begin{tabular}{@{}lccccc@{}}
\toprule
\textbf{Condition} & \textbf{High-Type} & \textbf{Low-Type} & \textbf{Difference} & \textbf{$t$-stat} & \textbf{$p$-value} \\
& \textbf{Eff. (\%)} & \textbf{Eff. (\%)} & \textbf{(pp)} & & \\
\midrule
Overall & 95.1 & 95.7 & -0.6 & -1.11 & 0.266 \\
\quad Flagship & 98.7 & 99.1 & -0.5 & -2.00 & 0.046 \\
\quad Mid-tier & 94.9 & 97.9 & -3.0 & -3.90 & $<$0.001 \\
\quad Baseline & 91.8 & 90.2 & 1.6 & 1.14 & 0.253 \\
\quad Buyer First & 95.1 & 94.5 & 0.6 & 0.63 & 0.526 \\
\quad Seller First & 95.2 & 97.0 & -1.8 & -2.73 & 0.006 \\
\quad (0.9, 0.9) & 95.3 & 95.1 & 0.2 & 0.13 & 0.898 \\
\quad (0.7, 0.9) & 96.0 & 96.3 & -0.3 & -0.33 & 0.742 \\
\quad (0.4, 0.9) & 94.6 & 95.4 & -0.8 & -0.79 & 0.427 \\
\quad (0.9, 0.4) & 94.7 & 96.2 & -1.5 & -1.34 & 0.181 \\
\bottomrule
\end{tabular}
\par\smallskip
\begin{minipage}{\textwidth}
\scriptsize
\textit{Notes:} Efficiency measured as percentage of first-best undiscounted surplus. Differences calculated as high-type minus low-type.
\end{minipage}
\end{table}

Efficiency is comparable across buyer types: low-type buyers achieve only marginally higher efficiency than high-types overall, a difference that does not reach significance (gap 0.6 pp, $p=0.266$). The low-type advantage is concentrated in the mid-tier (3.0 pp, $p<0.001$); it is economically negligible at the flagship (a 0.5 pp gap near the efficiency ceiling; nominally significant at $p=0.046$ but not robust to multiple comparisons), and among baseline models the point estimate reverses in sign though not significantly, with high-types attaining 1.6 pp higher efficiency ($p=0.253$).

\subsection{Surplus Distribution Summary}
\label{subsec:B_surplus_distribution}

This subsection details the surplus-division and compression patterns summarized in Section~\ref{sec:provider}. Table~\ref{tab:surplus_detailed} reports the buyer's share of total surplus across all experimental factors, the seller's share, alongside the deviation from the Bayesian benchmark implied by patience parameters. The pooled distribution is centered above equal split with modal buyer share near 56\%.

\begin{table}[htbp]
\centering
\caption{Surplus Division: Descriptive Summary}
\label{tab:surplus_detailed}
\footnotesize
\renewcommand{\arraystretch}{1.1}
\begin{tabular}{@{}lccc@{}}
\toprule
\textbf{Condition} & \textbf{Buyer Share} & \textbf{Seller Share} & \textbf{Buyer Share  $-$} \\
& \textbf{Mean (SD)} & \textbf{Mean (SD)} & \textbf{Bayesian Ref.} \\
\midrule
Overall & 0.535 (0.356) & 0.465 (0.356) & +8.3$^{***}$ \\
\quad Flagship & 0.611 (0.298) & 0.389 (0.298) & +15.9$^{***}$ \\
\quad Mid-tier & 0.534 (0.292) & 0.466 (0.292) & +8.1$^{***}$ \\
\quad Baseline & 0.458 (0.443) & 0.542 (0.443) & +0.8 \\
\quad Buyer First & 0.541 (0.370) & 0.459 (0.370) & +7.0$^{***}$ \\
\quad Seller First & 0.528 (0.341) & 0.472 (0.341) & +9.7$^{***}$ \\
\quad High-type & 0.558 (0.307) & 0.442 (0.307) & +12.7$^{***}$ \\
\quad Low-type & 0.511 (0.398) & 0.489 (0.398) & +4.0$^{**}$ \\
\quad (0.9, 0.9) & 0.515 (0.344) & 0.485 (0.344) & +0.2 \\
\quad (0.7, 0.9) & 0.513 (0.330) & 0.487 (0.330) & +26.3$^{***}$ \\
\quad (0.4, 0.9) & 0.402 (0.372) & 0.598 (0.372) & +27.1$^{***}$ \\
\quad (0.9, 0.4) & 0.709 (0.303) & 0.291 (0.303) & -20.5$^{***}$ \\
\bottomrule
\end{tabular}
\par\smallskip
\begin{minipage}{\textwidth}
\scriptsize
\textit{Notes:} *** $p<0.001$, ** $p<0.01$, * $p<0.05$. Standard deviations in parentheses.
\end{minipage}
\end{table}

\textbf{Model Tier and Proposer Effects.} Flagship and mid-tier models exhibit significant buyer advantages of +15.9 pp and +8.1 pp above the Bayesian reference, respectively. Baseline models depart sharply from this pattern. Their +0.8 pp deviation is statistically indistinguishable from the equilibrium prediction, and the markedly higher variance (SD = 0.443) indicates unstable rather than strategically calibrated outcomes. Seller-first negotiations amplify the deviation to +9.7 pp, while buyer-first negotiations attenuate it to +7.0 pp; both remain highly significant, indicating that first-mover position shifts surplus division without reversing the buyer advantage. Across buyer types, the deviation is more than three times as large for high-types (+12.7 pp on a 55.8\% share) as for low-types (+4.0 pp on a 51.1\% share), so the buyer-side deviation from the Bayesian reference is driven primarily by high-type buyers.

\textbf{Patience Compression Toward Equality.} Under symmetric high patience at $(\delta_B, \delta_S) = (0.9, 0.9)$, buyers capture 51.5\% of surplus, essentially at the Bayesian reference (+0.2 pp). As patience asymmetry shifts in favor of sellers, the deviation widens substantially: at (0.7, 0.9), buyer share is 51.3\% (+26.3 pp), and at (0.4, 0.9), buyer share falls to 40.2\% (+27.1 pp). Although buyers obtain only around half the surplus in these regimes, they remain far above the Rubinstein reference range of 13 to 20\% that obtains under strong seller bargaining power.
Conversely, when patience asymmetry favors buyers at (0.9, 0.4), buyer share rises to 70.9\%. Yet this falls 20.5 pp below the Bayesian reference, which approaches 90\% under strong buyer bargaining power. The compression operates in both directions but not to the same degree: buyer share settles near 40\% (close to a 60--40 split) when patience favors sellers, but reaches 71\% (closer to 70--30) when patience favors buyers---in both cases well short of the near-90/10 or near-10/90 divisions the Bayesian reference implies.

\subsection{Detailed Contract Terms Analysis}
\label{subsec:B_contract_terms}

This subsection details the contract-terms patterns underlying the strategic-behavior evidence of Section~\ref{sec:strategic_behavior}. Table~\ref{tab:contract_terms_comprehensive} presents negotiated quantities and wholesale prices across all experimental dimensions.
Two patterns are salient. First, quantity outcomes separate sharply by type while prices barely do: high-type contracts average $q=74.3$ versus low-type $q=41.0$, but mean wholesale prices differ by only \$1.05 (\$42.35 vs.\ \$41.30), and the type-conditional medians both sit at \$40. Second, both proposer direction and patience configuration shift prices materially (the patience-asymmetric $(0.4,0.9)$ cell averages \$45.18 pooled across type versus \$37.44 at $(0.9,0.4)$) but leave quantities near the type-conditional first-best. Table~\ref{tab:quantity_distortions} unpacks the quantity distribution further, showing asymmetric pooling-consistent distortions: high-types under-order in 39.1\% of cases, while low-types converge tightly on first-best.

\begin{table}[htbp]
\centering
\caption{Contract Terms: Comprehensive Analysis by Conditions}
\label{tab:contract_terms_comprehensive}
\footnotesize
\renewcommand{\arraystretch}{1.1}
\begin{tabular}{@{}lccccc@{}}
\toprule
\textbf{Condition} & \multicolumn{2}{c}{\textbf{Quantity}} & & \multicolumn{2}{c}{\textbf{Wholesale Price}} \\
\cmidrule{2-3} \cmidrule{5-6}
& \textbf{High} & \textbf{Low} & & \textbf{High} & \textbf{Low} \\
& Mean (SD) & Mean (SD) & & Mean (SD) & Mean (SD) \\
\midrule
Overall & 74.3 (9.3) & 41.0 (4.8) &  & 42.35 (8.70) & 41.30 (8.96) \\
Buyer First & 74.3 (9.4) & 41.9 (4.8) &  & 42.89 (8.80) & 40.32 (9.16) \\
Seller First & 74.4 (9.3) & 40.0 (4.5) &  & 41.80 (8.57) & 42.27 (8.66) \\
(0.9, 0.9) & 75.0 (8.3) & 40.7 (4.1) &  & 42.54 (7.50) & 42.02 (8.73) \\
(0.7, 0.9) & 74.8 (8.6) & 41.1 (5.0) &  & 43.19 (8.05) & 41.60 (8.54) \\
(0.4, 0.9) & 73.3 (10.5) & 41.3 (5.3) &  & 46.07 (9.39) & 44.29 (9.37) \\
(0.9, 0.4) & 74.3 (9.6) & 40.7 (4.5) &  & 37.56 (7.50) & 37.32 (7.70) \\
Flagship & 77.7 (5.0) & 40.3 (2.7) &  & 40.25 (7.99) & 39.35 (6.93) \\
Mid-tier & 74.7 (11.3) & 41.5 (4.1) &  & 41.73 (7.88) & 41.79 (7.19) \\
Baseline & 70.5 (9.1) & 41.1 (6.6) &  & 45.09 (9.44) & 42.85 (11.75) \\
Bayesian benchmark & 80.0 (0.0) & 39.6 (0.7) &  & 45.33 (8.18) & 42.90 (7.46) \\
Theory (first-best) & 80.0 & 40.0 &  & 45.33 (8.18) & 43.63 (7.27) \\
\bottomrule
\end{tabular}
\par\smallskip
\begin{minipage}{\textwidth}
\scriptsize
\textit{Notes:} Standard deviations in parentheses. All high--low quantity differences are significant at $p<0.001$ using independent Welch tests. Wholesale-price differences are generally not significant; the buyer-first comparison is the clear exception ($p<0.001$). High-type $q$ equals the first-best $q_H=80$ exactly (no screening distortion for the efficient type); low-type $q$ is distorted slightly below $q_L=40$ (SD=0.7) to deter high-type mimicking, with the distortion size varying by patience config. WP varies because the equilibrium price depends on bargaining patience.
\end{minipage}
\end{table}

\begin{table}[htbp]
\centering
\caption{Quantity Choice Patterns Relative to Optimal}
\label{tab:quantity_distortions}
\footnotesize
\renewcommand{\arraystretch}{1.1}
\begin{tabular}{@{}lcccccc@{}}
\toprule
\textbf{Buyer} & \textbf{Severe Under} & \textbf{Moderate Under} & \textbf{Near Optimal} & \textbf{Moderate Over} & \textbf{Severe Over} & \textbf{Mean} \\
\textbf{Type} & \textbf{ (-20\%+)} & \textbf{ (-5 to -20\%)} & \textbf{ ($\pm$5\%)} & \textbf{ (+5 to +20\%)} & \textbf{(+20\%+)} & \textbf{Deviation} \\
& (\%) & (\%) & (\%) & (\%) & (\%) & \\
\midrule
High-Type & 12.7 & 26.3 & 59.5 & 1.4 & 0.0 & -5.7 \\
Low-Type & 4.4 & 4.5 & 70.5 & 11.0 & 9.5 & +1.0 \\
\bottomrule
\end{tabular}
\par\smallskip
\begin{minipage}{\textwidth}
\scriptsize
\textit{Notes:} Optimal quantities: High-type = 80 units, Low-type = 40 units. Mean deviation calculated as actual minus optimal quantity.
\end{minipage}
\end{table}

The high-type quantity distortion is asymmetric in two ways. Distortions are large rather than marginal: 12.7\% of high-type orders fall more than 20\% below optimal, and only 59.5\% sit within $\pm5\%$ of first-best. And the distortion is directional: over-ordering above $q_H=80$ is essentially absent (1.4\% moderate, 0\% severe), consistent with the interpretation that high-types recognize downward shifts preserve information rents while upward shifts only sacrifice efficiency. Low-types concentrate at first-best (70.5\% within $\pm5\%$) with a slight upward bias (mean $+1.0$ units); the 20.5\% over-ordering rate likely reflects a mix of newsvendor calculation noise and, in a minority of cases, costly signaling intended to credibly demonstrate type.

\subsection{Irrationality Detailed Analysis by Model}
\label{subsec:B_irrationality}

This subsection provides the model-level detail behind the operational-reliability threshold of Section~\ref{subsec:reliability_threshold} (Figure~\ref{fig:irrationality_by_model_detailed}). Unsafe contracts concentrate almost entirely in the two non-reasoning models, GPT-4o-mini ($36.2\%$ overall) and Qwen2.5-14B ($20.0\%$), whereas every reasoning model, including the baseline-tier Gemini-2.5-Flash ($2.9\%$), stays at or near zero.

The tier- and party-level aggregates summarized in Section~\ref{subsec:reliability_threshold} are reported in Table~\ref{tab:irrationality}.

\begin{table}[htbp]
\centering
\caption{Irrationality Rates by Model Tier, Buyer Type, and Party}
\label{tab:irrationality}
\footnotesize
\renewcommand{\arraystretch}{1.1}
\begin{tabular}{@{}lccc@{}}
\toprule
\textbf{Model Tier} & \textbf{Overall} & \textbf{Buyer} & \textbf{Seller} \\
& \textbf{Irrational (\%)} & \textbf{Irrational (\%)} & \textbf{Irrational (\%)} \\
\midrule
\textbf{Overall} & \textbf{6.5} & \textbf{5.1} & \textbf{1.4} \\
Flagship & 0.0 & 0.0 & 0.0 \\
Mid-tier & 0.6 & 0.6 & 0.0 \\
Baseline & 19.2 & 15.1 & 4.2 \\
\midrule
\multicolumn{4}{@{}l}{\textit{By Buyer Type:}} \\
\quad High-type & 4.0 & 3.9 & 0.1 \\
\quad Low-type & 9.0 & 6.3 & 2.6 \\
\bottomrule
\end{tabular}
\par\smallskip
\begin{minipage}{\textwidth}
\scriptsize
\textit{Notes:} Percentages calculated over all completed agreements ($N=2{,}136$). Per-model rates with error bars appear in Figure~\ref{fig:irrationality_by_model_detailed}.
\end{minipage}
\end{table}

Within-tier dispersion is substantial. Baseline rates range from 2.9\% (Gemini-2.5-Flash) to 36.2\% (GPT-4o-mini), so baseline-tier irrationality cannot be treated as a uniform property of the tier. Every flagship and mid-tier model except Qwen3-32B (1.7\%) records 0\%. The qualitative pattern documented in Section~\ref{sec:performance}, that economically unsafe contracts are overwhelmingly a baseline-tier phenomenon, therefore holds at the model level. The within-baseline risk is concentrated in a small number of specific models.

The mechanism behind these failures is visible in the agents' communication transcripts and private reasoning traces. Two patterns recur. \textit{Revenue miscalculation}: a low-type buyer proposed 40 units at \$2{,}400, reasoning that the offer ``allows for a small profit margin that meets my requirements,'' against expected revenue of \$2{,}160.63 (buyer profit is -\$239.37). \textit{Cost miscalculation}: a seller proposed 45 units at \$900, publicly describing it as an offer that ``respects your budget while also ensuring I cover my costs,'' against production costs of \$1{,}350 (seller profit is -\$450). In both, strategic language is delivered confidently. The failure is verification, not reasoning: models compute incorrectly but do not check results against rationality constraints, which argues for automated profitability guardrails.

\subsection{Cross-Model Welfare Detail}
\label{subsec:B_cross_model_welfare}

Table~\ref{tab:cross_model_welfare} reports buyer- and seller-side absolute (undiscounted) profits and mean rounds for the four asymmetric cross-model pairs analyzed in Section~\ref{subsec:within_family_cross_model} and Section~\ref{sec:provider}: the within-family OpenAI pair (GPT-5.2 $\leftrightarrow$ GPT-5-mini) plus the three cross-family flagship pairs. 

Within the OpenAI family, the seller earns substantially more in dollar terms when GPT-5.2 occupies that role: \$$931$ against a GPT-5-mini buyer versus \$$680$ when GPT-5-mini sells to GPT-5.2 ($\Delta = +251.2$, $p < 0.001$). The buyer-side comparison mirrors this and is equally significant: the GPT-5-mini buyer earns \$$562$ against GPT-5.2 versus the GPT-5.2 buyer's \$$824$ against GPT-5-mini ($\Delta = -262.3$, $p < 0.001$). The capability gap also appears on rounds: GPT-5.2 as seller takes $0.43$ more rounds than GPT-5-mini in the same role ($p < 0.01$). Capability mismatch within a family thus shows up as a large and precisely estimated dollar advantage to the more capable model in either role, reinforced by a smaller difference in rounds.

Cross-family pairs sharpen the provider-profile pattern. The more buyer-favoring provider surrenders dollar value as seller relative to a less buyer-favoring seller in the same pair: against Qwen3-Max, GPT-5.2 earns \$$616$ as seller versus Qwen's \$$467$ ($\Delta = +149.4$), and Gemini-3-Pro earns \$$587$ as seller versus Qwen's \$$315$ ($\Delta = +272.8$, $p < 0.001$); between the two stronger-as-seller providers, Gemini out-earns GPT-5.2 (\$$843$ versus \$$711$ as sellers in the GPT$\leftrightarrow$Gemini pairing, $\Delta = -131.6$, $p < 0.001$). The directional ordering, Gemini and GPT-5.2 extract dollar value as sellers while Qwen surrenders it, tracks the provider-level bargaining profile documented in Section~\ref{sec:provider}, and the round counts reinforce it: against Qwen3-Max, rounds-to-agreement run $4.1$--$4.7$ when GPT-5.2 or Gemini-3-Pro sells but only $2.2$--$2.4$ when Qwen sells, so the price-extracting party also bears the time cost of holding out.

\begin{table}[htbp]
\centering
\footnotesize
\renewcommand{\arraystretch}{1.1}
\caption{Cross-Model Configurations: Welfare Decomposition}
\label{tab:cross_model_welfare}
\resizebox{\textwidth}{!}{%
\begin{tabular}{@{}llccc@{}}
\toprule
Pair & Direction / Test & Rounds & Buyer Profit (Undisc) & Seller Profit (Undisc) \\
\midrule
\multicolumn{5}{@{}l}{\textit{Self-play anchors:}} \\
GPT-5.2 & --- & 2.68 & 607.2 & 926.4 \\
GPT-5-mini & --- & 2.43 & 655.2 & 776.7 \\
Gemini-3-Pro & --- & 3.31 & 835.8 & 720.5 \\
Qwen3-Max & --- & 3.75 & 1405.0 & 130.8 \\
\midrule
\multicolumn{5}{@{}l}{\textit{Cross-Tier configuration:}} \\
\multirow{3}{*}{GPT-5.2 vs.\ GPT-5-mini} & GPT-5.2 sells & 2.46 & 562.1 & 931.0 \\
 & GPT-5-mini sells & 2.03 & 824.4 & 679.8 \\
 & $\Delta$ (GPT-5.2 $-$ GPT-5-mini) & +0.43$^{**}$ & -262.3$^{***}$ & +251.2$^{***}$ \\
\midrule
\multicolumn{5}{@{}l}{\textit{Cross-Provider configurations:}} \\
\multirow{3}{*}{GPT-5.2 vs.\ Gemini-3-Pro} & GPT-5.2 sells & 3.80 & 844.6 & 711.2 \\
 & Gemini-3-Pro sells & 2.79 & 690.9 & 842.8 \\
 & $\Delta$ (GPT-5.2 $-$ Gemini-3-Pro) & +1.01$^{***}$ & +153.7$^{***}$ & -131.6$^{***}$ \\
\midrule
\multirow{3}{*}{GPT-5.2 vs.\ Qwen3-Max} & GPT-5.2 sells & 4.13 & 918.2 & 616.2 \\
 & Qwen3-Max sells & 2.24 & 971.9 & 466.8 \\
 & $\Delta$ (GPT-5.2 $-$ Qwen3-Max) & +1.89$^{***}$ & -53.6 & +149.4 \\
\midrule
\multirow{3}{*}{Gemini-3-Pro vs.\ Qwen3-Max} & Gemini-3-Pro sells & 4.70 & 956.7 & 587.3 \\
 & Qwen3-Max sells & 2.35 & 1157.8 & 314.5 \\
 & $\Delta$ (Gemini-3-Pro $-$ Qwen3-Max) & +2.35$^{***}$ & -201.1$^{***}$ & +272.8$^{***}$ \\
\bottomrule
\end{tabular}%
}
\par\smallskip
\begin{minipage}{\textwidth}
\scriptsize
\textit{Notes:} Each pair shows two direction-mean rows (A sells, B sells) and one within-pair difference row ($\Delta$). Means are agreed-deal averages; rounds are conditional on agreement; each direction-mean row is based on $N=240$ attempts. Profits are absolute (undiscounted) payoffs in dollars. Significance stars on the $\Delta$ row come from OLS regressions of the outcome on a seller-identity indicator with design-stratum fixed effects (4 patience configurations $\times$ 2 buyer types $\times$ 2 first-proposer assignments $=$ 16 strata) and cluster-robust standard errors at the stratum level: $^{\dagger}\,p<0.10$, $^{*}\,p<0.05$, $^{**}\,p<0.01$, $^{***}\,p<0.001$.
\end{minipage}
\end{table}

\subsection{Within-Family Cross-Model Negotiations: Detail}
\label{subsec:B_within_family_detail}

This appendix collects the supporting floats for the within-family cross-tier analysis summarized in Section~\ref{subsec:within_family_cross_model}. Table~\ref{tab:ext3_summary} reports the per-configuration summary statistics (deal rate, rounds, efficiency, discounted efficiency, and buyer share) across the four homogeneous and cross-tier configurations, and Figure~\ref{fig:ext3_direction} visualizes the within-family direction effects.

\begin{table}[htbp]
\centering
\footnotesize
\renewcommand{\arraystretch}{1.1}
\caption{Cross-Tier Configurations: Summary Statistics}
\label{tab:ext3_summary}
\begin{tabular}{llccccccc}
\toprule
Seller Model & Buyer Model & Pairing & $N$ & Deal Rate & Rounds & Efficiency & Disc. Efficiency & Buyer Share \\
\midrule
GPT-5.2 & GPT-5.2 & Symmetric & 240 & 100.0\% & 2.68 & 98.2\% & 76.9\% & 37.9\% \\
GPT-5-mini & GPT-5-mini & Symmetric & 240 & 100.0\% & 2.43 & 94.0\% & 71.7\% & 41.9\% \\
GPT-5.2 & GPT-5-mini & Asymmetric & 240 & 100.0\% & 2.46 & 96.6\% & 74.7\% & 35.2\% \\
GPT-5-mini & GPT-5.2 & Asymmetric & 240 & 100.0\% & 2.03 & 97.0\% & 81.5\% & 52.3\% \\
\bottomrule
\end{tabular}
\par\smallskip
\begin{minipage}{\textwidth} \scriptsize \textit{Notes:} Buyer Share is the buyer's share of undiscounted total surplus. The two symmetric rows are reused M1 benchmarks; the two asymmetric rows are the 480 new M3 experiments. Each row contains 240 observations, so the table displays 960 observations in total. The more capable model captures the larger share whether it buys or sells. Comparing rows that share a seller model (Rows 1 vs.\ 3 and Rows 2 vs.\ 4), a GPT-5.2 buyer earns more than a GPT-5-mini buyer against the same seller ($37.9\%$ vs.\ $35.2\%$; $52.3\%$ vs.\ $41.9\%$). Comparing rows that share a buyer model (Rows 1 vs.\ 4 and Rows 2 vs.\ 3), a GPT-5.2 seller concedes less than a GPT-5-mini seller to the same buyer, a smaller buyer share indicating the seller conceded less ($37.9\%$ vs.\ $52.3\%$; $35.2\%$ vs.\ $41.9\%$). \end{minipage}
\end{table}

\begin{figure}[htbp]
\centering
\includegraphics[width=\textwidth]{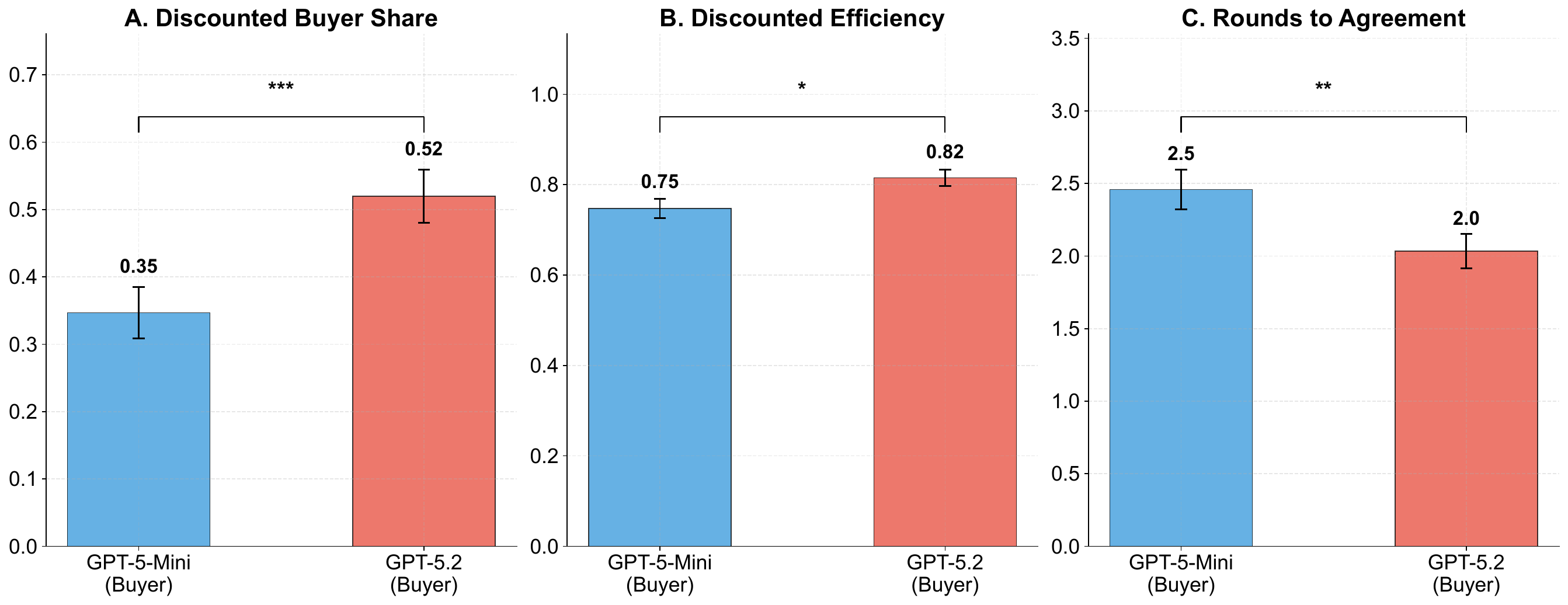}
\caption{Direction Effects in Cross-Model Negotiations}
\label{fig:ext3_direction}
\par\smallskip
\begin{minipage}{\textwidth}
\scriptsize
\textit{Notes:} Bars show direction means; error bars are 95\% confidence intervals. Brackets compare the two directional configurations. Significance stars are based on OLS regressions of each outcome on seller identity with fixed effects for the 16 design strata (four patience configurations $\times$ two buyer types $\times$ two first-proposer assignments) and cluster-robust standard errors at the stratum level. Significance levels: $^{\dagger}\,p<0.10$,
$^{*}\,p<0.05$, $^{**}\,p<0.01$, $^{***}\,p<0.001$.
\end{minipage}
\end{figure}

\subsection{Cross-Family Flagship Negotiations: Detail}
\label{subsec:B_flagship_detail}

This appendix collects the full performance summary and supporting figures for the cross-family flagship analysis summarized in Section~\ref{subsec:cross_family}. Table~\ref{tab:ext2_summary} reports deal rate, rounds, efficiency, discounted efficiency, buyer share, and the Bayesian reference for all six directional configurations. Figure~\ref{fig:ext2_advantage} decomposes the descriptive sources of Gemini's cross-family advantage (quantity gap, unit price, and surplus share across patience levels), and Figure~\ref{fig:ext2_concession} reports buyer- and seller-side concession patterns by patience configuration.

\begin{table}[htbp]
\centering
\footnotesize
\renewcommand{\arraystretch}{1.1}
\caption{Cross-Family Flagship Negotiations: Summary Statistics}
\label{tab:ext2_summary}
\begin{tabular}{llcccccc}
\toprule
Pair & Direction & Deal Rate & Rounds & Efficiency & DiscEfficiency & Buyer Share & Bayesian Ref. \\
\midrule
GPT $\leftrightarrow$ Gemini & GPT sells & 100.0\% & 3.80 & 99.6\% & 63.2\% & 54.3\% & 45.2\% \\
GPT $\leftrightarrow$ Gemini & Gemini sells & 100.0\% & 2.79 & 97.9\% & 73.2\% & 43.0\% & 45.2\% \\
GPT $\leftrightarrow$ Qwen & GPT sells & 100.0\% & 4.13 & 98.4\% & 57.1\% & 59.1\% & 45.2\% \\
GPT $\leftrightarrow$ Qwen & Qwen sells & 100.0\% & 2.24 & 91.9\% & 73.9\% & 66.4\% & 45.2\% \\
Gemini $\leftrightarrow$ Qwen & Gemini sells & 100.0\% & 4.70 & 99.0\% & 51.8\% & 61.2\% & 45.2\% \\
Gemini $\leftrightarrow$ Qwen & Qwen sells & 100.0\% & 2.35 & 94.0\% & 71.3\% & 79.0\% & 45.2\% \\
\bottomrule
\end{tabular}
\par\smallskip
\begin{minipage}{\textwidth}
\scriptsize
\textit{Notes:} We use DiscEfficiency to denote discounted efficiency. The Bayesian column reports the PBE buyer share benchmark under asymmetric demand information, evaluated at the patience configuration and proposer order of each cell. Each pair is tested in both directions across the 16 design strata (four patience configurations $\times$ two buyer types $\times$ two first-proposer assignments).
\end{minipage}
\end{table}

\paragraph{Additional cross-flagship statistics.} Three descriptive patterns supplement the figures. First, on alignment with the Bayesian benchmark, under seller-patient conditions ($\delta_B = 0.4$, $\delta_S = 0.9$), where the Bayesian benchmark allocates the buyer only 13.1\% of surplus, Gemini achieves 50.2\% ($+$37.1 pp deviation), while GPT achieves 35.0\% ($+$21.8 pp) and Qwen 53.1\% ($+$40.0 pp); these cross-family buyer advantages thus reflect not closer alignment with the benchmark reference but a consistent ability to extract surplus above it. Second, on sensitivity to the buyer's informational position, Gemini's surplus share varies least across buyer types (high: 66.1\%, low: 67.2\%, gap: $-$1.1 pp), compared to GPT (58.6\% vs.\ 50.8\%, gap: 7.8 pp) and Qwen (62.5\% vs.\ 57.8\%, gap: 4.6 pp). Third, on the efficiency decomposition for the Gemini~$\leftrightarrow$~Qwen direction, the total gap from first-best is 3.50 pp, dominated by suboptimal contract terms (3.42 pp, 97.7\%), with no deal failures (both directions reach agreement in all negotiations) and a negligible irrational component (0.08 pp).

\begin{figure}[htbp]
\centering
\includegraphics[width=\textwidth]{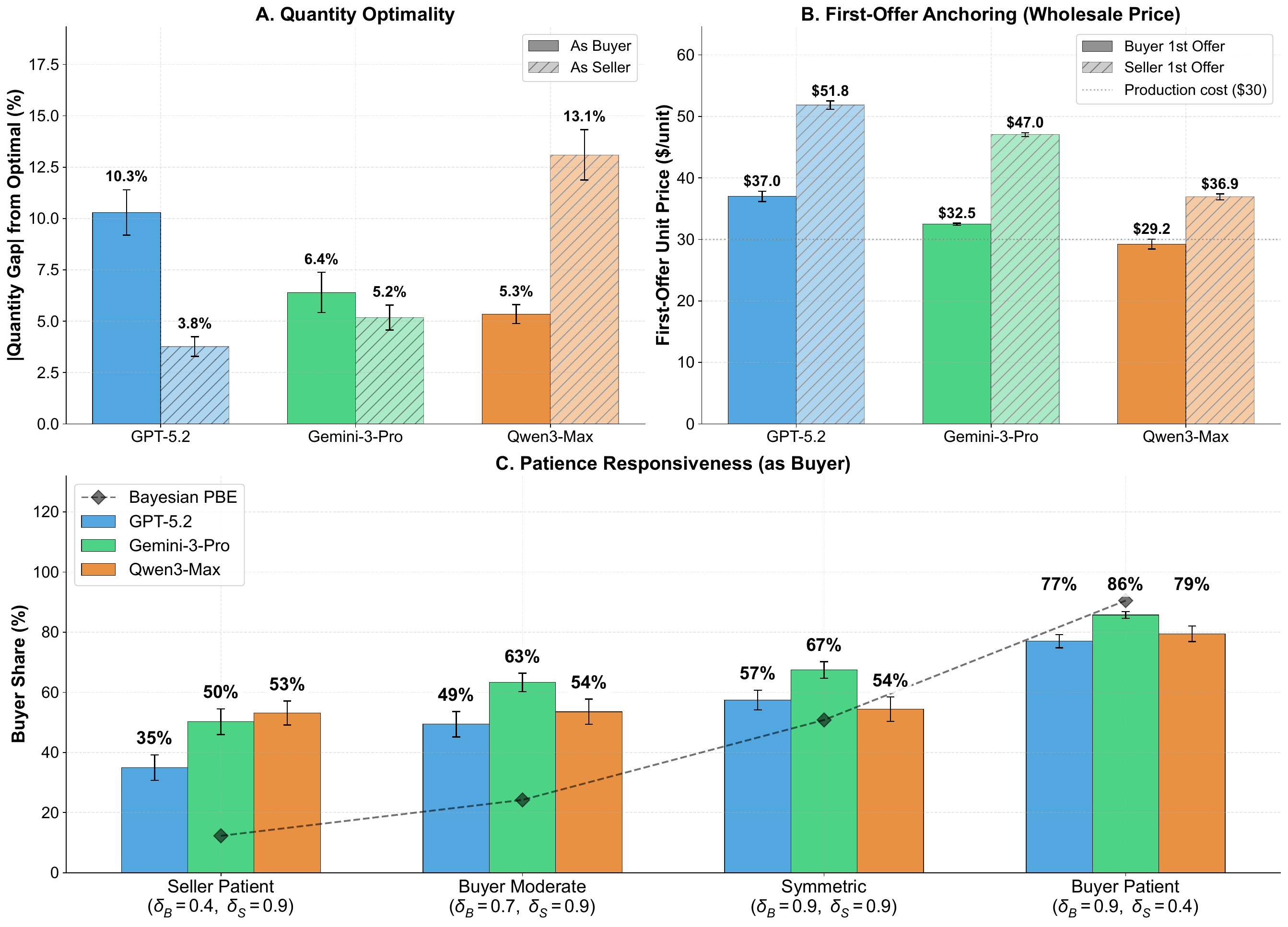}
\caption{Sources of Model Advantage in Cross-Flagship Negotiations}
\label{fig:ext2_advantage}
\par\smallskip
\begin{minipage}{\textwidth}
\scriptsize
\textit{Notes:} Sample restricted to negotiations that reached agreement. Panel~(a) quantity gap = (optimal $-$ actual) / optimal. Panel~(b) dashed horizontal line marks the production cost (\$30/unit) as reference.
Error bars are 95\% confidence intervals.
\end{minipage}
\end{figure}

\begin{figure}[htbp]
\centering
\includegraphics[width=\textwidth]{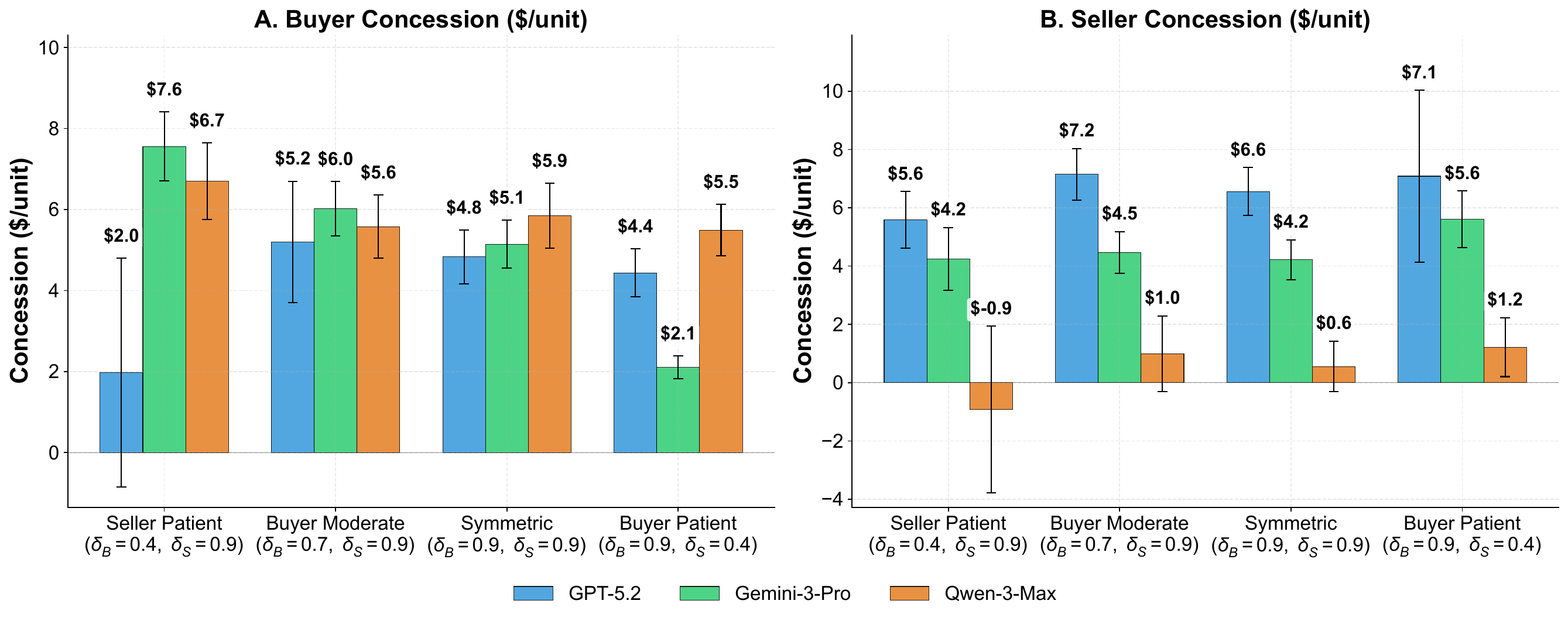}

\caption{Concession Patterns in Cross-Flagship Negotiations by Patience Configuration}
\label{fig:ext2_concession}
\par\smallskip
\begin{minipage}{\textwidth}
\scriptsize
\textit{Notes:} Sample restricted to negotiations that reached agreement. Panel~(a) Buyer concession: Qwen's buyer concessions are nearly flat across patience cells (\$5.5--\$6.7 per unit), whereas Gemini's decline from \$7.6 under seller-patient conditions to \$2.1 under buyer-patient conditions, where Qwen concedes approximately $2.6\times$ as much as Gemini. Panel~(b) Seller concession: Qwen barely concedes as seller, while GPT makes the largest seller-side adjustments.
Error bars are 95\% confidence intervals.
\end{minipage}
\end{figure}

\clearpage
\section{Strategic Behavior Details}
\label{app:strategic_behavior_details}
\label{app:strategic_detail}

This appendix provides detailed analyses supporting the process and strategic-behavior evidence summarized in Section~\ref{sec:strategic_behavior} and Section~\ref{sec:provider}. We present comprehensive anchoring analysis, LLM-inferred buyer strategy distributions, information disclosure patterns, deceptive tactics impact, and extended reasoning content analysis.

\subsection{An Illustrative Negotiation}
\label{subsec:illustrative_negotiation}

Figure~\ref{fig:neg_example_gemini} traces one representative Gemini-3-Pro self-play negotiation under buyer-first, high-patience conditions ($\delta_B = 0.9$, $\delta_S = 0.9$, high-type buyer). It illustrates the paper's central empirical pattern: the agents converge on the efficient quantity but take two rounds rather than the Bayesian benchmark of one.

\begin{figure}[htbp]
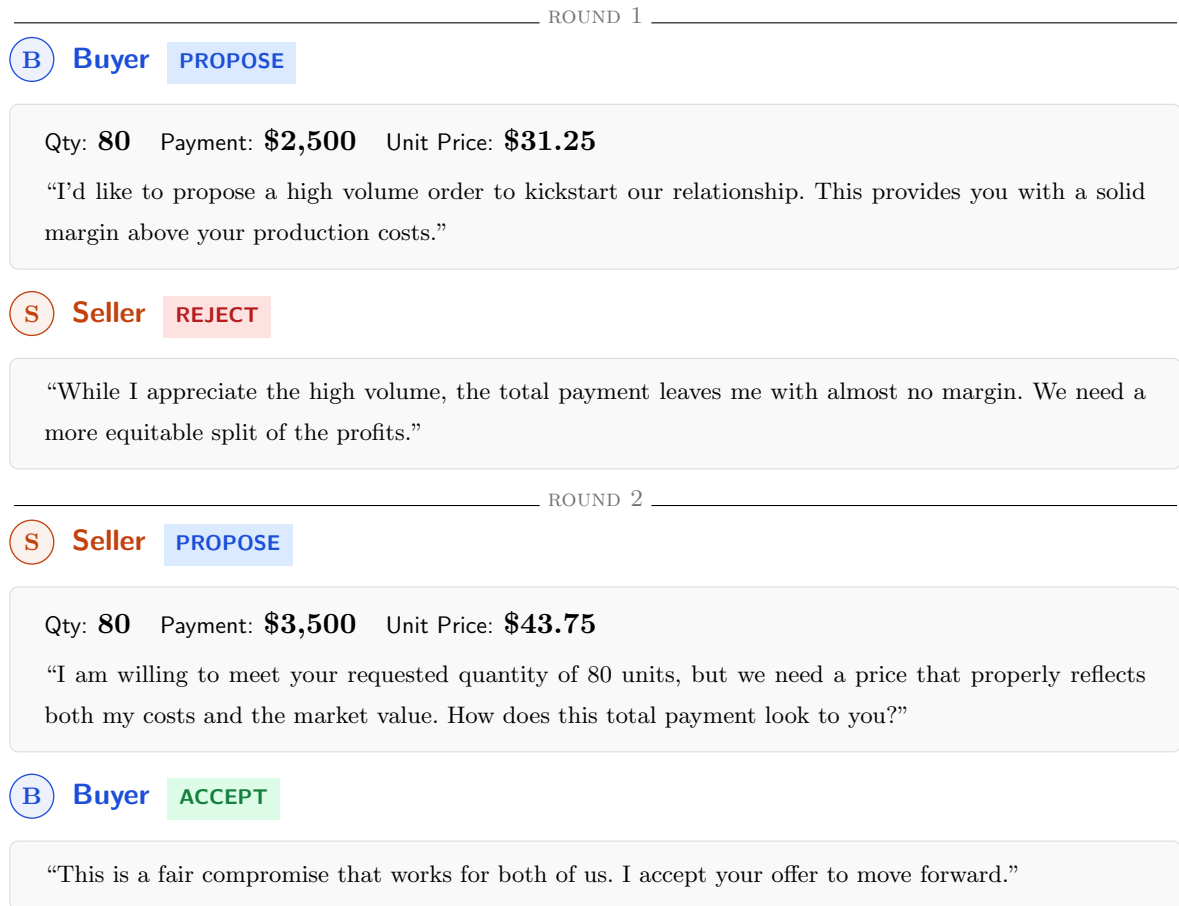

\centering
\begin{minipage}{0.94\textwidth}

{\centering\rule{18em}{0.4pt}\;{\footnotesize\sffamily\color{black!55}\textsc{round 1}}\;\rule{18em}{0.4pt}\par}
\vspace{4pt}

\noindent
\tikz[baseline=(char.base)]{\node[shape=circle, draw=buyerB, fill=buyerB!8, inner sep=1.2pt, line width=0.5pt, text=buyerB, minimum size=1.5em] (char) {\footnotesize\bfseries B};}\; \textbf{\sffamily\color{buyerB}Buyer}\; \colorbox{proposebg}{\color{proposefg}\scriptsize\sffamily\bfseries\strut\,PROPOSE\,}\par\smallskip

\begin{tcolorbox}[enhanced, breakable, colback=gray!5, colframe=gray!28, boxrule=0.4pt, arc=3pt, left=10pt, right=10pt, top=6pt, bottom=6pt, before skip=3pt, after skip=8pt]
\textsf{\footnotesize Qty:} \textbf{80}\quad \textsf{\footnotesize Payment:} \textbf{\$2{,}500}\quad \textsf{\footnotesize Unit Price:} \textbf{\$31.25}\par
\vspace{3pt}
\footnotesize``I'd like to propose a high volume order to kickstart our relationship. This provides you with a solid margin above your production costs.''
\end{tcolorbox}

\noindent
\tikz[baseline=(char.base)]{\node[shape=circle, draw=sellerS, fill=sellerS!8, inner sep=1.2pt, line width=0.5pt, text=sellerS, minimum size=1.5em] (char) {\footnotesize\bfseries S};}\; \textbf{\sffamily\color{sellerS}Seller}\; \colorbox{rejectbg}{\color{rejectfg}\scriptsize\sffamily\bfseries\strut\,REJECT\,}\par\smallskip

\begin{tcolorbox}[enhanced, breakable, colback=gray!5, colframe=gray!28, boxrule=0.4pt, arc=3pt, left=10pt, right=10pt, top=6pt, bottom=6pt, before skip=3pt, after skip=8pt]
\footnotesize``While I appreciate the high volume, the total payment leaves me with almost no margin. We need a more equitable split of the profits.''
\end{tcolorbox}

{\centering\rule{18em}{0.4pt}\;{\footnotesize\sffamily\color{black!55}\textsc{round 2}}\;\rule{18em}{0.4pt}\par}
\vspace{4pt}

\noindent
\tikz[baseline=(char.base)]{\node[shape=circle, draw=sellerS, fill=sellerS!8, inner sep=1.2pt, line width=0.5pt, text=sellerS, minimum size=1.5em] (char) {\footnotesize\bfseries S};}\; \textbf{\sffamily\color{sellerS}Seller}\; \colorbox{proposebg}{\color{proposefg}\scriptsize\sffamily\bfseries\strut\,PROPOSE\,}\par\smallskip

\begin{tcolorbox}[enhanced, breakable, colback=gray!5, colframe=gray!28, boxrule=0.4pt, arc=3pt, left=10pt, right=10pt, top=6pt, bottom=6pt, before skip=3pt, after skip=8pt]
\textsf{\footnotesize Qty:} \textbf{80}\quad \textsf{\footnotesize Payment:} \textbf{\$3{,}500}\quad \textsf{\footnotesize Unit Price:} \textbf{\$43.75}\par
\vspace{3pt}
\footnotesize``I am willing to meet your requested quantity of 80 units, but we need a price that properly reflects both my costs and the market value. How does this total payment look to you?''
\end{tcolorbox}

\noindent
\tikz[baseline=(char.base)]{\node[shape=circle, draw=buyerB, fill=buyerB!8, inner sep=1.2pt, line width=0.5pt, text=buyerB, minimum size=1.5em] (char) {\footnotesize\bfseries B};}\; \textbf{\sffamily\color{buyerB}Buyer}\; \colorbox{acceptbg}{\color{acceptfg}\scriptsize\sffamily\bfseries\strut\,ACCEPT\,}\par\smallskip

\begin{tcolorbox}[enhanced, breakable, colback=gray!5, colframe=gray!28, boxrule=0.4pt, arc=3pt, left=10pt, right=10pt, top=6pt, bottom=6pt, before skip=3pt, after skip=2pt]
\footnotesize``This is a fair compromise that works for both of us. I accept your offer to move forward.''
\end{tcolorbox}
~\\[-2mm]
\end{minipage}
\caption{Example of an LLM Negotiation Process (Gemini-3-Pro Self-Play Negotiation)}
\label{fig:neg_example_gemini}
\end{figure}

\subsection{Detailed Anchoring Analysis}
\label{subsec:C_anchoring}

We examine how opening offers shape final contract terms. Figure~\ref{fig:opening_offers} displays opening-offer patterns compared against the Bayesian benchmark for bargaining shares, summarizing the buyer-side type separation and seller-side expected-value pooling reported in Section~\ref{sec:strategic_behavior}.

\begin{figure}[ht]
\centering
\includegraphics[width=\textwidth]{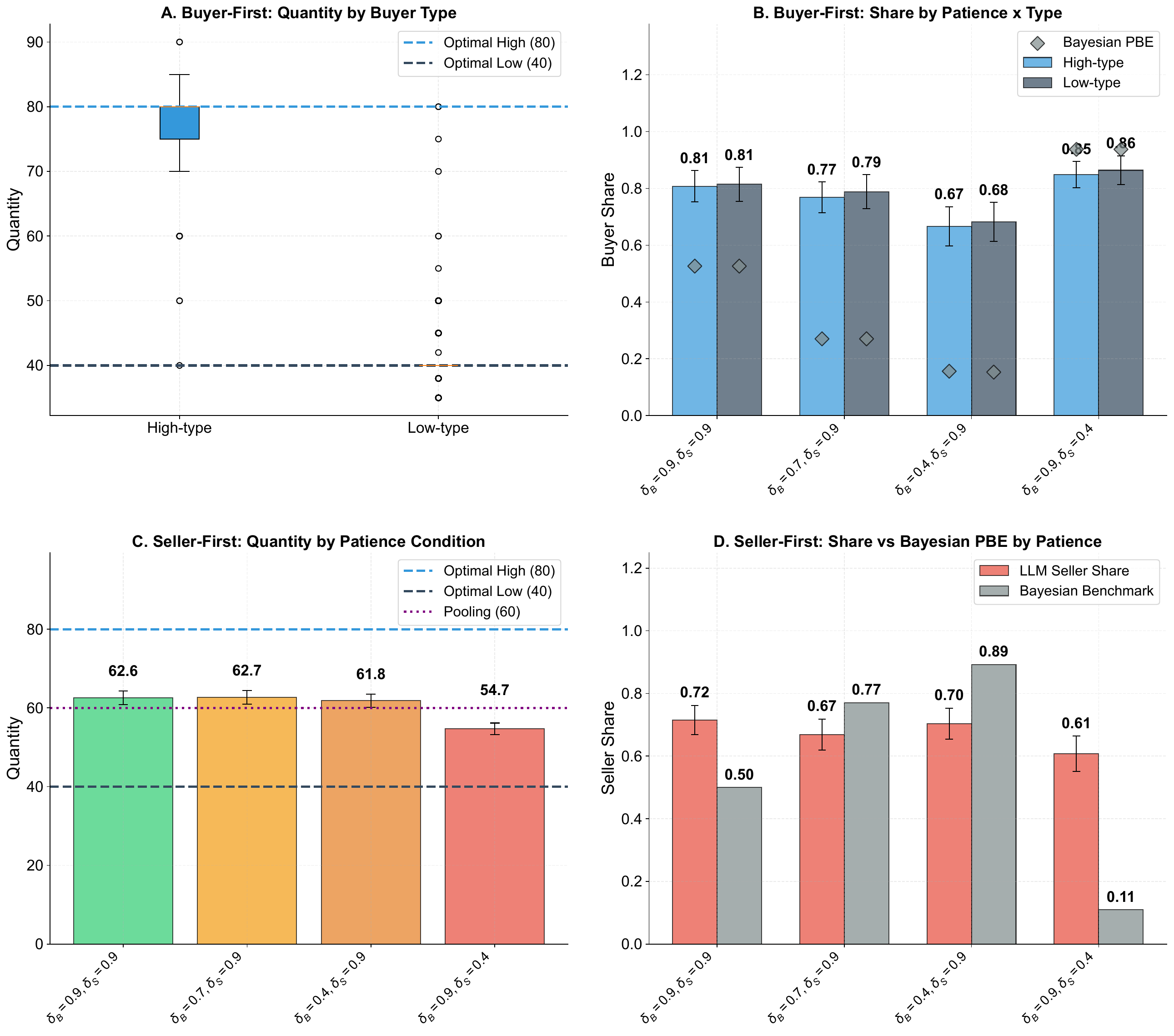}
\caption{Opening Offers and Anchoring Effects}
\label{fig:opening_offers}
\end{figure}

Table~\ref{tab:anchoring_correlations} presents correlation coefficients between opening and final terms across proposer conditions and buyer types.

\begin{table}[htbp]
\centering
\caption{Correlations Between Opening and Final Contract Terms}
\label{tab:anchoring_correlations}
\footnotesize
\renewcommand{\arraystretch}{1.1}
\begin{tabular}{@{}lccccc@{}}
\toprule
\textbf{Proposer} & \textbf{Buyer Type} & \textbf{Quantity} & \textbf{$p$-value} & \textbf{Payment} & \textbf{$p$-value} \\
& & \textbf{Correlation} & & \textbf{Correlation} & \\
\midrule
\multicolumn{6}{l}{\textit{Buyer-First Negotiations}} \\
\midrule
Overall & All & 0.911 & $<$0.001 & 0.862 & $<$0.001 \\
 & High-type & 0.349 & $<$0.001 & 0.687 & $<$0.001 \\
 & Low-type & 0.310 & $<$0.001 & 0.628 & $<$0.001 \\
\midrule
\multicolumn{6}{l}{\textit{Seller-First Negotiations}} \\
\midrule
Overall & All & 0.164 & $<$0.001 & 0.381 & $<$0.001 \\
 & High-type & 0.435 & $<$0.001 & 0.725 & $<$0.001 \\
 & Low-type & 0.269 & $<$0.001 & 0.478 & $<$0.001 \\
\bottomrule
\end{tabular}
\par\smallskip
\begin{minipage}{\textwidth}
\scriptsize
\textit{Notes:} Pearson correlation coefficients between opening offer terms and final negotiated contract terms. Buyer-first N=1,080; Seller-first N=1,080.
\end{minipage}
\end{table}

\textbf{Buyer-First Negotiations Show Type Separation, Moderate Quantity Anchoring, and Strong Payment Anchoring.} When buyers propose first, overall quantity correlation of 0.911 indicates opening quantities strongly predict final outcomes. This high correlation reflects type separation: buyers signal their type through initial quantity proposals, and these signals persist into final contracts. Decomposing by buyer type shows that both types anchor at moderate strength: high-type buyers show a quantity correlation of 0.349 and low-type buyers 0.310, indicating that opening quantities meaningfully predict final terms for both, with substantial adjustment through bargaining.

Payment anchoring follows different patterns. Both buyer types exhibit strong payment correlations, with high-types at 0.687 and low-types at 0.628, both highly significant. This suggests that while quantities may adjust through negotiation, payment terms remain anchored to opening proposals regardless of buyer type. The asymmetry between quantity and payment anchoring is best read as descriptive process evidence that different parts of the contract adjust at different rates during bargaining.

\textbf{Seller-First Negotiations Show Weaker Anchoring.} When sellers propose first, both quantity and payment correlations decline substantially. Overall quantity correlation drops to 0.164, and payment correlation to 0.381, indicating sellers' opening offers provide weaker anchors than buyers' proposals. This asymmetry may reflect sellers' pooling strategy: by proposing expected-value quantities averaging 60.5 units rather than the branch-specific screening contracts in the Bayesian benchmark, sellers create opening positions subject to greater adjustment through negotiation.

Decomposing seller-first negotiations by buyer type reveals that anchoring strength varies systematically with the realized type. Quantity correlations are moderate when the buyer is a high-type ($r=0.435$, $p<0.001$) but weaker when the buyer is a low-type ($r=0.269$, $p<0.001$). Payment correlations follow the same ordering and are noticeably stronger than the corresponding quantity correlations ($r=0.725$ for high-types and $r=0.478$ for low-types, both $p<0.001$, versus $r=0.435$ and $r=0.269$ for quantity). The within-type correlations are sharper than the pooled estimates because pooling mixes negotiations whose realized contracts diverge in opposite directions from a common opening offer. Even for high-types, where anchoring is strongest, correlations remain well below the buyer-first benchmark, consistent with seller opening offers establishing reference points that influence but do not determine final terms.

\subsection{Inferred Buyer Strategy Distributions}
\label{subsec:C_strategy}

Because the experimental protocol records public messages and private reasoning traces but does not elicit self-reported strategy labels, we classify buyer strategy from the recorded turn-level text using an independent language-model coder (Claude Haiku 4.5). For each buyer turn, the coder observes the public message and private reasoning trace and assigns one of four categories: \textit{signal} (truthfully conveying the buyer's type to separate), \textit{mimic} (misrepresenting type by posing as the other type), \textit{optimal} (straightforward payoff maximization with no type-strategic intent), or \textit{other}.
Table~\ref{tab:strategy_inferred_detailed} reports the distribution by buyer type and patience, Table~\ref{tab:strategy_inferred_capability} by capability tier, and Figure~\ref{fig:strategy_inferred_appendix} visualizes the type-and-patience breakdown.

\begin{figure}[ht]
\centering
\includegraphics[width=\textwidth]{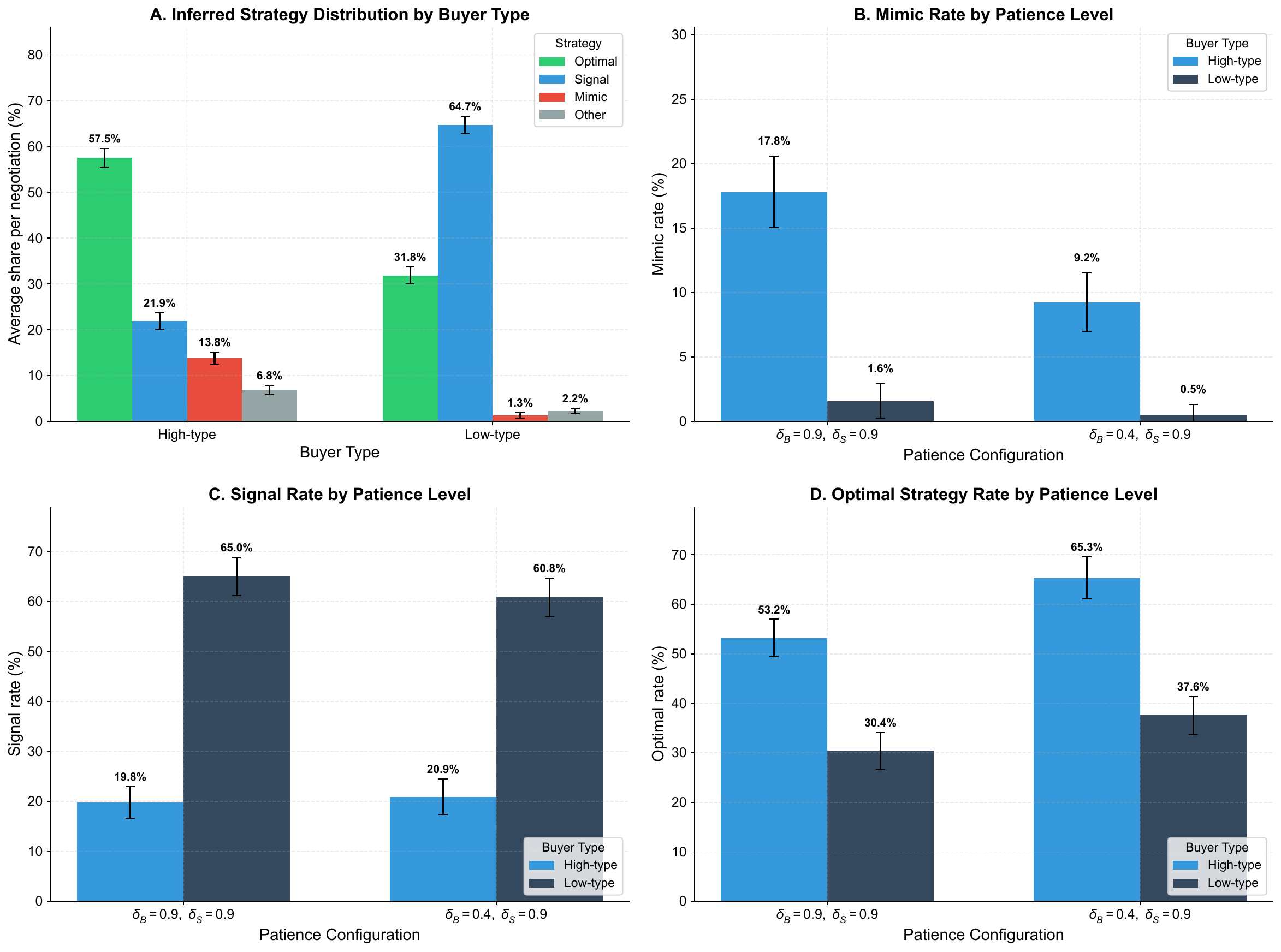}
\caption{LLM-Inferred Buyer Strategy by Type and Patience}
\par\smallskip
\begin{minipage}{\textwidth}
\scriptsize
\textit{Notes:} Bars report average within-negotiation shares of classified buyer turns; error bars are 95\% confidence intervals.
\end{minipage}
\label{fig:strategy_inferred_appendix}
\end{figure}

\begin{table}[htbp]
\centering
\caption{LLM-Inferred Buyer Strategy: by Type and Patience}
\label{tab:strategy_inferred_detailed}
\footnotesize
\renewcommand{\arraystretch}{1.1}
\begin{tabular}{@{}lccccc@{}}
\toprule
\textbf{Condition} & \textbf{$N$} & \textbf{Optimal} & \textbf{Signal} & \textbf{Mimic} & \textbf{Other} \\
& & \textbf{(\%)} & \textbf{(\%)} & \textbf{(\%)} & \textbf{(\%)} \\
\midrule
\multicolumn{6}{l}{\textit{By Buyer Type}} \\
\midrule
High-Type & 1,071 & 57.5 & 21.9 & 13.8 & 6.8 \\
Low-Type & 1,079 & 31.8 & 64.7 & 1.3 & 2.2 \\
Holm-adjusted $p$-value & & $<$0.001 & $<$0.001 & $<$0.001 & $<$0.001 \\
\midrule
\multicolumn{6}{l}{\textit{By Patience Configuration (High-Type Buyers)}} \\
\midrule
(0.9, 0.9) & 269 & 53.2 & 19.8 & 17.8 & 9.2 \\
(0.7, 0.9) & 269 & 56.7 & 23.2 & 14.0 & 6.2 \\
(0.4, 0.9) & 266 & 65.3 & 20.9 & 9.2 & 4.5 \\
(0.9, 0.4) & 267 & 54.8 & 23.7 & 14.0 & 7.4 \\
Holm-adjusted $p$-value & & $<$0.001 & 0.340 & $<$0.001 & 0.013 \\
\midrule
\multicolumn{6}{l}{\textit{By Patience Configuration (Low-Type Buyers)}} \\
\midrule
(0.9, 0.9) & 270 & 30.4 & 65.0 & 1.6 & 3.1 \\
(0.7, 0.9) & 270 & 30.9 & 66.1 & 1.0 & 1.9 \\
(0.4, 0.9) & 270 & 37.6 & 60.8 & 0.5 & 1.1 \\
(0.9, 0.4) & 269 & 28.5 & 66.7 & 2.0 & 2.9 \\
Holm-adjusted $p$-value & & 0.027 & 0.289 & 0.289 & 0.029 \\
\bottomrule
\end{tabular}
\par\medskip\begin{minipage}{\textwidth}\scriptsize\textit{Notes:} The sample comprises 6{,}279 classified buyer turns across 2{,}150 negotiations. Each percentage is the average within-negotiation share of classified buyer turns. The buyer-type panel uses independent Welch tests; the patience panels use Welch ANOVA. Holm correction is applied across the four strategy outcomes within each panel.\end{minipage}
\end{table}

\textbf{Type Asymmetry Aligns with Theory.} The two type-strategic categories split sharply, in the directions \citet{feng2015dynamic} predict. The average mimic rate per negotiation is $13.8\%$ for high-type buyers versus $1.3\%$ for low-type buyers, consistent with the high type's incentive to pose as a low-demand buyer and avoid surplus extraction. Conversely, the average signaling rate per negotiation is $64.7\%$ for low-type buyers versus $21.9\%$ for high-type buyers, consistent with the low type's incentive to separate credibly. Both differences remain significant after Holm correction ($p<0.001$), and non-type-strategic ``optimal'' play is the modal high-type category ($57.5\%$).

\textbf{Patience Effects Contradict Theory.} The comparative static, however, does not align. \citet{feng2015dynamic} predict that high-type mimicking should intensify when buyer patience is low, since a revealed high-type is then more exposed to seller extraction. We find the opposite: the average mimic rate per negotiation is \textit{lowest} under the most buyer-impatient configuration ($\delta_B = 0.4$, $\delta_S = 0.9$) at $9.2\%$ and \textit{highest} under symmetric high patience ($\delta_B = \delta_S = 0.9$) at $17.8\%$ (Holm-adjusted Welch ANOVA, $p<0.001$). Agents thus exhibit behavior consistent with high-type mimicking but do not calibrate it to the patience-based incentive the theory emphasizes; this is the same qualitative anomaly that the buyer-side quantity distortions in Section~\ref{sec:strategic_behavior} exhibit.

\textbf{Model Capability Effects.} The average mimic rate per negotiation rises with model capability (Table~\ref{tab:strategy_inferred_capability}): $11.1\%$ for flagship buyers, $6.9\%$ for mid-tier buyers, and $4.5\%$ for baseline buyers (Holm-adjusted Welch ANOVA, $p<0.001$). More capable buyers are more often classified as mimicking the other type, consistent with the reasoning capacity a pooling strategy requires; signaling, by contrast, is common across all tiers ($37.7$--$50.6\%$). This gradient is consistent with greater strategic sophistication, though the inferred mimic rate should be read as approximate.

\begin{table}[htbp]
\centering
\caption{LLM-Inferred Buyer Strategy by Model Capability Tier}
\label{tab:strategy_inferred_capability}
\footnotesize
\renewcommand{\arraystretch}{1.1}
\begin{tabular}{@{}lccccc@{}}
\toprule
\textbf{Model Tier} & \textbf{$N$} & \textbf{Optimal (\%)} & \textbf{Signal (\%)} & \textbf{Mimic (\%)} & \textbf{Other (\%)} \\
\midrule
Flagship & 718 & 40.3 & 41.7 & 11.1 & 6.9 \\
Mid-tier & 715 & 51.1 & 37.7 & 6.9 & 4.2 \\
Baseline & 717 & 42.5 & 50.6 & 4.5 & 2.4 \\
\midrule
Holm-adjusted $p$-value & & $<$0.001 & $<$0.001 & $<$0.001 & $<$0.001 \\
\bottomrule
\end{tabular}
\par\medskip\begin{minipage}{\textwidth}\scriptsize\textit{Notes:} The sample comprises 6{,}279 classified buyer turns across 2{,}150 negotiations. Each percentage is the average within-negotiation share of classified buyer turns. Welch ANOVA compares capability tiers, with Holm correction across the four strategy outcomes.\end{minipage}
\end{table}

\subsection{Information Disclosure and Public--Private Divergence}
\label{subsec:C_disclosure}
We first describe buyers' disclosure rates by type, then turn to the four public--private divergence tactics and their effect on outcomes. A language-model classifier\footnote{Classification was performed using Claude Haiku 4.5, which is independent of all three studied agent vendors (OpenAI, Google, Alibaba).
} coded each buyer turn's public message into disclosure categories: direct truthful (explicitly revealing type), indirect truthful (implying type through context), deceptive (posing as the other type), or withholding (avoiding disclosure). Table~\ref{tab:disclosure_detailed} reports the resulting rates.

\begin{table}[htbp]
\centering
\caption{Information Disclosure Patterns by Buyer Type}
\label{tab:disclosure_detailed}
\footnotesize
\renewcommand{\arraystretch}{1.1}
\begin{tabular}{@{}lcccccc@{}}
\toprule
\textbf{Buyer} & \textbf{N} & \textbf{Any} & \textbf{Direct} & \textbf{Indirect} & \textbf{Deceptive} & \textbf{Withhold} \\
\textbf{Type} & & \textbf{Truthful} & \textbf{Truthful} & \textbf{Truthful} & & \\
& & \textbf{(\%)} & \textbf{(\%)} & \textbf{(\%)} & \textbf{(\%)} & \textbf{(\%)} \\
\midrule
High-Type & 1,078 & 48.2 & 27.2 & 21.0 & 4.0 & 47.8 \\
Low-Type & 1,080 & 52.1 & 21.3 & 30.8 & 1.4 & 46.5 \\
\midrule
Difference & & -3.9* & +5.9*** & -9.8*** & +2.6*** & +1.2 \\
\bottomrule
\end{tabular}
\par\smallskip
\begin{minipage}{\textwidth}
\scriptsize
\textit{Notes:} Each percentage is the average within-negotiation share of classified buyer turns. $N$ denotes negotiations. The difference row reports high-type minus low-type percentage points. Stars are based on independent Welch tests with Holm correction across the five outcomes. *** $p<0.001$, ** $p<0.01$, * $p<0.05$.
\end{minipage}
\end{table}

\textbf{Disclosure Rate Asymmetries.} The average truthful-disclosure rate per negotiation is higher for low-type than high-type buyers (52.1 versus 48.2\%, Holm-adjusted $p<0.05$). The two types differ in mode, however: high-types use \textit{direct} disclosure more (27.2 versus 21.3\%, Holm-adjusted $p<0.001$), whereas low-types rely more on \textit{indirect} disclosure (30.8 versus 21.0\%, Holm-adjusted $p<0.001$), so the net truthfulness gap is carried by indirect statements. High-types withhold marginally more (47.8 versus 46.5\%, not significant). At the net level the pattern aligns with signaling theory: low-types disclose more to separate from high-types, while high-types more often withhold or send deceptive type signals.

\textbf{Deception Concentrated Among High-Types.} The average deceptive-disclosure rate per negotiation is low but higher among high-type buyers (4.0 versus 1.4\%, Holm-adjusted $p<0.001$). This aligns with the theoretical prediction that high-types have a structural incentive to mimic low-types to avoid surplus extraction.

\textbf{From disclosure to public--private divergence.} The disclosure rates above describe what the public message asserts about the buyer's type; they do not speak to whether the public framing aligns with private reasoning. We next compare buyers' public messages to their private reasoning traces and identify four recurring forms of public--private divergence. \textbf{Fairness Claim Mismatch} occurs when buyers publicly appeal to fairness while privately reasoning in purely self-interested terms. \textbf{Profit Hiding} occurs when buyers conceal or understate expected profitability, portraying acceptable terms as marginally viable when private reasoning reveals substantial gains. \textbf{False Constraint} and \textbf{False Finality} occur when buyers claim non-existent constraints or declare offers final while privately intending continued negotiation. These labels come from a language-model classifier rather than validated human annotation and should be read as exploratory indicators of public--private divergence rather than validated prevalence estimates of intentional deception. Table~\ref{tab:deception_tactics} reports prevalence.

\begin{table}[htbp]
\centering
\caption{Public--Private Message Divergence Tactics}
\label{tab:deception_tactics}
\footnotesize
\renewcommand{\arraystretch}{1.1}
\begin{tabular}{lcc}
\toprule
\textbf{Tactic} & \textbf{Prevalence (\%)} & \textbf{Observed Count} \\
\midrule
Fairness Claim Mismatch & 56.5 & 7,352 \\
Profit Hiding & 14.1 & 1,831 \\
False Constraint & 9.4 & 1,222 \\
False Finality & 0.7 & 91 \\
\midrule
\textbf{Total Turns Analyzed} & & \textbf{13,023} \\
\bottomrule
\end{tabular}
\par\smallskip
\begin{minipage}{\textwidth}
\scriptsize
\textit{Notes:} This table is descriptive at the turn level: percentages are calculated over all 13{,}023 cache-hit turns, and some turns exhibit multiple tactics. Negotiation-level role comparisons appear in Table~\ref{tab:deception_tactics_family_role}.
\end{minipage}
\end{table}

The pooled turn-level prevalence in Table~\ref{tab:deception_tactics} averages across two distinctions that turn out to matter: which side of the negotiation produced the turn, and which provider family generated the agent. Table~\ref{tab:deception_tactics_family_role} instead compares average buyer and seller tactic rates within the same negotiation. Using buyer-minus-seller differences, the split reveals that the asymmetry is not uniform across families. Within OpenAI, buyer and seller fairness-mismatch rates are nearly identical ($+0.1$ pp), and the $+2.8$ pp profit-hiding difference is not significant after Holm correction. Google buyers have lower rates than Google sellers for both fairness mismatch ($-9.3$ pp) and profit hiding ($-3.0$ pp), whereas Qwen buyers have higher rates than Qwen sellers ($+8.7$ and $+3.3$ pp, respectively). These patterns suggest that public--private divergence in framing is a population-average behavior, but one whose role incidence varies in sign across provider families.

\begin{table}[htbp]
\centering
\caption{Public--Private Divergence Tactics by Provider Family and Role}
\label{tab:deception_tactics_family_role}
\footnotesize
\renewcommand{\arraystretch}{1.1}
\begin{tabular}{@{}llccccc@{}}
\toprule
\textbf{Family} & \textbf{Role} & \textbf{N} & \textbf{Fairness Mismatch} & \textbf{Profit Hiding} & \textbf{False Constraint} & \textbf{False Finality} \\
& & & \textbf{(\%)} & \textbf{(\%)} & \textbf{(\%)} & \textbf{(\%)} \\
\midrule
OpenAI & Buyer & 720 & 39.1 & 17.0 & 5.8 & 1.2 \\
 & Seller & 720 & 39.1 & 14.2 & 9.4 & 1.2 \\
 & $\Delta$ (B$-$S, pp) & & +0.1 & +2.8 & -3.6*** & 0.0 \\
\midrule
Google & Buyer & 718 & 63.2 & 10.1 & 8.2 & 0.2 \\
 & Seller & 718 & 72.5 & 13.1 & 6.7 & 0.1 \\
 & $\Delta$ (B$-$S, pp) & & -9.3*** & -3.0** & +1.5 & +0.1 \\
\midrule
Qwen & Buyer & 718 & 59.4 & 15.8 & 11.9 & 0.2 \\
 & Seller & 718 & 50.8 & 12.5 & 5.8 & 0.3 \\
 & $\Delta$ (B$-$S, pp) & & +8.7*** & +3.3** & +6.1*** & -0.1 \\
\bottomrule
\end{tabular}
\par\smallskip
\begin{minipage}{\textwidth}
\scriptsize
\textit{Notes:} Each percentage is the average within-negotiation share of cache-hit turns exhibiting the tactic. $N$ denotes negotiations with classified buyer and seller turns. The $\Delta$ row reports the within-negotiation buyer-minus-seller difference in percentage points. Stars are based on paired t-tests with Holm correction across the four tactics within each family. *** $p<0.001$, ** $p<0.01$, * $p<0.05$.
\end{minipage}
\end{table}

\subsection{Process Traces: Why Agents Take Extra Rounds}
\label{subsec:C_process_efficiency}

This appendix presents the full process-trace evidence behind the round-count gap summarized in Section~\ref{sec:strategic_behavior}. The round-count gap documented in the main text has a clear process-level signature: LLM agents recognize the structure of the bargaining problem but substitute robust heuristics for the branch-specific execution the PBE prescribes. We document the substitution in two places, opening offers and inferred strategies, and show that both contribute to the deviation from equilibrium timing.

On opening offers, the substitution depends on which side moves first. When buyers open, they reproduce the type separation predicted by the \citet{feng2015dynamic} truth-telling equilibrium: high-type quantities average $77.8$ units and low-type quantities average $41.6$ units, with opening-to-final correlations of $0.911$ for quantity and $0.862$ for payment. When sellers open, by contrast, they propose $60.5$ units on average, essentially the uniform-prior expected value of $60$, rather than the branch-specific screening menu the PBE prescribes; opening-to-final correlations fall to $0.164$ for quantity and $0.381$ for payment. This pooling substitution contributes to the timing gap on the seller-first side: a seller who opens at a pooling quantity cannot reach the separating contract in round one and defers separation to later rounds. But seller-first negotiations are not measurably slower than buyer-first ones (rounds do not differ significantly by first-proposer order, $F=0.12$, n.s.; Table~\ref{tab:main_results}), so pooling substitution alone does not explain the pooled $2.98$-round average against the $1.25$-round equilibrium prediction; on the buyer-first side, where opening offers already separate by type, the residual delay instead reflects post-agreement price haggling over an already-anchored quantity (Appendix~\ref{subsec:illustrative_negotiation}). Wholesale prices show minimal type-based differentiation in either case, with medians at $40$ for both buyer types, indicating that LLM agents recover the quantity-separating signal but not the type-dependent price separation.

A complementary signature appears in how agents handle the high type's mimicking incentive (Appendix~\ref{subsec:C_strategy}). The average mimic rate per negotiation is $13.8\%$ for high-type buyers versus $1.3\%$ for low-type buyers (Holm-adjusted $p<0.001$). High-type realized quantities are also $7.1\%$ below first-best, consistent with this interpretation, although the quantity distortion alone does not establish mimicking. The comparative static, however, runs the wrong way: theory predicts more high-type mimicking under low buyer patience, but the average mimic rate instead declines from $17.8\%$ under symmetric high patience to $9.2\%$ under low buyer patience (Holm-adjusted Welch ANOVA, $p<0.001$). Agents enact the static mimicking incentive yet do not calibrate it to the dynamic patience incentive on which the equilibrium turns, a behavior--theory gap that accompanies the implementation failures behind the timing-and-efficiency gap.

\subsection{Process Traces: How Buyers Exploit the Verbal Channel}
\label{subsec:C_process_verbal}
\label{subsec:framing_channel}
The provider-level patterns documented in Section~\ref{sec:provider} are accompanied by a role-asymmetric pattern in how the verbal channel gets used, though the asymmetry is itself provider-conditional. An LLM classifier flags fairness-claim mismatch (public appeals to fairness paired with self-interested private reasoning) in roughly $56\%$ of classified turns overall and profit hiding in about $14\%$. As detailed in Appendix~\ref{subsec:C_disclosure} (Table~\ref{tab:deception_tactics_family_role}), this asymmetry is provider-specific: Google buyers show markedly lower rates than Google sellers, OpenAI shows no meaningful gap, and Qwen's gap runs the opposite way.

\subsection{Strategic Patience: Per-Model Heterogeneity (R3)}
\label{subsec:C_patience_heterogeneity}

This subsection reports the full strategic-patience analysis summarized in Section~\ref{subsec:strategic-patience} of the main paper: the headline gain table (Table~\ref{tab:ext9_summary}), the buyer-payoff curves by prompted patience (Figure~\ref{fig:strategic_patience_buyer}), and the per-model decomposition (Figure~\ref{fig:strategic_patience_heterogeneity}) showing which models translate structural patience into realized payoff.

\begin{table}[htbp]
\centering
\footnotesize
\caption{Gains from Optimally Prompted Strategic Patience}
\label{tab:ext9_summary}
\begin{tabular}{@{}lcccrl@{}}
\toprule
Party & True $\delta^{\mathrm{econ}}$ & \makecell{Matched\\avg-optimal models} & \makecell{Matched\\per-condition optimal} & \makecell{Mean gain\\$[$95\% CI$]$} & Conclusion \\
\midrule
Buyer  & 0.9 & 4/9 & 20/36 (56\%) & \makecell[r]{$+74.81$\\$[+29.43,\,+130.17]$} & \makecell[l]{Matched-high modal;\\understate for most} \\
Seller  & 0.9 & 9/9 & 26/36 (72\%) & \makecell[r]{$+18.28$\\$[+4.35,\,+35.06]$} & Matched-high dominant \\
Buyer  & 0.7 & 4/9 & 13/36 (36\%) & \makecell[r]{$+41.11$\\$[+25.11,\,+58.41]$} & \makecell[l]{Matched most common;\\3 overstate, 2 understate} \\
Seller  & 0.7 & 2/9 & 11/36 (31\%) & \makecell[r]{$+66.53$\\$[+39.05,\,+100.50]$} & \makecell[l]{High patience dominant;\\two matched exceptions} \\
\bottomrule
\end{tabular}

\par\smallskip
\begin{minipage}{\textwidth}
\scriptsize
\textit{Notes:} For each cell (model $\times$ buyer-type $\times$ first-proposer), we identify the empirically best prompted $\delta^{\mathrm{strat}}$ (including the matched value $\delta^{\mathrm{strat}} = \delta^{\mathrm{econ}}$ as one of the alternatives) and compute gain $=$ best payoff $-$ matched payoff $\geq 0$. \textit{Per-condition optimal} reports the count of cells in which the matched value is itself the best; \textit{avg-optimal models} reports the count of models for which the matched value yields the highest mean payoff after averaging across the four conditions. \textit{Mean gain} averages the cell-level gains across the 36 cells; positive values indicate that some strategic alternative beats matched on average. Brackets are 95\% percentile bootstrap CIs over the 36 cells (2{,}000 resamples).

\end{minipage}

\end{table}

\begin{figure}[htbp]
\centering
\includegraphics[width=\textwidth]{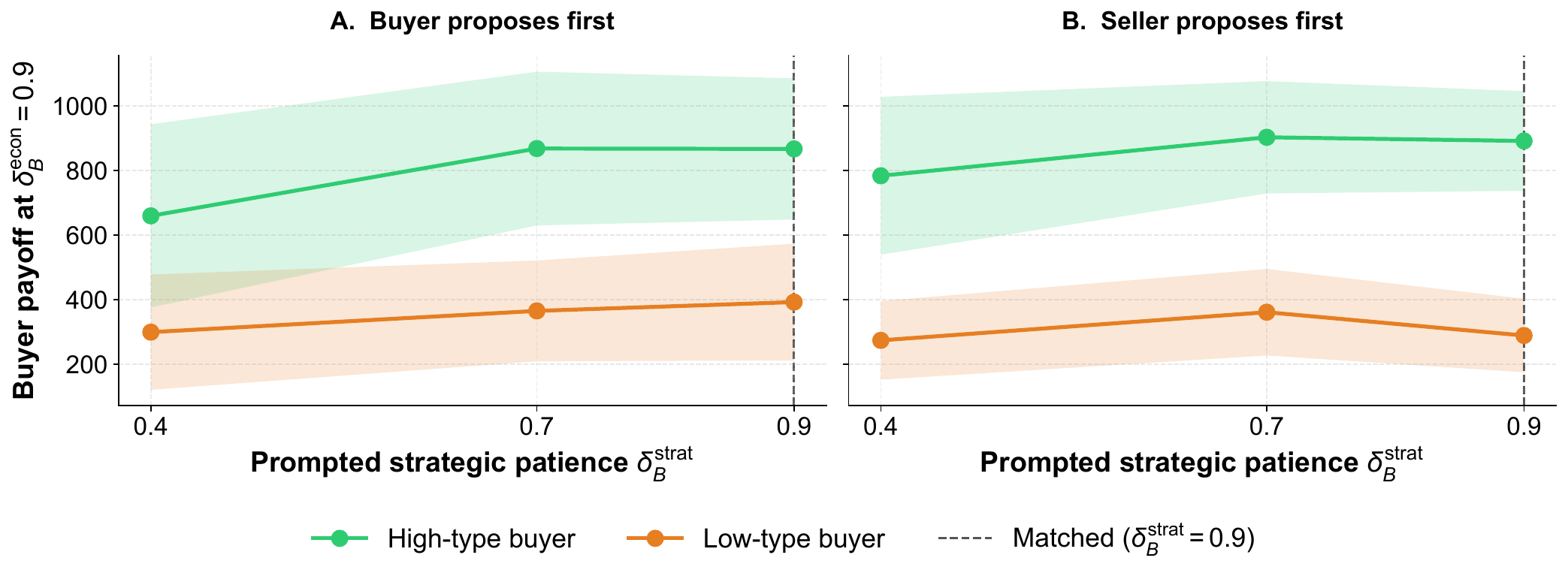}
\caption{Buyer Payoff by Prompted Strategic Patience}
\label{fig:strategic_patience_buyer}
\vspace{0.5em}
\begin{minipage}{0.95\textwidth}
\scriptsize
\textit{Notes.} Buyer payoff at $\delta^{\text{econ}}_B = 0.9$ is plotted as a function of prompted strategic patience $\delta^{\text{strat}}_B$, separately for H-type and L-type buyers. Panel~A restricts to negotiations in which the buyer proposes first; Panel~B to those in which the seller proposes first. Each line averages across the nine main verbal models, with shaded bands giving 95\% confidence intervals across models. The right edge of each panel and the dashed vertical line both mark the matched choice $\delta^{\text{strat}}_B = \delta^{\text{econ}}_B = 0.9$.
\end{minipage}
\end{figure}

\begin{figure}[htbp]
\centering
\includegraphics[width=\textwidth]{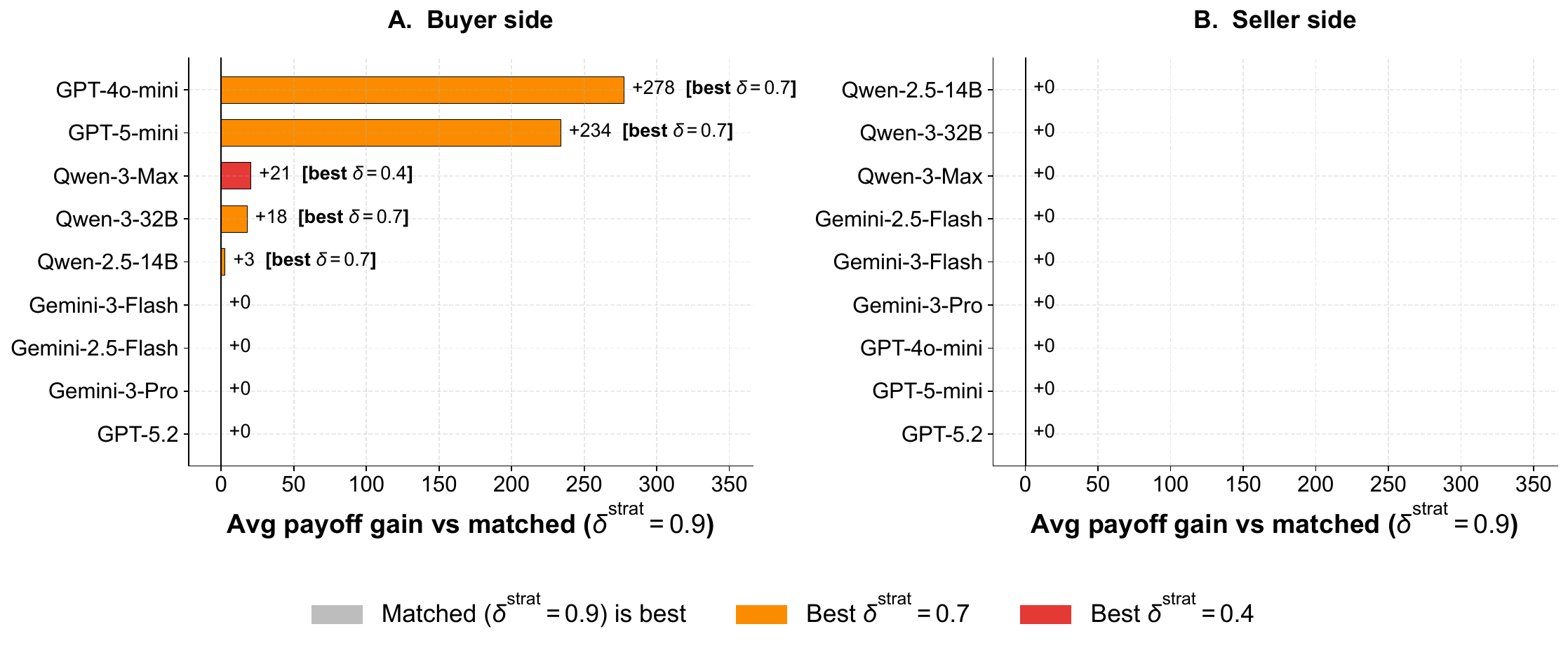}
\caption{Per-model Gain from Optimal Strategic Patience}
\label{fig:strategic_patience_heterogeneity}
\vspace{0.5em}
\begin{minipage}{0.95\textwidth}
\scriptsize
\textit{Notes.} Bars show the gain from choosing the empirically best prompted $\delta^{\text{strat}}$ relative to the matched baseline $\delta^{\text{strat}} = \delta^{\text{econ}} = 0.9$, averaged across the four buyer type $\times$ first proposer conditions. Panel A: buyers. Panel B: sellers. Near-zero gray bars indicate that the matched baseline ($\delta^{\text{strat}} = \delta^{\text{econ}}$) is average-optimal. The pattern is asymmetric across roles. On the seller side (Panel B) matching is average-optimal for all nine models, so every bar is essentially flat. On the buyer side (Panel A) five of nine models realize higher payoff by understating patience to $\delta^{\text{strat}}_B = 0.7$ (or $0.4$ for Qwen3-Max), with the largest gains for GPT-4o-mini ($+278$) and GPT-5-mini ($+234$); the four remaining models (GPT-5.2 and the three Gemini models) are average-optimal at the matched value. For buyer models that benefit from understating patience, the gain arises through higher agreement rates, more favorable terms, or shorter bargaining, depending on the model.
\end{minipage}
\end{figure}

\clearpage
\section{Design and Robustness Extension Program}
\label{sec:extension_program}

This appendix reports the complete design and robustness extension program supporting the paper's main findings. Section~\ref{subsec:inventory} lists every LLM-to-LLM negotiation conducted across both the main analysis and the design and robustness extensions. The seven completed extensions are reported in full in the paper: a structured-communication treatment with verbal messages removed (R1; Section~\ref{subsec:r1_structured}), a no-discounting bargaining treatment that removes the discounting framework from agent prompts (R2; Section~\ref{app:ext_unprompted}), a strategic-patience analysis (R3; Section~\ref{subsec:C_patience_heterogeneity}), a reasoning-effort ablation on GPT-5.2 (R4; Section~\ref{subsec:reasoning_effort}), a retail-price robustness check on the OpenAI family (R5; Section~\ref{subsec:retail_price_robustness}), a sensitivity analysis over the seller's prior belief about buyer type (R6; Section~\ref{subsec:prior_sensitivity}), and a parameter-size ablation within the Qwen3 family (R7; Section~\ref{subsec:parameter_size}). R1--R3 are the configuration analyses reported in Section~\ref{sec:design_choice}; R4--R7 are the additional robustness extensions. The extensions probe whether the headline findings of Sections~\ref{sec:performance}--\ref{sec:provider} generalize beyond the experimental choices made in the main analysis, along dimensions of communication mode, time-preference framing, prompted strategic patience, inference-time compute, surplus magnitude, prior beliefs, and model size.

\subsection{Complete Experimental Inventory}
\label{subsec:inventory}

Table~\ref{tab:inventory} accounts for all LLM-to-LLM negotiations conducted in this paper. The inventory is organized into two tiers: the main analysis (M1--M3), whose findings are reported in the main text, and the design and robustness extensions (R1--R7), whose findings are reported in Section~\ref{sec:design_choice} and the appendix sections referenced above. M1--M3 together contribute 4{,}080 negotiations and 12{,}490 bargaining rounds; the completed design and robustness extensions (R1--R7) contribute an additional 5{,}760 negotiations and 19{,}102 rounds. The full program therefore totals 9{,}840 unique negotiations and 31{,}592 rounds across nine models from three providers.

\begin{table}[htbp]
\centering
\caption{Experiment Inventory: Main Analysis and Design and Robustness Extensions}
\label{tab:inventory}
\footnotesize
\renewcommand{\arraystretch}{1.1}
\setlength{\tabcolsep}{3pt}
\resizebox{\textwidth}{!}{%
\begin{tabular}{lccccclrrr}
\toprule
 & \multicolumn{4}{c}{Condition Factors} & & & & & \\
\cmidrule(lr){2-5}
Block & Buyer Types & First Proposers & Patience Levels & Other & Cond. & Models & Exps & Rounds & Reps \\
\midrule
\rowcolor{gray!20}
\multicolumn{10}{l}{\textit{Main Analysis}} \\
\midrule
\makecell[l]{M1: Verbal main\\factorial} & 2 (H, L) & 2 (B, S) & 4 & --- & 16 & 9 (3 providers $\times$ 3 tiers) & 2{,}160 & 6{,}607 & 15.0 \\
\midrule
\makecell[l]{M2: Cross-family\\flagship} & 2 (H, L) & 2 (B, S) & 4 & \makecell{2 direc-\\tions} & 16 & \makecell{3 (GPT-5.2,\\Gemini-3-Pro,\\Qwen3-Max)} & 1{,}440 & 4{,}805 & 15.0 \\
\midrule
\makecell[l]{M3: Cross-capability} & 2 (H, L) & 2 (B, S) & 4 & 2 directions & 32 & \makecell{1 pair (GPT-5.2\\$\leftrightarrow$ GPT-5-mini)} & 480 & 1{,}078 & 15.0 \\
\midrule
\rowcolor{yellow!10}
\textbf{Main analysis subtotal} & & & & & & & \textbf{4{,}080} & \textbf{12{,}490} & \\
\midrule
\rowcolor{gray!20}
\multicolumn{10}{l}{\textit{Design and Robustness Extensions}} \\
\midrule
\makecell[l]{R1: Structured\\communication} & 2 (H, L) & 2 (B, S) & 4 & --- & 16 & 9 (3 providers $\times$ 3 tiers) & 2{,}160 & 7{,}331 & 15.0 \\
\midrule
\makecell[l]{R2: No-discounting\\bargaining} & 2 (H, L) & 2 (B, S) & --- & no $\delta$ & 4 & 9 (3 providers $\times$ 3 tiers) & 540 & 2{,}229 & 15.0 \\
\midrule
\makecell[l]{R3: Strategic patience\\$(\delta_B,\delta_S)=(0.9,0.7)$} & 2 (H, L) & 2 (B, S) & 1 & --- & 4 & 9 (all main models) & 540\textsuperscript{a} & 1{,}755 & 15.0 \\
\midrule
\makecell[l]{R4: Reasoning effort\\(GPT-5.2 only)} & 2 (H, L) & 2 (B, S) & 4 & 3 efforts & 48 & 1 (low/med/high) & 480\textsuperscript{b} & 1{,}316 & 15.0 \\
\midrule
\makecell[l]{R5: Retail price 120\\(OpenAI models)} & 2 (H, L) & 2 (B, S) & 4 & --- & 16 & 3 (OpenAI family) & 720 & 2{,}328 & 15.0 \\
\midrule
\makecell[l]{R6: Prior sensitivity\\($\beta \in \{0.3, 0.5, 0.7\}$)} & 2 (H, L) & 2 (B, S) & 1 & 3 priors & 12 & 9 (all main models) & 1{,}080\textsuperscript{c} & 3{,}116 & 15.0 \\
\midrule
\makecell[l]{R7: Parameter size\\(Qwen3-14B)} & 2 (H, L) & 2 (B, S) & 4 & --- & 16 & 1 (Qwen3-14B) & 240\textsuperscript{d} & 1{,}027 & 15.0 \\
\midrule
\rowcolor{yellow!10}
\textbf{Extension subtotal (R1--R7)} & & & & & & & \textbf{5{,}760} & \textbf{19{,}102} & \\
\midrule
\textbf{Full program total} & & & & & & & \textbf{9{,}840} & \textbf{31{,}592} & \\
\bottomrule
\end{tabular}%
}
\par\smallskip
\begin{minipage}{\textwidth}
\scriptsize
\textit{Notes:} M1--M3 constitute the main analysis reported in the main text; R1--R7 are the design and robustness extensions reported in full in Appendix~\ref{sec:extension_program}. H = high-type buyer, L = low-type buyer. B = buyer proposes first, S = seller proposes first. Patience levels: ($\delta_B, \delta_S$) $\in$ \{(0.9, 0.9), (0.9, 0.4), (0.4, 0.9), (0.7, 0.9)\}. M3 counts 480 new cross-tier experiments; the 480 observations from the two homogeneous M1 benchmarks displayed in Table~\ref{tab:ext3_summary} are not counted again. Efforts: low, medium, high reasoning. \textsuperscript{a}540 new strategic-patience experiments; the full analysis reuses 2,160 M1 observations ($N=2{,}700$). \textsuperscript{b}480 new experiments (low + high effort); the medium-effort baseline (+240) is reused from M1 and not counted in the R4 row or in the grand total. \textsuperscript{c}1{,}080 new experiments ($\beta = 0.3$ and $\beta = 0.7$); the $\beta = 0.5$ baseline (+540) is reused from M1 and not counted in the R6 row or in the grand total. \textsuperscript{d}240 new Qwen3-14B experiments; the Qwen3-32B and Qwen3-Max legs of the $N=720$ three-size comparison reported in Section~\ref{subsec:parameter_size} (+480) are reused from M1 self-play data and not counted in the R7 row or in the grand total.
\end{minipage}
\end{table}

\clearpage
\subsection{Structured Communication (R1): Model-Level Detail}
\label{subsec:r1_structured}

This subsection reports the detailed results underlying the structured-communication treatment (R1) summarized in Section~\ref{subsec:r1_maintext}. Table~\ref{tab:ext1_struct_vs_verbal} gives the full performance comparison across agreement, rounds, and efficiency by model tier, first proposer, buyer type, and patience, with between-treatment tests. Removing verbal communication measurably changes execution but does not overturn the main performance pattern. Structured-only bargaining lowers agreement overall ($\chi^2=27.41$, $p<0.001$), increases rounds modestly ($|t|=3.23$, $p<0.01$), and reduces undiscounted efficiency ($|t|=5.05$, $p<0.001$). The agreement decline is concentrated outside the flagship tier, while efficiency falls significantly in each tier; nevertheless, structured agents still reach high absolute agreement ($96.4\%$) and efficiency ($92.8\%$), with only a modest increase in rounds ($2.98$ to $3.15$). Table~\ref{tab:ext1_model_detail} then reports, for each model, buyer surplus share, efficiency, and rounds under the verbal and structured treatments, with within-model significance tests.

\begin{table}[htbp]
\centering
\caption{Structured vs.\ Verbal Agents}
\label{tab:ext1_struct_vs_verbal}
\footnotesize
\renewcommand{\arraystretch}{1}
\begin{tabular}{@{}lccc|ccc@{}}
\toprule
\textbf{Condition} & \multicolumn{3}{c}{\textbf{Structured Agents}} & \multicolumn{3}{c}{\textbf{Verbal Agents}} \\
& \textbf{Agreement} & \textbf{Rounds} & \textbf{Efficiency} & \textbf{Agreement} & \textbf{Rounds} & \textbf{Efficiency} \\
& (\%) & & (\%) & (\%) & & (\%) \\
\midrule
\textbf{Overall} & \textbf{96.4} & \textbf{3.15} & \textbf{92.8} & \textbf{98.9} & \textbf{2.98} & \textbf{95.4} \\
\midrule
\multicolumn{7}{l}{\textit{By Model Tier:}} \\
\quad Flagship & 100.0 & 3.23 & 98.3 & 100.0 & 3.25 & 98.9 \\
\quad Mid-tier & 98.8 & 3.39 & 92.5 & 99.9 & 2.93 & 96.4 \\
\quad Baseline & 90.6 & 2.81 & 87.7 & 96.8 & 2.75 & 91.0 \\
\midrule
\multicolumn{7}{l}{\textit{By First Proposer:}} \\
\quad Buyer First & 95.5 & 3.10 & 92.8 & 98.4 & 2.97 & 94.8 \\
\quad Seller First & 97.4 & 3.20 & 92.9 & 99.4 & 2.99 & 96.1 \\
\midrule
\multicolumn{7}{l}{\textit{By Buyer Type:}} \\
\quad High-type & 97.0 & 2.94 & 93.0 & 99.5 & 2.88 & 95.1 \\
\quad Low-type & 95.8 & 3.36 & 92.7 & 98.2 & 3.08 & 95.7 \\
\midrule
\multicolumn{7}{l}{\textit{By Patience:}} \\
\quad (0.9, 0.9) & 97.8 & 3.57 & 94.7 & 98.0 & 3.39 & 95.2 \\
\quad (0.7, 0.9) & 96.9 & 3.33 & 93.1 & 99.4 & 3.09 & 96.1 \\
\quad (0.4, 0.9) & 94.6 & 2.86 & 90.2 & 99.3 & 2.63 & 95.0 \\
\quad (0.9, 0.4) & 96.5 & 2.83 & 93.4 & 98.9 & 2.82 & 95.5 \\
\midrule
\multicolumn{7}{l}{\textit{Between-Treatment Tests (Structured vs Verbal)}} \\
\quad Overall & 27.41*** & 3.23** & 5.05*** & --- & --- & --- \\
\quad By Model Tier: & & & & & & \\
\qquad Flagship & 0.00 & 0.31 & 2.58** & --- & --- & --- \\
\qquad Mid-tier & 4.93* & 4.54*** & 5.18*** & --- & --- & --- \\
\qquad Baseline & 22.71*** & 0.52 & 2.57* & --- & --- & --- \\
\quad By First Proposer: & & & & & & \\
\qquad Buyer First & 15.02*** & 1.69$^\dagger$ & 2.51* & --- & --- & --- \\
\qquad Seller First & 11.62*** & 2.89** & 4.94*** & --- & --- & --- \\
\quad By Buyer Type: & & & & & & \\
\qquad High-type & 18.59*** & 0.86 & 3.24** & --- & --- & --- \\
\qquad Low-type & 10.06** & 3.69*** & 3.88*** & --- & --- & --- \\
\bottomrule
\end{tabular}
\par\smallskip
\begin{minipage}{\textwidth}
\scriptsize
\textit{Notes:} Between-treatment test statistics: $\chi^2$ for Agreement, $|t|$ for Rounds and Efficiency. $^\dagger p < 0.10$, * $p < 0.05$, ** $p < 0.01$, *** $p < 0.001$. Efficiency measured as percentage of first-best undiscounted surplus.
\end{minipage}
\end{table}

\begin{table}[htbp]
\centering
\caption{Model-Level Verbal vs.\ Structured Comparison}
\label{tab:ext1_model_detail}
\footnotesize
\renewcommand{\arraystretch}{1.1}
\begin{tabular}{llcccccc}
\toprule
 & & \multicolumn{2}{c}{Buyer Share (\%)} & \multicolumn{2}{c}{Efficiency (\%)} & \multicolumn{2}{c}{Rounds} \\
\cmidrule(lr){3-4} \cmidrule(lr){5-6} \cmidrule(lr){7-8}
Model & Family & Verbal & Structured & Verbal & Structured & Verbal & Structured \\
\midrule
GPT-4o-mini & OpenAI & 39.8 & 64.1$^{***}$ & 84.6 & 74.7$^{**}$ & 3.40 & 3.52 \\
GPT-5-mini & OpenAI & 41.9 & 51.3$^{**}$ & 94.0 & 92.9 & 2.43 & 3.63$^{***}$ \\
GPT-5.2 & OpenAI & 37.9 & 40.5 & 98.2 & 97.0$^{*}$ & 2.68 & 2.91$^{*}$ \\
Gemini-2.5-Flash & Google & 44.3 & 41.7 & 98.7 & 97.1 & 2.45 & 2.82$^{*}$ \\
Gemini-3-Flash & Google & 51.4 & 45.9$^{***}$ & 99.1 & 98.9 & 3.11 & 3.24 \\
Gemini-3-Pro & Google & 53.9 & 50.2$^{*}$ & 99.8 & 100.0$^{***}$ & 3.31 & 3.31 \\
Qwen2.5-14B & Qwen & 52.7 & 50.0 & 89.8 & 91.4 & 2.48 & 2.19$^{\dagger}$ \\
Qwen3-32B & Qwen & 66.8 & 60.2 & 96.1 & 85.7$^{***}$ & 3.26 & 3.30 \\
Qwen3-Max & Qwen & 91.4 & 91.0 & 98.7 & 97.9$^{*}$ & 3.75 & 3.45$^{*}$ \\
All models &  & 53.5 & 54.8 & 95.4 & 92.8 & 2.98 & 3.15 \\
\bottomrule
\end{tabular}
\par\smallskip
\begin{minipage}{\textwidth}
\scriptsize
\textit{Notes:} Significance stars on structured values: $t$-test comparing structured vs verbal within each model. $^\dagger p < 0.10$, * $p < 0.05$, ** $p < 0.01$, *** $p < 0.001$.
\end{minipage}
\end{table}

Pooled buyer share changes little (53.5\% verbal vs.\ 54.8\% structured), but model-level distributional shifts diverge. OpenAI's mid-tier and baseline models shift surplus \textit{toward} buyers when communication is removed: GPT-5-mini's buyer share rises from 41.9\% to 51.3\% ($d = -0.27$, $p = 0.003$), and GPT-4o-mini shifts even more sharply (39.8\% $\to$ 64.1\%, $d = -0.33$, $p = 0.001$). These models use verbal messages in ways that moderate the buyer advantage: their sellers negotiate more effectively with language than without it. Gemini's mid-tier and flagship models move in the opposite direction: Gemini-3-Flash buyer share drops from 51.4\% to 45.9\% ($d = 0.30$, $p = 0.001$), and Gemini-3-Pro from 53.9\% to 50.2\% ($d = 0.20$, $p = 0.033$); for these two models, verbal communication amplifies the buyer advantage rather than moderating it. The remaining five models, GPT-5.2, Gemini-2.5-Flash, Qwen2.5-14B, Qwen3-32B, and Qwen3-Max, show no detectable buyer-share shift. Thus R1 rules out a purely language-based explanation for the main distributional profiles while showing that the verbal channel remains an execution aid and a model-specific distributional lever.

\clearpage
\subsection{No-Discounting Bargaining (R2)}
\label{app:ext_unprompted}
This subsection reports the full no-discounting bargaining extension (R2) summarized in Section~\ref{subsec:r2_maintext} of the main paper. It gives the design, interpretive frame, and the complete set of empirical results behind the headline comparison (Table~\ref{tab:ext_unprompted_efficiency_delay}), including the cuts by first-proposer role and buyer type (Table~\ref{tab:ext_unprompted_summary_nodup}).

\begin{table}[htbp]
\centering
\caption{Negotiation Performance: Symmetric Patience ($\delta=(0.9,0.9)$) vs.\ No Discounting}
\label{tab:ext_unprompted_efficiency_delay}
\scriptsize
\renewcommand{\arraystretch}{1.15}
\setlength{\tabcolsep}{3pt}
\begin{tabular*}{\linewidth}{@{\extracolsep{\fill}}lccccc|ccccc@{}}
\toprule
\textbf{Group} & \multicolumn{5}{c}{\textbf{Symmetric} $\delta=(0.9,0.9)$} & \multicolumn{5}{c}{\textbf{No Discounting}} \\
& \textbf{Eff.} & \textbf{Rounds} & \textbf{Round 1} & \textbf{Buyer} & \textbf{Irr.} & \textbf{Eff.} & \textbf{Rounds} & \textbf{Round 1} & \textbf{Buyer} & \textbf{Irr.} \\
& & & \textbf{accept} & \textbf{share} & \textbf{rate} & & & \textbf{accept} & \textbf{share} & \textbf{rate} \\
& (\%) & & (\%) & (\%) & (\%) & (\%) & & (\%) & (\%) & (\%) \\
\midrule
\multicolumn{11}{@{}l}{\textit{By Tier:}} \\
\quad Baseline & 89.1 & 2.83 & 10.0 & 41.5 & 22.4 & 89.8 & 3.66 & 1.1 & 46.4 & 16.5 \\
\quad Mid-tier & 97.3 & 3.42 & 15.0 & 50.7 & 0.6 & 95.9 & 3.92 & 11.1 & 53.7 & 0.0 \\
\quad Flagship & 99.2 & 3.88 & 0.6 & 61.9 & 0.0 & 99.0 & 4.45 & 0.6 & 61.8 & 0.0 \\
\midrule
\multicolumn{11}{@{}l}{\textit{By Provider:}} \\
\quad OpenAI & 90.6 & 3.22 & 5.0 & 32.4 & 15.9 & 90.1 & 3.79 & 3.3 & 40.2 & 12.4 \\
\quad Google & 99.6 & 3.84 & 5.6 & 51.0 & 0.6  & 99.6 & 4.91 & 0.0 & 48.6 & 0.0 \\
\quad Qwen   & 95.3 & 3.09 & 15.0 & 70.2 & 6.1  & 95.0 & 3.33 & 9.4 & 72.7 & 3.9 \\
\midrule
\multicolumn{11}{@{}l}{\textit{By Model:}} \\
\multicolumn{11}{@{}l}{\quad \textit{OpenAI:}} \\
\quad \quad GPT-5.2 & 98.8 & 3.60 & 0.0 & 39.3 & 0.0 & 98.4 & 4.70 & 0.0 & 38.7 & 0.0 \\
\quad \quad GPT-5-mini & 95.8 & 2.72 & 15.0 & 35.9 & 0.0 & 93.3 & 3.13 & 10.0 & 37.5 & 0.0 \\
\quad \quad GPT-4o-mini & 77.4 & 3.36 & 0.0 & 20.1 & 54.0 & 78.7 & 3.50 & 0.0 & 45.2 & 42.0 \\
\multicolumn{11}{@{}l}{\quad \textit{Google:}} \\
\quad \quad Gemini-3-Pro & 99.9 & 4.33 & 0.0 & 56.0 & 0.0 & 100.0 & 4.83 & 0.0 & 54.8 & 0.0 \\
\quad \quad Gemini-3-Flash & 99.8 & 4.42 & 0.0 & 51.8 & 0.0 & 99.8 & 5.28 & 0.0 & 52.5 & 0.0 \\
\quad \quad Gemini-2.5-Flash & 99.1 & 2.77 & 16.7 & 45.2 & 1.7 & 99.0 & 4.62 & 0.0 & 38.6 & 0.0 \\
\multicolumn{11}{@{}l}{\quad \textit{Qwen:}} \\
\quad \quad Qwen3-Max & 99.0 & 3.70 & 1.7 & 90.4 & 0.0 & 98.7 & 3.82 & 1.7 & 91.9 & 0.0 \\
\quad \quad Qwen3-32B & 96.2 & 3.14 & 30.0 & 64.4 & 1.7 & 94.6 & 3.35 & 23.3 & 71.1 & 0.0 \\
\quad \quad Qwen2.5-14B & 90.7 & 2.45 & 13.3 & 55.6 & 16.7 & 91.6 & 2.83 & 3.3 & 55.1 & 11.7 \\
\midrule
\textbf{Overall} & \textbf{95.2} & \textbf{3.39} & \textbf{8.5} & \textbf{51.5} & \textbf{7.4} & \textbf{94.9} & \textbf{4.02} & \textbf{4.3} & \textbf{54.1} & \textbf{5.3} \\
\bottomrule\\
\end{tabular*}

\begin{tabular*}{\linewidth}{@{\extracolsep{\fill}}lccccc@{}}
\multicolumn{6}{@{}l}{\textit{Between-treatment tests (Symmetric vs No Discounting): $|t|$ Welch / $\chi^2$}} \\
\addlinespace[0.2em]
& \textbf{Eff.} & \textbf{Rounds} & \textbf{Round-1 accept} & \textbf{Buyer share} & \textbf{Irr. rate} \\
\midrule
\quad Overall & 0.29 & 6.49*** & 8.19** & 1.25 & 1.95 \\
\multicolumn{6}{@{}l}{\quad By Tier:} \\
\quad \quad Baseline & 0.30 & 4.48*** & 13.55*** & 1.02 & 1.88 \\
\quad \quad Mid-tier & 1.21 & 2.79** & 1.20 & 1.05 & 1.01 \\
\quad \quad Flagship & 0.70 & 4.72*** & 0.00 & 0.04 & --- \\
\bottomrule\\
\end{tabular*}

\begin{tabular*}{\linewidth}{@{\extracolsep{\fill}}p{0.62\linewidth}l@{}}
\multicolumn{2}{@{}l}{\textit{Provider-level buyer-share tests (Symmetric vs No Discounting): $|t|$ Welch on raw share}} \\
\addlinespace[0.2em]
\quad OpenAI pooled & 1.73$^\dagger$ \\
\quad Google pooled & 1.86$^\dagger$ \\
\quad Qwen pooled & 0.78 \\
\addlinespace[0.2em]
\quad Treatment $\times$ Provider interaction (OLS $F$-test) & 2.44$^\dagger$ \\
\addlinespace[0.6em]
\multicolumn{2}{@{}l}{\textit{Capability gradient: Cochran--Armitage linear-trend test on irrationality across tiers}} \\
\addlinespace[0.2em]
\quad Symmetric $\delta=(0.9,0.9)$ & $|Z|=7.93***$ \quad ($p=0.000$) \\
\quad No discounting & $|Z|=6.82***$ \quad ($p=0.000$) \\
\bottomrule
\end{tabular*}
\par\smallskip
\begin{minipage}{\textwidth}
\scriptsize
\textit{Notes:} $N=60$ per (model, treatment); $N=540$ per treatment. Round-1 accept: share agreeing in the first round. Buyer share: buyer's share of realized undiscounted surplus. Irr.\ rate: share of agreed deals where either party accepted a contract with negative undiscounted profit. Efficiency and first-round acceptance averaged over all experiments; rounds-to-agreement, buyer share, and irrationality rate averaged over agreed deals only. All quantities computed on an undiscounted basis to compare treatments on common footing (the no-discounting treatment makes discounted and undiscounted profits identical by construction). Test column reports $|t|$ (Welch) for Eff., Rounds, and Buyer share; $\chi^2$ (Pearson) for Round-1 accept and Irr.\ rate. Cochran--Armitage trend test uses scores $\{0,1,2\}$ on Either-irrational across tiers (Baseline $\to$ Mid $\to$ Flagship), reported separately within each treatment. $^\dagger p<0.10$, * $p<0.05$, ** $p<0.01$, *** $p<0.001$.
\end{minipage}
\end{table}

\subsubsection{Design}

 We retain the main-analysis factorial structure (two buyer types, two proposer assignments, nine models, fifteen replications per cell) but modify both buyer and seller prompts to remove all references to $\delta$, patience, effective utility, and time pressure. Agents are told that they negotiate over multiple rounds with a ten-round cap, that failure to agree yields zero, and that profit is computed from the contract terms as in the main analysis. No discount factor appears in the prompt or in the utility computation. The resulting design yields $N = 540$ negotiations. Appendix~\ref{app:prompts_r2} reports the modified prompt text, with removed content shown in strikethrough.

\subsubsection{Interpretive Frame}

We present R2 as an external-validity test of the main findings rather than as a mechanism test. Removing discounting from the prompt changes several features simultaneously: the utility formula, the patience concept, and the cue that time matters. We do not attempt to separate their contributions. The informative question is whether the qualitative patterns of the main analysis (the capability-patience driver separation, provider-specific surplus profiles, capability-dependent irrationality) replicate under a prompting regime that omits the discounting framework. Because no imposed $\delta$ structure exists in R2, there is no Feng et al.\ benchmark for surplus division; we compare against the symmetric high-patience $(0.9, 0.9)$ cell as the closest main-analysis analogue.

\subsubsection{Empirical Results}

Table~\ref{tab:ext_unprompted_efficiency_delay} compares the
symmetric-patience baseline with the no-discounting treatment
(no $\delta$ in the prompt or payoff computation) on five outcomes:
undiscounted efficiency, rounds-to-agreement, first-round acceptance, buyer share, and irrationality rate, all reported on a common undiscounted basis so the two treatments are compared on equal footing. The three panels serve distinct purposes. Panel~A maps the descriptive landscape, reporting each outcome by tier, by provider,
by model, and overall, for both treatments.\footnote{Patterns by first-proposer role and buyer type are stable on agreement and efficiency, while rounds increase significantly in every reported cut; Table~\ref{tab:ext_unprompted_summary_nodup} reports these comparisons.} Panel~B tests whether the treatments differ in central tendency, overall and within each tier, using Welch $|t|$ tests for continuous outcomes and $\chi^2$ tests for binary outcomes. Panel~C tests for heterogeneity: provider-level Welch tests, a Treatment $\times$ Provider interaction $F$-test on raw buyer share, and a Cochran--Armitage linear-trend test \citep{cochran1954some, armitage1955tests} for irrationality across capability tiers within each treatment.

The next three paragraphs each draw on one slice of the table: the high-efficiency-with-delay pattern (Panels~A and~B), the cross-provider distributional profile (the \textit{By Provider} block of Panel~A together with the interaction in Panel~C), and the capability--irrationality gradient (the \textit{Irr.\ rate} column of Panel~A together with the trend test in Panel~C).

\paragraph{Efficiency and delay.} Allocative efficiency is statistically indistinguishable across the two treatments at the aggregate level ($95.2\%$ vs.\ $94.9\%$; $|t| = 0.29$) and within every capability tier. Delay, by contrast, increases broadly: rounds-to-agreement rise from $3.39$ to $4.02$ in the pooled sample ($|t| = 6.49$, $p < 0.001$), from $2.83$ to $3.66$ among baseline models ($|t| = 4.48$, $p < 0.001$), from $3.42$ to $3.92$ among mid-tier models ($|t| = 2.79$, $p < 0.01$), and from $3.88$ to $4.45$ among flagship models ($|t| = 4.72$, $p < 0.001$). Round-1 acceptance also falls from $8.5\%$ to $4.3\%$ ($\chi^2 = 8.19$, $p < 0.01$). Removing the discount factor therefore acts as a tempo cue, slowing closure without altering allocative efficiency.

\paragraph{Provider-specific distributional profiles.} The \textit{By Provider} block of Table~\ref{tab:ext_unprompted_efficiency_delay} reports realized buyer surplus shares pooled within each model family. We report raw buyer share rather than deviation from a PBE benchmark because the \citet{feng2015dynamic} PBE buyer share is undefined when $\delta_B \delta_S = 1$ (the alternating-offer game admits a continuum of subgame-perfect divisions when delay is costless), so a $\Delta$-against-PBE construct is not portable across the two treatments. The provider-level pattern from Section~\ref{sec:provider} reappears under R2: OpenAI remains seller-favoring (buyer share $32.4\% \to 40.2\%$), Google tracks an even split ($51.0\% \to 48.6\%$), and Qwen remains strongly buyer-favoring ($70.2\% \to 72.7\%$). The Treatment $\times$ Provider interaction is marginal ($F = 2.44$, $p = 0.087$; Panel C), but the rank ordering OpenAI $\prec$ Google $\prec$ Qwen is preserved. The cross-family seller-favoring/near-neutral/buyer-favoring profile is therefore a property of the underlying agents rather than an artifact of how patience was prompted.

\paragraph{Capability-dependent irrationality gradient.} The \textit{Irr.\ rate} column of Table~\ref{tab:ext_unprompted_efficiency_delay} reports, by tier and treatment, the share of agreed deals in which either party accepted a contract yielding negative undiscounted profit. The capability ordering from the main analysis survives the removal of discounting: irrationality declines monotonically from baseline ($22.4\%$ symmetric, $16.5\%$ no-discount) through mid-tier ($0.6\%$, $0.0\%$) to flagship ($0.0\%$, $0.0\%$). A Cochran--Armitage linear-trend test confirms the gradient within each treatment (symmetric: $|Z| = 7.93$; no-discount: $|Z| = 6.82$; both $p < 0.001$; Panel C). Baseline irrationality is if anything lower under no discounting ($16.5\%$ versus $22.4\%$), and the within-tier between-treatment $\chi^2$ tests in Panel B are nonsignificant (baseline $\chi^2 = 1.88$; mid-tier essentially unchanged, $\chi^2 = 1.01$); flagship has no between-treatment variation to test, since both treatments show $0.0\%$ irrationality. Capability, not prompt structure, governs whether agents accept loss-making contracts. Removing discounting leaves the capability gradient intact, indicating that rationality at the contract level is an agent property.

\begin{table}[htbp]
\centering
\caption{Negotiation Performance by Other Dimensions: Symmetric Patience vs.\ No Discounting}
\label{tab:ext_unprompted_summary_nodup}
\footnotesize
\renewcommand{\arraystretch}{1.1}
\begin{tabular}{@{}lccc|ccc@{}}
\toprule
\textbf{Condition} & \multicolumn{3}{c}{\textbf{Symmetric Patience} $\delta=(0.9,0.9)$} & \multicolumn{3}{c}{\textbf{No Discounting}} \\
& \textbf{Agreement} & \textbf{Rounds} & \textbf{Efficiency} & \textbf{Agreement} & \textbf{Rounds} & \textbf{Efficiency} \\
& (\%) & & (\%) & (\%) & & (\%) \\
\midrule
\multicolumn{7}{@{}l}{\textit{By First Proposer:}} \\
\quad Buyer first & 97.4 & 3.38 & 94.6 & 97.8 & 4.13 & 95.0 \\
\quad Seller first & 98.5 & 3.40 & 95.8 & 98.5 & 3.91 & 94.8 \\
\midrule
\multicolumn{7}{@{}l}{\textit{By Buyer Type:}} \\
\quad High-type & 98.9 & 3.33 & 95.3 & 98.9 & 4.03 & 95.2 \\
\quad Low-type & 97.0 & 3.45 & 95.1 & 97.4 & 4.00 & 94.6 \\
\midrule
\multicolumn{7}{@{}l}{\textit{Between-Treatment Tests (Symmetric vs No Discounting)}} \\
\addlinespace[0.2em]
\multicolumn{7}{@{}l}{\quad By First Proposer:} \\
\quad \quad Buyer first & 0.08 & 5.29*** & 0.31 & --- & --- & --- \\
\quad \quad Seller first & 0.00 & 3.85*** & 0.81 & --- & --- & --- \\
\multicolumn{7}{@{}l}{\quad By Buyer Type:} \\
\quad \quad High-type & 0.00 & 5.40*** & 0.04 & --- & --- & --- \\
\quad \quad Low-type & 0.07 & 3.85*** & 0.33 & --- & --- & --- \\
\bottomrule
\end{tabular}
\par\smallskip
\begin{minipage}{\textwidth}
\scriptsize
\textit{Notes:} $N=60$ per (model, treatment); $N=540$ per treatment. Companion to Table~\ref{tab:ext_unprompted_efficiency_delay}, which reports Overall and By Tier; this table presents only the By First Proposer and By Buyer Type cuts to avoid duplication. Agreement reported over all experiments; Rounds over agreed deals only; Efficiency over all experiments (failed deals contribute 0). Efficiency is undiscounted realized surplus relative to first-best. Between-treatment test statistics: $\chi^2$ \citep{pearson1900criterion} for Agreement, $|t|$ (Welch \citeyearpar{welch1947generalization}) for Rounds and Efficiency. Significance: $^\dagger p<0.10$, * $p<0.05$, ** $p<0.01$, *** $p<0.001$.
\end{minipage}
\end{table}

\paragraph{Deployment interpretation.} R2 is a closer analogue to production-style prompts that specify contractual terms and objectives without parameterizing agent utility with a time-preference coefficient. On the three structural margins that organize the main analysis, the qualitative patterns persist: allocative efficiency, the raw cross-provider distributional ordering, and the capability-irrationality gradient all reproduce. The main detectable change is longer bargaining, which carries no allocative-efficiency loss in the no-discounting treatment because delay is not penalized. The extension does not replace the main-analysis benchmark against the PBE of \citet{feng2015dynamic}, which remains necessary for interpreting deviations from normative predictions. Instead, it shows that those deviations persist when explicit discount-factor language is omitted, while that language mainly affects negotiation tempo.

\clearpage
\subsection{Reasoning Effort Ablation (R4)}
\label{subsec:reasoning_effort}

This extension probes whether additional inference-time computation materially changes bargaining outcomes within a fixed model. If more reasoning per turn sharpened strategic execution, we would expect more aggressive opening offers, faster convergence, or tighter surplus splits. We test this by running GPT-5.2 at low, medium, and high reasoning effort levels across all 16 conditions ($N = 720$ experiments).

Reasoning effort affects negotiation tempo but not undiscounted allocative performance. Mean rounds decline from 2.95 at low effort to 2.68 at medium effort and 2.53 at high effort ($F = 11.22$, $p = 1.6 \times 10^{-5}$). Discounted efficiency rises from 70.7\% to 76.9\% to 78.5\% ($F = 24.59$, $p = 4.7 \times 10^{-11}$). For both outcomes, the low--medium and low--high differences survive Holm correction, whereas the medium--high difference does not. Undiscounted efficiency ($F = 1.68$, $p = 0.188$) and buyer share ($F = 2.21$, $p = 0.110$) do not differ in the run-level omnibus tests. Deal rates are 100\% at every effort level. Table~\ref{tab:ext4_summary} reports the full results.

\begin{table}[htbp]
\centering
\footnotesize
\renewcommand{\arraystretch}{1.1}
\caption{Reasoning Effort Ablation: Summary Statistics (GPT-5.2)}
\label{tab:ext4_summary}
\begin{tabular}{lccccccc}
\toprule
Effort & $N$ & Deal Rate & Rounds & Efficiency & DiscEfficiency & Buyer Share \\
\midrule
Low & 240 & 100.0\% & 2.95 & 97.2\% & 70.7\% & 41.4\% \\
Medium & 240 & 100.0\% & 2.68 & 98.2\% & 76.9\% & 37.9\% \\
High & 240 & 100.0\% & 2.53 & 97.7\% & 78.5\% & 36.5\% \\
\midrule
Run-level ANOVA $F$ & & & 11.22 & 1.68 & 24.59 & 2.21 \\
Omnibus $p$-value & & & $<0.001$ & 0.188 & $<0.001$ & 0.110 \\
\bottomrule
\end{tabular}
\par\smallskip
\begin{minipage}{\textwidth}
\scriptsize
\textit{Notes:} We use DiscEfficiency to denote discounted efficiency. The omnibus tests are ordinary one-way ANOVAs at the negotiation level. Pairwise comparisons use Welch tests with Holm correction across the three effort-level contrasts within each outcome.
\end{minipage}
\end{table}

Across all effort levels, buyer surplus share stays within a 36.5\%--41.4\% band, and no buyer-share pairwise comparison survives Holm correction. Thus, within GPT-5.2 and over the effort levels tested here, additional inference-time computation changes negotiation speed but not the division of surplus in the run-level analysis.

The ablation therefore rejects a blanket null effect of reasoning effort. Greater effort primarily reduces negotiation delay and improves time-adjusted performance, without significantly improving undiscounted allocative efficiency. This pattern is not uniquely diagnostic of any single internal mechanism.

\clearpage
\subsection{Retail Price Robustness (R5)}
\label{subsec:retail_price_robustness}

The main analysis establishes its findings at a retail price of 60 (margin $=30$, 50\% gross margin). A natural concern is whether these patterns (high efficiency with delayed agreement, provider-specific surplus division, and capability threshold) are specific to this particular surplus level or represent general properties of LLM bargaining. If surplus division patterns stem from training-encoded behavioral tendencies rather than strategic calculation, they should be relatively insensitive to the absolute size of the surplus: a rational agent would negotiate harder when more money is on the table, but a training-driven bias should produce similar percentage shares regardless of magnitude. We test this by tripling the margin to 90 (retail price 120, 75\% gross margin), creating a larger surplus to divide. Only OpenAI models (GPT-5.2, GPT-5-mini, GPT-4o-mini) are included. We compare their RP120 results ($N = 720$) against their RP60 baseline ($N = 720$), with 15 replications per condition in both.

\paragraph{The high-efficiency-with-delay pattern generalizes across price levels.} Undiscounted efficiency remains high at both price levels, dipping modestly from 92.3\% at RP60 to 89.5\% at RP120 ($p = 0.014$); discounted efficiency is statistically equivalent (66.7\% vs.\ 64.2\%, $p = 0.070$). The agents' ability to find efficient deals is largely insensitive to the size of the surplus available.

\paragraph{Buyer advantage responds differently by capability tier.} Buyer surplus share rises from 39.9\% at RP60 to 52.7\% at RP120. As Table~\ref{tab:ext5_summary} and Figure~\ref{fig:ext5_comparison} show, this shift is driven mainly by GPT-4o-mini, whose buyer share jumps from 39.8\% to 67.1\% ($p < 0.001$); GPT-5-mini shows a smaller but significant rise (41.9\% $\to$ 50.4\%, $p = 0.006$), while GPT-5.2 is essentially flat (37.9\% $\to$ 40.7\%, $p = 0.254$).

This pattern is descriptive rather than mechanically identified. The flagship model maintains a comparatively stable buyer share across the larger-surplus environment, while GPT-5-mini and especially GPT-4o-mini become more buyer-favoring at RP120. This tier gradient is consistent with greater environment sensitivity below the capability threshold, but the design does not identify the internal mechanism generating the response.

Notably, GPT-4o-mini's higher buyer share translates into significantly higher \textit{absolute} buyer profit at RP120. Among rational deals (excluding agreements in which either party accepts negative profit), GPT-4o-mini earns \$3,349 per deal compared to \$2,059 for GPT-5.2 ($t = 7.62$, $p < 0.001$), with a buyer share gap of 23.9 percentage points ($t = 9.14$, $p < 0.001$). Even accounting for GPT-4o-mini's 11.8\% irrational deal rate and 15.4\% deal failure rate, its expected buyer profit (\$2,782) still significantly exceeds GPT-5.2's (\$2,059; $t = 4.23$, $p < 0.001$). This counterintuitive result arises because GPT-5.2 negotiates ``fairer'' deals that allocate more surplus to the seller (\$2,823 vs.\ \$1,699 for GPT-4o-mini), while GPT-4o-mini's aggressive buyer behavior, though less reliable, captures more value when it succeeds. The implication for AI-mediated procurement is that deploying a more capable model as buyer does not necessarily maximize buyer value; capability promotes balanced outcomes rather than partisan ones.

\begin{table}[htbp]
\centering
\footnotesize
\renewcommand{\arraystretch}{1.1}
\caption{Retail Price Comparison: Summary Statistics (OpenAI Models)}
\label{tab:ext5_summary}
\begin{tabular}{clccccc}
\toprule
Retail Price & Model & $N$ & Deal Rate & Rounds & Efficiency & Buyer Share \\
\midrule
60  & GPT-4o-mini & 240 & 90.8\% & 3.40 & 84.6\% & 39.8\% \\
60  & GPT-5-mini  & 240 & 100\%  & 2.43 & 94.0\% & 41.9\% \\
60  & GPT-5.2     & 240 & 100\%  & 2.68 & 98.2\% & 37.9\% \\
\midrule
120 & GPT-4o-mini & 240 & 84.6\% & 3.82 & 79.2\% & 67.1\% \\
120 & GPT-5-mini  & 240 & 100\%  & 2.18 & 91.7\% & 50.4\% \\
120 & GPT-5.2     & 240 & 100\%  & 2.75 & 97.6\% & 40.7\% \\
\bottomrule
\end{tabular}
\par\smallskip
\begin{minipage}{\textwidth}
\scriptsize
\textit{Notes:} $N=240$ per (model, retail price). Deal Rate is over all experiments; Rounds is over agreed deals only; Efficiency is undiscounted realized surplus relative to first-best (failed deals contribute 0); Buyer Share is the buyer's share of realized undiscounted surplus among agreed deals.
\end{minipage}
\end{table}

\begin{figure}[H]
\centering
\includegraphics[width=0.95\textwidth]{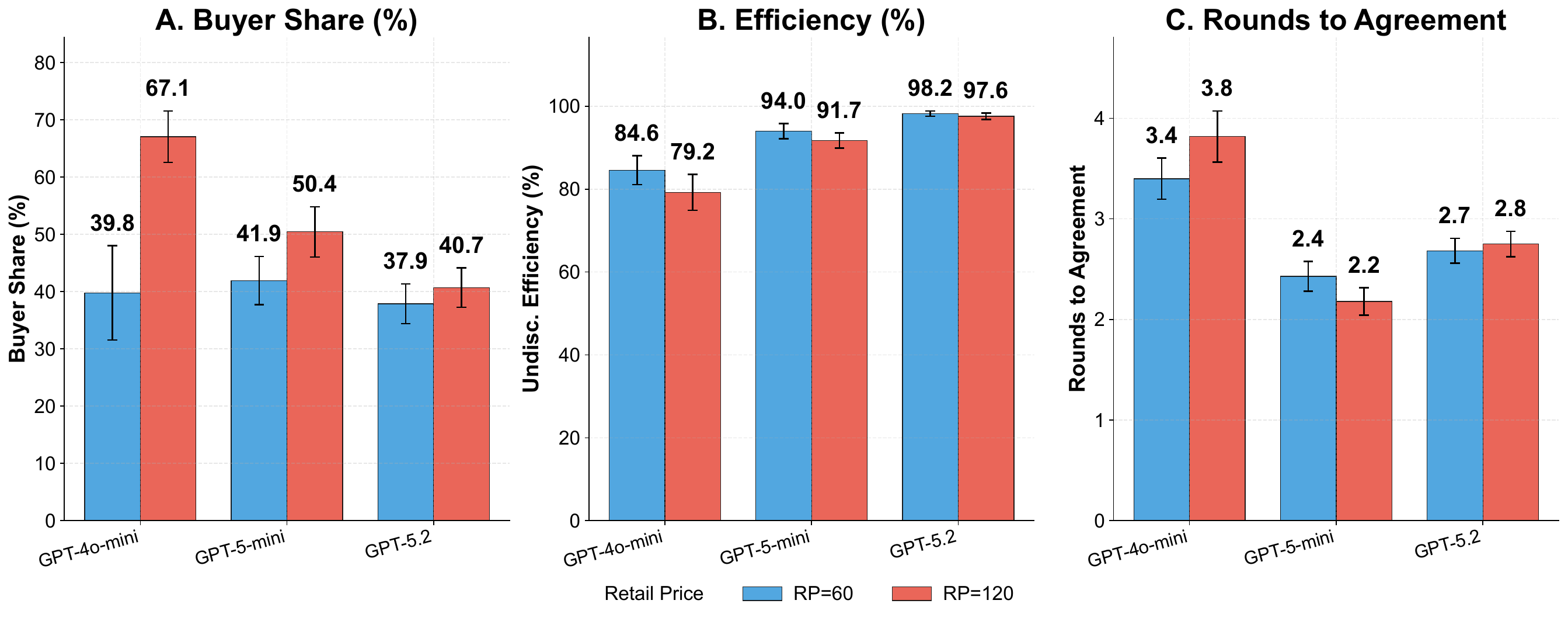}
\caption{Retail Price Comparison Across OpenAI Models}
\par\smallskip
\begin{minipage}{\textwidth}
\scriptsize
\textit{Notes:} RP60 (blue) vs.\ RP120 (red) for buyer surplus share, undiscounted efficiency, and rounds to agreement.
\end{minipage}
\label{fig:ext5_comparison}
\end{figure}

\paragraph{Patience effects and negotiation speed at higher margins.} Negotiations conclude at a similar pace at RP120 (2.87 vs.\ 2.82 rounds, $t = 0.63$, $p = 0.53$). The qualitative pattern of patience effects is preserved: under buyer-patient conditions ($\delta_B = 0.9$, $\delta_S = 0.4$), buyer shares reach 75.5\% at RP120 (vs.\ 62.9\% at RP60); under seller-patient conditions ($\delta_B = 0.4$, $\delta_S = 0.9$), buyer shares fall to 37.4\% at RP120 (vs.\ 23.4\% at RP60). The direction of patience effects is consistent across price levels, though the magnitude of buyer-share variation is somewhat compressed at RP120.

\clearpage
\subsection{Sensitivity to Prior Probability of High-Type Buyer (R6)}
\label{subsec:prior_sensitivity}

The main analysis fixes the seller's prior belief about the buyer being high-type at $\beta = 0.5$. This extension varies the prior to $\beta = 0.3$ (seller believes buyer is more likely low-type) and $\beta = 0.7$ (seller believes buyer is more likely high-type), restricting the patience configuration to ($\delta_B = 0.4$, $\delta_S = 0.9$), where the buyer is impatient and the seller is patient. This condition is chosen because the theoretical equilibrium structure \citep{feng2015dynamic} is most prior-sensitive under low buyer patience: the thresholds that determine signaling, pooling, and screening regimes shift with the prior, and quantity distortion occurs in this regime. By contrast, under high buyer patience the buyer truth-tells with first-best quantities regardless of $\beta$, making the prior less consequential. All nine main-analysis models are included, with 15 replications per condition ($N = 540$ per prior level, $N = 1{,}620$ total).

\paragraph{Efficiency and deal rates are robust to prior beliefs.} Deal rates remain near-universal across prior levels (98.1\%, 99.3\%, and 97.8\% at $\beta = 0.3$, $0.5$, and $0.7$), with no meaningful difference. Undiscounted efficiency likewise stays high (92.7\%, 95.0\%, and 93.6\%), without a systematic trend in the prior. Table~\ref{tab:ext6_summary} reports the full results.

\begin{table}[htbp]
\centering
\footnotesize
\renewcommand{\arraystretch}{1.1}
\caption{Prior Sensitivity: Summary Statistics ($\delta_B = 0.4$, $\delta_S = 0.9$, All 9 Models Pooled)}
\label{tab:ext6_summary}
\begin{tabular}{ccccccc}
\toprule
Prior ($\beta$) & $N$ & Deal Rate & Rounds & Efficiency & Buyer Share & Wholesale Price \\
\midrule
0.3 & 540 & 98.1\% & 2.61 & 92.7 & 44.2\% & \$44.01 \\
0.5 & 540 & 99.3\% & 2.63 & 95.0 & 40.2\% & \$45.19 \\
0.7 & 540 & 97.8\% & 2.86 & 93.6 & 37.1\% & \$44.99 \\
\bottomrule
\end{tabular}
\par\smallskip
\begin{minipage}{\textwidth}
\scriptsize
\textit{Notes:} $N=540$ per prior level, pooled across all 9 models. Deal Rate, Rounds, Efficiency, and Buyer Share as defined in Table~\ref{tab:ext5_summary}. Wholesale Price is the mean agreed unit price (payment/quantity) among agreed deals.
\end{minipage}
\end{table}

\paragraph{Negotiation speed varies with prior.} The ANOVA for rounds is significant ($F = 4.78$, $p = 0.009$). Negotiations at $\beta = 0.7$ take the longest (2.86 rounds), followed by $\beta = 0.5$ (2.63 rounds) and $\beta = 0.3$ (2.61 rounds). Higher priors, where the seller believes the buyer is likely high-type, lead to longer negotiations as sellers bargain more aggressively.

\paragraph{Surplus division shifts with seller beliefs.} Buyer surplus share trends downward as the seller's prior about high-type increases: 44.2\% at $\beta = 0.3$, 40.2\% at $\beta = 0.5$, and 37.1\% at $\beta = 0.7$. The ANOVA is significant ($F = 3.21$, $p = 0.040$), and the directional pattern is consistent with rational Bayesian updating: a seller who believes the buyer is likely high-type should demand more, extracting a larger share.

\paragraph{Model families respond differently to priors.} The prior's effect varies across model families
. Gemini models show a monotonic decline in buyer share from $\beta = 0.3$ to $\beta = 0.7$ (31.1\% $\to$ 28.5\% $\to$ 19.2\%), consistent with directionally rational behavior: higher priors prompt harder seller bargaining. OpenAI models show the same monotonic pattern (28.1\% $\to$ 23.4\% $\to$ 20.1\%), declining steadily as the prior rises. Qwen models, by contrast, show no meaningful response to the prior. Buyer share remains near 70\% at all three prior levels (72.8\%, 68.3\%, 71.0\%), reflecting the strong buyer bias documented in the main analysis. The Qwen family's dominant buyer-favoring tendency overrides the informational content of the prior.

\clearpage
\subsection{Parameter Size Ablation (R7)}
\label{subsec:parameter_size}

The main analysis compares models across providers, confounding architecture, training data, and parameter count. This extension isolates the effect of model size by comparing three models from the same family: Qwen3-14B (14 billion parameters), Qwen3-32B (32 billion), and Qwen3-Max (flagship, undisclosed but substantially larger). All three share the Qwen3 architecture, allowing a cleaner size comparison. The extension covers the 16-condition factorial ($N = 720$ across the three sizes; 240 attempts per model).

\paragraph{Deal rates remain near-universal across model size.} All three sizes reach agreement in essentially all negotiations (14B 100.0\%, 32B 99.6\%, Max 100.0\%), so deal-making itself does not scale with size in this family; the reliability difference instead appears in the rate of economically irrational agreements (discussed below). Table~\ref{tab:ext7_summary} reports the full results.

\begin{table}[htbp]
\centering
\footnotesize
\renewcommand{\arraystretch}{1.1}
\caption{Parameter Size Ablation: Summary Statistics (Qwen3 Family; Irr.\ = Irrational Deal Rate)}
\label{tab:ext7_summary}
\begin{tabular}{lccccccl}
\toprule
Size & $N$ & Deal Rate & Rounds & Efficiency & DiscEfficiency & Buyer Share & Irr. \\
\midrule
14B & 240 & 100.0\% & 4.28 & 94.1\% & 51.6\% & 63.1\% & 5.4\% \\
32B & 240 & 99.6\% & 3.26 & 96.1\% & 62.6\% & 66.8\% & 1.7\% \\
Max & 240 & 100.0\% & 3.75 & 98.7\% & 50.8\% & 91.4\% & 0.0\% \\
\bottomrule
\end{tabular}
\par\smallskip
\begin{minipage}{\textwidth}
\scriptsize
\textit{Notes:} We use DiscEfficiency to denote discounted efficiency.
\end{minipage}
\end{table}

\paragraph{Negotiation speed varies modestly with model size.} The 14B model is the slowest to converge, averaging 4.28 rounds, compared with 3.75 for Max and 3.26 for 32B. The ANOVA for rounds across the three sizes is significant ($F = 18.0$, $p < 0.001$), with the 14B-vs-Max pairwise difference at $t = 3.32$, $p < 0.001$; 32B is also significantly faster than Max ($t = 3.15$, $p = 0.002$), so negotiation speed does not improve monotonically with size in this family. Figure~\ref{fig:ext7_verbal} shows the pattern across buyer share, efficiency, and rounds.

\paragraph{Efficiency and the cost of delay.} Undiscounted efficiency rises modestly with size (94.1\% at 14B, 96.1\% at 32B, 98.7\% at Max; $p < 0.001$ for the 14B-vs-Max contrast). Discounted efficiency, which penalizes the extra rounds, does not fall monotonically with size: 32B attains the highest discounted efficiency (62.6\%), with 14B (51.6\%) and Max (50.8\%) close together, mirroring the round counts since 32B is the fastest to converge. The smallest model's longer negotiations are not severe enough to devastate the value it eventually creates.

\begin{figure}[htbp]
\centering
\includegraphics[width=0.95\textwidth]{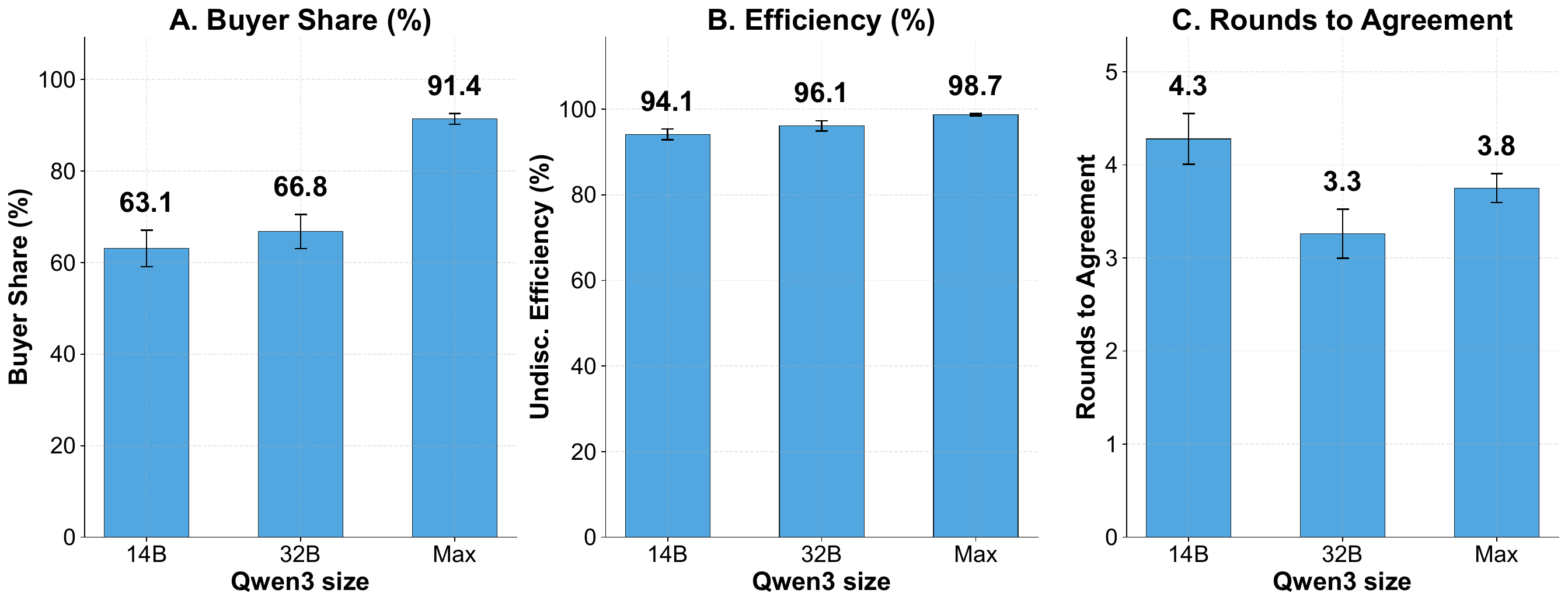}
\caption{Parameter Size Effects (Qwen3 Family)}
\par\smallskip
\begin{minipage}{\textwidth}
\scriptsize
\textit{Notes:} Panels show buyer surplus share, undiscounted efficiency, and rounds to agreement. The 14B model takes more rounds than 32B or Max ($p < 0.001$). The Max model achieves the highest efficiency and buyer share.
\end{minipage}
\label{fig:ext7_verbal}
\end{figure}

\paragraph{Irrational deals decline with size.} The rate of irrational deals declines with size: 5.4\% at 14B, 1.7\% at 32B, and 0.0\% at Max. This pattern suggests a within-family reliability gradient in the Qwen size comparison: value-destroying agreements become less frequent as model size rises within this architecture family. Because the ladder contains three Qwen points and Max's parameter count is undisclosed, we interpret it as within-family evidence rather than a general scaling law.

\clearpage
\section{Bayesian Benchmark Implementation Details}
\label{Bayesian-Details}
\subsection{Problem Setup and Notation}

Let the buyer's private type be $i \in \{H,L\}$ with prior $P(i=H)=\beta$.
A contract is a pair $(q,T)$, where $q$ is order quantity and $T$ is the transfer from buyer to seller.
The buyer's demand is $D_i \sim \mathcal{N}(\mu_i,\sigma_i^2)$.
In the baseline calibration,
\[
(\mu_H,\mu_L,\sigma_H,\sigma_L)=(80,40,10,10), \qquad r=60, \qquad c=30, \qquad \beta=0.5.
\]

If agreement occurs in round $\tau$, the buyer's and seller's discounted utilities are
\begin{align}
U_{B,i}^{(\tau)}(q,T) &= \delta_B^{\tau-1}\bigl(R_i(q)-T\bigr), \\
U_S^{(\tau)}(q,T) &= \delta_S^{\tau-1}\bigl(T-cq\bigr),
\end{align}
where
\begin{equation}
R_i(q)=r\,\mathbb{E}[\min(D_i,q)].
\end{equation}

For normal demand, $R_i(q)$ admits the closed form
\begin{equation}
R_i(q)
=
r\left[
\mu_i \Phi(z_i)-\sigma_i \phi(z_i)+q\bigl(1-\Phi(z_i)\bigr)
\right],
\qquad
z_i=\frac{q-\mu_i}{\sigma_i},
\end{equation}
with $\Phi(\cdot)$ and $\phi(\cdot)$ denoting the standard normal cdf and pdf.

The type-$i$ first-best quantity solves the newsvendor condition
\begin{equation}
F_i(\hat q_i)=\frac{r-c}{r},
\end{equation}
and the associated maximum total surplus is
\begin{equation}
\hat\pi_i = R_i(\hat q_i)-c\hat q_i.
\end{equation}

In our calibration, $(r-c)/r=0.5$, so $\hat q_H=80$ and $\hat q_L=40$.

\subsection{Complete-Information Building Blocks}

Under complete information, \citet{feng2015dynamic} reduce the game to Rubinstein bargaining over the type-specific surplus $\hat\pi_i$.

If the seller proposes first, the equilibrium contract is $(\hat q_i,\hat T_i^S)$ with
\begin{equation}
\hat T_i^S
=
c\hat q_i
+
\frac{1-\delta_B}{1-\delta_B\delta_S}\hat\pi_i.
\label{eq:appendixG_Ts}
\end{equation}
Hence the seller and buyer shares of surplus are
\begin{equation}
s_S^{S\text{-first}}=\frac{1-\delta_B}{1-\delta_B\delta_S},
\qquad
s_B^{S\text{-first}}=\frac{\delta_B(1-\delta_S)}{1-\delta_B\delta_S}.
\label{eq:appendixG_seller_first_shares}
\end{equation}

If the buyer proposes first, the equilibrium contract is $(\hat q_i,\hat T_i^B)$ with
\begin{equation}
\hat T_i^B
=
c\hat q_i
+
\frac{\delta_S(1-\delta_B)}{1-\delta_B\delta_S}\hat\pi_i.
\label{eq:appendixG_Tb}
\end{equation}
The corresponding shares are
\begin{equation}
s_S^{B\text{-first}}=\frac{\delta_S(1-\delta_B)}{1-\delta_B\delta_S},
\qquad
s_B^{B\text{-first}}=\frac{1-\delta_S}{1-\delta_B\delta_S}.
\label{eq:appendixG_buyer_first_shares}
\end{equation}

These complete-information objects are building blocks for the asymmetric-information benchmark.
They are not, by themselves, the full incomplete-information prediction.

\subsection{Patience Regions}

Following Proposition~2 of \citet{feng2015dynamic}, define the high-type information advantage at the low-type first-best quantity,
\begin{equation}
\Delta = R_H(\hat q_L)-R_L(\hat q_L),
\end{equation}
and
\begin{equation}
k=\frac{\Delta}{\hat\pi_H-\hat\pi_L}.
\end{equation}

The buyer's patience region is determined by
\begin{equation}
\bar\delta_B(\delta_S)=\frac{k}{1+\delta_S(k-1)},
\qquad
\underline\delta_B(\delta_S)=\frac{k-1+\delta_S}{k\delta_S}.
\label{eq:appendixG_thresholds}
\end{equation}

The buyer is in the HIGH region when $\delta_B \geq \bar\delta_B(\delta_S)$, in the MEDIUM region when $\underline\delta_B(\delta_S) \leq \delta_B < \bar\delta_B(\delta_S)$, and in the LOW region otherwise.

For our calibration, $k \approx 0.1995$.
When $\delta_S=0.9$, this gives
\[
\underline\delta_B(0.9)\approx 0.554,
\qquad
\bar\delta_B(0.9)\approx 0.714,
\]
so the three buyer-patience values used in the paper map to LOW ($0.4$), MEDIUM ($0.7$), and HIGH ($0.9$) exactly as intended.

\subsection{Asymmetric-Information Contract Objects}

The seller-initiated PBE in \citet{feng2015dynamic} is piecewise, with the realized branch depending on the patience region and on prior cutoffs within that region.
The benchmark implementation uses the following equilibrium objects.

\paragraph{Pooling / intermediate contract $(q_I,T_I)$.}
In the seller-first pooling region, the seller offers a single contract acceptable to both types:
\begin{equation}
T_I
=
R_H(q_I)-\frac{\delta_B(1-\delta_S)}{1-\delta_B\delta_S}\hat\pi_H
=
R_L(q_I)-\frac{\delta_B(1-\delta_S)}{1-\delta_B\delta_S}\hat\pi_L.
\label{eq:appendixG_TI}
\end{equation}

\paragraph{Seller-screening contract in the MEDIUM region.}
In the $(R,A)$-Scr branch of Proposition~4, the seller uses the same quantity object $q_I$ but perturbs the transfer by a small $\varepsilon>0$ so that only the low type accepts in round~1:
\begin{equation}
T_I
=
R_H(q_I)-\frac{\delta_B(1-\delta_S)}{1-\delta_B\delta_S}\hat\pi_H+\varepsilon
=
R_L(q_I)-\frac{\delta_B(1-\delta_S)}{1-\delta_B\delta_S}\hat\pi_L.
\label{eq:appendixG_TI_screen}
\end{equation}

\paragraph{Low-type signaling contract $(q_D,T_D)$.}
In the LOW buyer-patience region, the low type separates by distorting quantity downward.
The signaling contract $(q_D,T_D)$ with $q_D<\hat q_L$ solves
\begin{align}
R_H(q_D)-T_D &= \frac{1-\delta_S}{1-\delta_B\delta_S}\hat\pi_H,
\label{eq:appendixG_qd1}\\
T_D-cq_D &= \frac{\delta_S(1-\delta_B)}{1-\delta_B\delta_S}\hat\pi_L.
\label{eq:appendixG_qd2}
\end{align}
Equation~\eqref{eq:appendixG_qd1} makes the high type indifferent to mimicking, and Equation~\eqref{eq:appendixG_qd2} gives the seller the low-type continuation value.

\paragraph{Mixed low-patience counteroffer $(q_M,T_M)$.}
When the LOW-region equilibrium falls into the $(A,R)$-Mix branch of Proposition~5, the low type's round-2 counteroffer $(q_M,T_M)$ satisfies
\begin{align}
R_H(q_M)-T_M
&=
\min\left\{
\frac{R_H(q_S)-T_S}{\delta_B},
\;
R_H(\hat q_L)-\hat T_L^B
\right\},
\label{eq:appendixG_qm1}\\
T_M-cq_M
&=
\frac{\delta_S(1-\delta_B)}{1-\delta_B\delta_S}\hat\pi_L,
\label{eq:appendixG_qm2}
\end{align}
with $q_D \le q_M \le \hat q_L$.
The corresponding seller first-round transfer cutoff $\bar T_H^S$ is pinned down by the indifference condition in Proposition~5(a).

\paragraph{Mixed seller offer $(q_X,T_X)$.}
In the $(R,A)$-Mix branch of Proposition~5, the seller's round-1 offer $(q_X,T_X)$ satisfies $q_X<q_D$ and solves the fixed-point indifference conditions in Proposition~5(b).
These equations are implemented numerically in the validation agent.
They are not used as simple Rubinstein-share shortcuts.

\subsection{How the Benchmark Is Constructed in This Paper}

The incomplete-information PBE in \citet{feng2015dynamic} is derived for seller-initiated bargaining.
Our experimental design also includes buyer-first conditions, so the implemented benchmark separates two cases:
seller-first cells are benchmarked directly to Feng's branch-specific PBE, whereas buyer-first cells are benchmarked to a separately derived buyer-first separating equilibrium built from Feng's complete-information contracts and low-type signaling conditions.

\paragraph{Buyer-first conditions.}
Because \citet{feng2015dynamic} do not derive the buyer-initiated game formally, we verify here the separating buyer-first equilibrium used in the benchmark implementation. Nature draws buyer type $i\in\{H,L\}$ with prior $\beta$; the buyer observes $i$ and proposes $(q,T)$ in round~1; the seller updates beliefs and accepts or rejects; after rejection, the game continues with the seller proposing in round~2. For a degenerate posterior on type $i$, the seller's round-1 continuation value from rejecting and entering the round-2 complete-information seller-proposer subgame is
\[
V_{S,i}^{\mathrm{cont}}
=
\delta_S\frac{1-\delta_B}{1-\delta_B\delta_S}\hat\pi_i.
\]
Hence any accepted round-1 type-$i$ contract must give the seller at least $V_{S,i}^{\mathrm{cont}}$.

The buyer-first benchmark uses the following candidate separating offers:
\[
(q_H^\star,T_H^\star)=(\hat q_H,\hat T_H^B),
\]
and
\[
(q_L^\star,T_L^\star)=
\begin{cases}
(\hat q_L,\hat T_L^B), & \delta_B\geq \underline\delta_B(\delta_S),\\
(q_D,T_D), & \delta_B<\underline\delta_B(\delta_S),
\end{cases}
\]
where $(\hat q_i,\hat T_i^B)$ is the buyer-proposer complete-information contract from Equation~\eqref{eq:appendixG_Tb}, $(q_D,T_D)$ is the low-type signaling contract from Equations~\eqref{eq:appendixG_qd1}--\eqref{eq:appendixG_qd2}, and $\underline\delta_B(\delta_S)$ is the lower patience cutoff in Equation~\eqref{eq:appendixG_thresholds}.

Seller participation binds on path because the buyer optimally offers the lowest transfer consistent with acceptance. For $i\in\{H,L\}$, substituting Equation~\eqref{eq:appendixG_Tb} yields
\[
\hat T_i^B-c\hat q_i
=
\delta_S\frac{1-\delta_B}{1-\delta_B\delta_S}\hat\pi_i
=
V_{S,i}^{\mathrm{cont}},
\]
so the seller is indifferent and accepts by convention. In the low-patience signaling regime, Equation~\eqref{eq:appendixG_qd2} gives the same equality for $(q_D,T_D)$ with $i=L$.

Type-$H$ incentive compatibility determines the threshold. If type $L$ offers $(\hat q_L,\hat T_L^B)$, type $H$'s payoff from mimicking is
\[
R_H(\hat q_L)-\hat T_L^B
=
\hat\pi_L+\Delta-\delta_S\frac{1-\delta_B}{1-\delta_B\delta_S}\hat\pi_L,
\]
where $\Delta=R_H(\hat q_L)-R_L(\hat q_L)$. Comparing this with type $H$'s equilibrium payoff
\[
R_H(\hat q_H)-\hat T_H^B
=
\frac{1-\delta_S}{1-\delta_B\delta_S}\hat\pi_H
\]
shows that $H$ weakly prefers its own contract iff
\[
\frac{1-\delta_S}{1-\delta_B\delta_S}\bigl(\hat\pi_H-\hat\pi_L\bigr)\geq \Delta,
\]
which is equivalent to
\[
\frac{1-\delta_S}{1-\delta_B\delta_S}\geq k
\qquad\Longleftrightarrow\qquad
\delta_B\geq \underline\delta_B(\delta_S).
\]
Thus, above the cutoff, the no-distortion low-type offer $(\hat q_L,\hat T_L^B)$ is incentive compatible; below the cutoff, type $H$ would mimic it. In that low-patience region, the signaling contract $(q_D,T_D)$ restores separation because Equation~\eqref{eq:appendixG_qd1} makes type $H$ exactly indifferent between its own contract and mimicking the low type, while Equation~\eqref{eq:appendixG_qd2} preserves seller participation.

Type-$L$ does not mimic type $H$. The key monotonicity is that $R_H(q)-R_L(q)$ is increasing in $q$, so higher quantities are relatively more attractive to the high type. In the no-distortion regime, once type $H$ weakly prefers $(\hat q_H,\hat T_H^B)$ to $(\hat q_L,\hat T_L^B)$, type $L$ strictly prefers $(\hat q_L,\hat T_L^B)$ to $(\hat q_H,\hat T_H^B)$. In the low-patience regime, the same single-crossing logic implies that if type $H$ is exactly indifferent between $(\hat q_H,\hat T_H^B)$ and $(q_D,T_D)$, then type $L$ strictly prefers $(q_D,T_D)$ to the high-type contract. Therefore type $L$ never gains from mimicking type $H$.

To complete the equilibrium, we use off-path beliefs that attribute a deviation to the type for whom it could be profitable under acceptance. If a deviation is attributed to type $H$, the seller applies the posterior-$H$ participation threshold, and no accepted deviation improves on $(\hat q_H,\hat T_H^B)$ because that contract already solves type $H$'s round-1 maximization subject to seller participation under posterior $H$. If a deviation is attributed to type $L$, the seller applies the posterior-$L$ participation threshold, and no accepted deviation improves on the relevant low-type benchmark contract: $(\hat q_L,\hat T_L^B)$ in the no-distortion region, or $(q_D,T_D)$ in the signaling region, where the high-type no-mimicking constraint also binds. Pooling outcomes are ruled out by the same logic: type $H$ can deviate to an offer that is strictly preferred under posterior $H$ and unattractive to type $L$, so the seller's off-path beliefs do not sustain a pooling allocation. These beliefs therefore support the separating allocation above. For the benchmark implementation, this existence result is sufficient; we do not require a stronger uniqueness claim.

Under this buyer-first separating equilibrium, both types settle in round~1. In the main design with $\delta_S=0.9$, the threshold $\underline\delta_B(0.9)\approx 0.554$ places the HIGH ($\delta_B=0.9$) and MEDIUM ($\delta_B=0.7$) conditions in the no-distortion regime and the LOW ($\delta_B=0.4$) condition in the signaling regime. In the seller-impatient condition $(\delta_B,\delta_S)=(0.9,0.4)$, the threshold is negative, so the no-distortion regime applies. These are the buyer-first benchmark outcomes implemented in the validation agent and summarized in Table~\ref{tab:appendixG_mapping}; Figure~\ref{fig:truth_telling_regime} visualizes the corresponding truth-telling regions under the baseline calibration.

Relative to Feng's seller-first PBE, only the lower threshold $\underline\delta_B(\delta_S)$ matters here. The upper threshold $\bar\delta_B(\delta_S)$ separates pooling from screening in the seller-first game but has no analogue in the buyer-first game because the informed party proposes, eliminating the uninformed-proposer pooling temptation. Prior cutoffs on $\beta$ likewise play no role: separation is achieved by the buyer's own offer, so the seller's posterior is degenerate on path regardless of the prior.

\paragraph{Seller-first conditions.}
For seller-initiated cells, we benchmark directly to Feng's PBE.
We first classify the patience region using Proposition~2 and then solve the relevant branch of Propositions~3--5 at $(\delta_B,\delta_S,\beta)$.

Within each patience region, Feng defines prior cutoffs
\[
\beta_h,\ \bar\beta_h,\ \beta_m,\ \bar\beta_m,\ \beta_\ell,\ \bar\beta_\ell
\]
that separate pooling, screening, signaling, and mixed branches.
We compute these cutoffs numerically and then solve the corresponding branch-specific contract equations.
Criteria~1--3 and Lemmas~1--2 determine how rejected seller offers are interpreted and which buyer counteroffers are belief-revealing, pooling, or mixed.

In the main 16-condition design, $\beta=0.5$ in every cell.
Under this calibration, the seller-first benchmark reported in the paper has the following realized timing pattern:
\begin{itemize}
\item type-$H$ buyers accept the seller's round-1 offer, and
\item type-$L$ buyers reject in round~1 and make the equilibrium round-2 counteroffer, which the seller accepts.
\end{itemize}

In HIGH and MEDIUM patience, the type-$L$ round-2 counteroffer is $(\hat q_L,\hat T_L^B)$.
In LOW patience, the type-$L$ round-2 counteroffer is the low-type equilibrium separating contract obtained from Proposition~5, that is, the signaling contract $(q_D,T_D)$ or the corresponding mixed-case object when the numerically selected branch requires it.

This yields exactly the benchmark timing reported in the main text:
\begin{align}
\text{Buyer-first benchmark rounds} &= 1.00, \\
\text{Seller-first benchmark rounds} &= 1.50, \\
\text{Overall benchmark rounds} &= 1.25.
\end{align}

\begin{table}[htbp]
\centering
\caption{Bayesian Benchmark Mapping for the Main 16-Condition Design ($\beta=0.5$)}
\label{tab:appendixG_mapping}
\footnotesize
\setlength{\tabcolsep}{3pt}
\renewcommand{\arraystretch}{1.1}
\begin{tabular}{@{}llll@{}}
\toprule
\textbf{First proposer} & \textbf{Buyer patience} & \textbf{Type-$H$ path} & \textbf{Type-$L$ path} \\
\midrule
Buyer & \makecell[l]{HIGH /\\MEDIUM} & \makecell[l]{$(\hat q_H,\hat T_H^B)$,\\accept in round~1} & \makecell[l]{$(\hat q_L,\hat T_L^B)$,\\accept in round~1} \\
Buyer & LOW & \makecell[l]{$(\hat q_H,\hat T_H^B)$,\\accept in round~1} & \makecell[l]{$(q_D,T_D)$,\\accept in round~1} \\
Seller & \makecell[l]{HIGH /\\MEDIUM} & \makecell[l]{seller offer accepted\\in round~1} & \makecell[l]{reject in round~1, then\\$(\hat q_L,\hat T_L^B)$ in round~2} \\
Seller & LOW & \makecell[l]{seller offer accepted\\in round~1} & \makecell[l]{reject in round~1, then LOW-region\\equilibrium counteroffer in round~2} \\
\bottomrule
\end{tabular}%

\par\smallskip
\begin{minipage}{\textwidth}
\scriptsize
\textit{Notes:} The LOW-region seller-first counteroffer is $(q_D,T_D)$ in the pure signaling branch and the corresponding mixed-case counteroffer when Proposition~5 selects a mixed branch numerically.
\end{minipage}
\end{table}

\subsection{Validation and Interpretation}

For each experimental cell, the benchmark implementation performs four steps:
\begin{enumerate}[(1)]
\item compute $\hat q_i$ and $\hat\pi_i$ for $i\in\{H,L\}$;
\item classify the patience region using Equation~\eqref{eq:appendixG_thresholds};
\item if the seller proposes first, solve the branch-specific equilibrium object(s) required by Propositions~3--5; if the buyer proposes first, construct the type-contingent buyer proposal from the buyer-initiated complete-information contract stated after Proposition~1, Proposition~2, and Lemma~2;
\item execute the implied accept/reject path and record agreement round, contract terms, profits, and surplus shares.
\end{enumerate}

This validation benchmark reproduces the analytical properties used in the paper:
(i) immediate agreement in all buyer-first cells,
(ii) agreement in at most two rounds in all seller-first cells,
(iii) truthful first-best quantities whenever the equilibrium calls for full revelation, and
(iv) branch-specific quantity distortion whenever separation requires signaling.

A final clarification is important.
The Rubinstein shares in Equations~\eqref{eq:appendixG_seller_first_shares}--\eqref{eq:appendixG_buyer_first_shares} are complete-information building blocks.
They are not the full asymmetric-information surplus benchmark.
Whenever the paper refers to the ``Bayesian benchmark,'' it means the branch-specific incomplete-information implementation described above, not Rubinstein shares taken in isolation.

\begin{figure}[ht]
\centering
\includegraphics[width=0.7\textwidth]{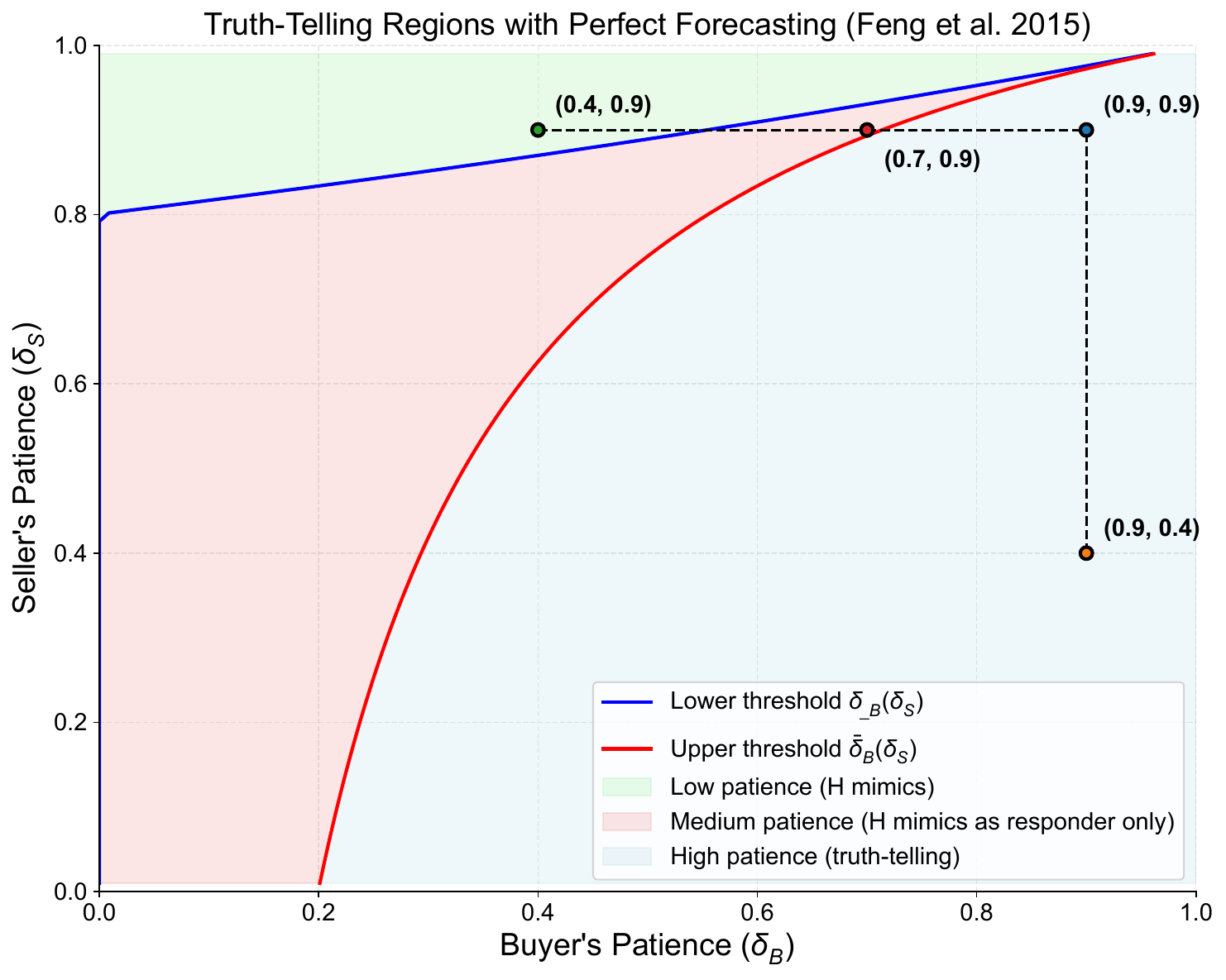}
\caption{Buyer Truth-Telling Regions Under the Baseline Calibration}
\label{fig:truth_telling_regime}
\par\smallskip
\begin{minipage}{\textwidth}
\scriptsize
\textit{Notes:} The figure reproduces Proposition~2 of \citet{feng2015dynamic}. The calibration is $r=60$, $c=30$, $\beta=0.5$, $D_H \sim \mathcal{N}(80,10^2)$, and $D_L \sim \mathcal{N}(40,10^2)$. Under this calibration, $\hat q_H=80$, $\hat q_L=40$, and the buyer-patience values used in the main design map cleanly into the LOW, MEDIUM, and HIGH regions.
\end{minipage}
\end{figure}

\end{APPENDICES}

\end{document}